\documentclass[10pt,journal,compsoc]{IEEEtran}

\usepackage{caption}
\usepackage{graphicx}
\usepackage{comment}
\usepackage{amsmath} %
\usepackage{amssymb}
\usepackage{array}
\usepackage{color}
\usepackage{epsfig}
\usepackage{marvosym}
\usepackage{multirow}
\usepackage{float}
\usepackage{xspace}
\usepackage{lineno}
\usepackage{amsfonts}
\usepackage{booktabs,siunitx}
\usepackage{bbm}
\usepackage[colorlinks,
            linkcolor=red,
            anchorcolor=red,
            citecolor=blue]{hyperref}
\usepackage{fixltx2e}
\usepackage{mathrsfs}

\ifCLASSOPTIONcompsoc
  \usepackage[nocompress]{cite}
\else
  \usepackage{cite}
\fi

\ifCLASSINFOpdf
\else
\fi

\newcommand{\bx}{\mathbf{x}}

\newcommand{\bS}{\mathbf{S}}
\newcommand{\bs}{\mathbf{s}}

\newcommand{\bt}{\mathbf{t}}

\newcommand{\bD}{\mathbf{D}}
\newcommand{\bz}{\mathbf{z}}

\newcommand{\bC}{\mathbf{C}}

\newcommand{\bc}{\mathbf{c}}

\newcommand{\bT}{\mathbf{T}}

\newcommand{\bd}{\mathbf{d}}
\newcommand{\br}{\mathbf{r}}

\newcommand{\nR}{\mathbb{R}}

\newcommand{\cL}{\mathcal{L}}

\makeatletter
\DeclareRobustCommand\onedot{\futurelet\@let@token\@onedot}
\def\@onedot{\ifx\@let@token.\else.\null\fi\xspace}
\def\eg{e.g\onedot} 
\def\ie{i.e\onedot}

\def\etal{\textit{et~al}\onedot} 
\def\Fig{Fig\onedot}   
\makeatother

\newcommand{\figref}[1]{\Fig~\ref{#1}}
\newcommand{\secref}[1]{Section~\ref{#1}}

\renewcommand{\eqref}[1]{Eq.~\ref{#1}}
\newcommand{\tabref}[1]{Table~\ref{#1}}

\newcommand*\rot{\rotatebox{90}}
\newcommand{\boldparagraph}[1]{\vspace{0.2cm}\noindent{\bf #1:} }

\newif\ifcomment
\commenttrue
\ifcomment
	\newcommand{\yl}[1]{ \noindent {\color{cyan} {\bf Yiyi:} {#1}} }
	
	\newcommand{\md}[1]{\noindent{\color{black}{#1}}}
	\newcommand{\mdi}[1]{\noindent{\color{black}{#1}}}
\else
	\newcommand{\ag}[1]{}
	\newcommand{\yl}[1]{}
\fi

\begin{document}

\title{PanopticNeRF-360: Panoramic \\ 3D-to-2D Label Transfer in Urban Scenes}

\author{Xiao~Fu,~
        Shangzhan~Zhang,~
        Tianrun~Chen,~
        Yichong~Lu,~
        Xiaowei~Zhou,~
        Andreas~Geiger,~
        Yiyi~Liao $\textsuperscript{\Letter}$~%
\IEEEcompsocitemizethanks{
\IEEEcompsocthanksitem X. Fu, S. Zhang, T. Chen, Y. Lu, X. Zhou and Y. Liao are affiliated with Zhejiang University, China.\protect
\IEEEcompsocthanksitem A. Geiger is affiliated with the Autonomous Vision Group (AVG) at the University of Tübingen and Tübingen AI Center, Germany.\protect
\IEEEcompsocthanksitem $\textsuperscript{\Letter}$Corresponding author.\protect
\IEEEcompsocthanksitem Project page: \href{https://fuxiao0719.github.io/projects/panopticnerf360/}{https://fuxiao0719.github.io/projects/panopticnerf360/}\protect
\IEEEcompsocthanksitem Code and data: \href{https://github.com/fuxiao0719/panopticnerf}{https://github.com/fuxiao0719/panopticnerf}\protect

}%
}

\markboth{ }
{Shell \MakeLowercase{\textit{et al.}}: Bare Advanced Demo of IEEEtran.cls for IEEE Computer Society Journals}

\IEEEtitleabstractindextext{%
\begin{abstract}
Training perception systems for self-driving cars requires substantial 2D annotations that are labor-intensive to manual label. While existing datasets provide rich annotations on pre-recorded sequences, they fall short in labeling rarely encountered viewpoints, potentially hampering the generalization ability for perception models. In this paper, we present PanopticNeRF-360, a novel approach that combines coarse 3D annotations with noisy 2D semantic cues to generate high-quality panoptic labels and images from any viewpoint. Our key insight lies in exploiting the complementarity of 3D and 2D priors to mutually enhance geometry and semantics. Specifically, we propose to leverage coarse 3D bounding primitives and noisy 2D semantic and instance predictions to guide geometry optimization, by encouraging predicted labels to match panoptic pseudo ground truth. Simultaneously, the improved geometry assists in filtering 3D\&2D annotation noise by fusing semantics in 3D space via a learned semantic field. To further enhance appearance, we combine MLP and hash grids to yield hybrid scene features, striking a balance between high-frequency appearance and contiguous semantics. Our experiments demonstrate PanopticNeRF-360's state-of-the-art performance over label transfer methods on the challenging urban scenes of the KITTI-360 dataset. Moreover, PanopticNeRF-360 enables omnidirectional rendering of high-fidelity, multi-view and spatiotemporally consistent appearance, semantic and instance labels.

\end{abstract}

}

\maketitle
\IEEEdisplaynontitleabstractindextext
\IEEEpeerreviewmaketitle
\ifCLASSOPTIONcompsoc
\IEEEraisesectionheading{\section{Introduction}\label{sec:introduction}}
\else
\section{Introduction}
\label{sec:introduction}
\fi

\begin{figure*}[ht]
\centering
\includegraphics[width=\linewidth]{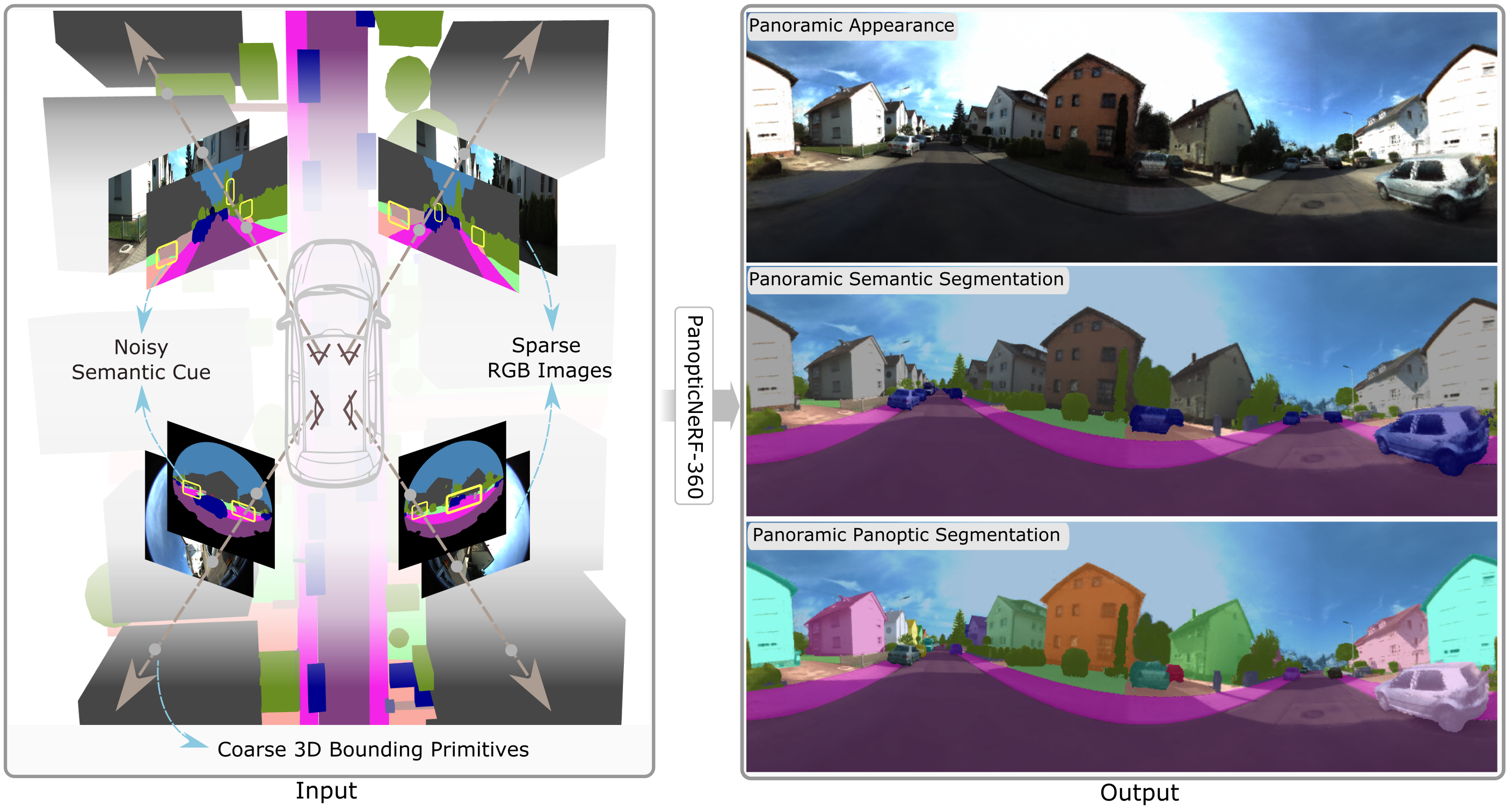}
\caption{\textbf{PanopticNeRF-360} takes as input a set of sparse forward-facing stereo images and two side-facing fisheye images, coarse 3D bounding primitives and noisy 2D semantic predictions (yellow boxes highlight inaccurate predictions). By inferring in 3D space, our model generates multi-view consistent semantic and instance labels in 2D image space via volume rendering. Our holistic formulation allows for rendering panoramic appearance and label maps~(output).}
\label{fig:teaser}
\end{figure*}

\IEEEPARstart{I}{n} autonomous driving, large-scale semantic instance annotations of real-world scenes are foundational for bootstrapping perception models~\cite{vaswani2017attention,dosovitskiy2020image,he2022masked}. Manually annotation of semantic and instance masks at the pixel level is recognized to be labor-intensive and costly. For example, a single street scene image demands approximately 1.5 hours for comprehensive instance annotation~\cite{cordts2016cityscapes}. Furthermore, going beyond the traditional task of annotating pre-recorded data from fixed viewpoints, offering RGB images and annotations from various new viewpoints carries significant value. This augmentation of viewpoint diversity holds the potential to enhance the generalization capabilities of perception models. Therefore, in this paper, we aim to devise a semi-automated framework that involves relatively low-cost human labor for generating high-fidelity labels and RGB images from both pre-recorded and novel viewpoints. 

3D-to-2D label transfer has great potential in this area~\cite{xie2016semantic,liao2021kitti,huang2019apolloscape}. By annotating coarse bounding primitives in 3D space and propagating these manually annotated coarse 3D primitives to dense and multi-view consistent 2D semantic and instance annotations, it significantly reduces the labeling time to 0.75 minutes per image~\cite{liao2021kitti}, yielding a $\sim$120x speedup compared to 2D per-pixel labeling~\cite{cordts2016cityscapes}. Existing methods \cite{xie2016semantic,liao2021kitti} automatically infer dense 2D labels leveraging 3D annotations, 2D pre-trained models (\eg, noisy 2D semantic predictions) and 2D image cues via conditional random fields (CRF). This CRF-based methodology necessitates intermediary 3D reconstructions for the projection of non-occluded 3D points. A limitation lies in the fact that the 3D reconstruction cannot be collaboratively and jointly optimized within the CRF framework, and as a consequence, any inaccuracies in the reconstruction phase propagate to the label transfer outputs. Besides, these methods are incapable of transferring labels to novel viewpoints.

\begin{figure*}[ht]
\centering
\includegraphics[width=\textwidth]{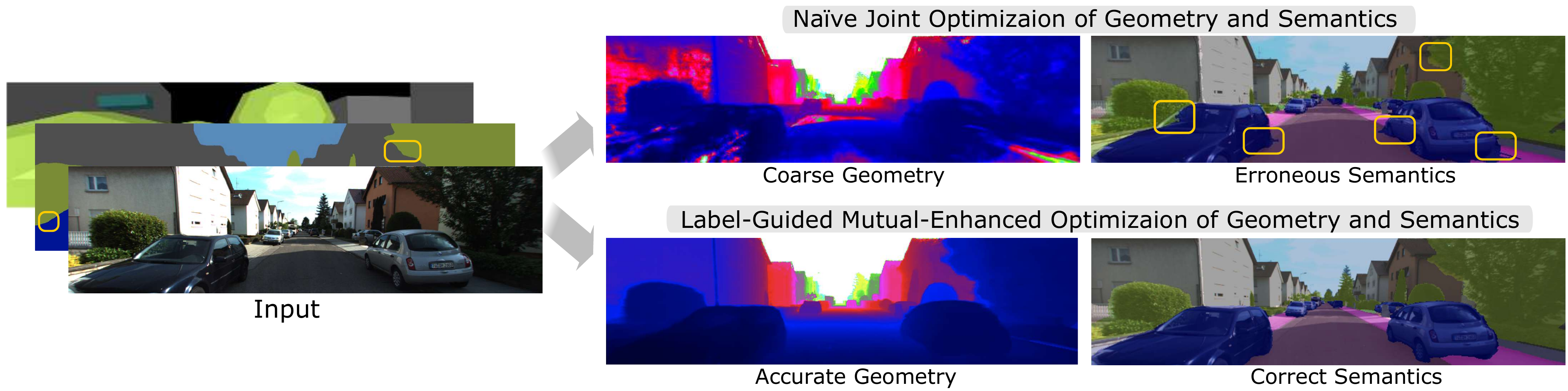}
\caption{\textbf{Comparison of Different Geometry and Semantic Optimization Methods.} Compared to the na\"ive joint optimization of geometry and semantics (top), ours leads to mutual enhancement of geometry and semantics (bottom).
}
\label{fig:naive_comparison}
\end{figure*}

In this paper, we introduce PanopticNeRF-360, a novel method that utilizes a $360^{\circ}$ Neural Radiance Field (NeRF)~\cite{mildenhall2020nerf} to estimate geometry and semantics in a joint and differentiable manner.  
PanopticNeRF-360 takes as input a set of sparse forward-facing stereo images and two side-facing fisheye images, as well as coarse 3D annotations and noisy 2D semantic predictions. By inferring semantic and instance labels in 3D space, the model renders dense 2D semantic and instance labels, \ie, panoptic segmentation labels~\cite{kirillov2019panoptic}, at novel viewpoints (see ~\figref{fig:teaser}). Note that our model enables rendering images from diverse viewpoints and even panoramic images, by incorporating the fisheye images of a large field of view. However, obtaining accurate geometry and semantics is non-trivial in urban scenes. In driving scenarios, where we have sparse input views with frequent over-exposure (particularly common for fisheye views due to directly facing of the sun), reconstructing high-quality geometry using NeRF is challenging. 
Moreover, inferring precise semantics in 3D space is also difficult given imprecise geometry and coarse 3D annotations with many overlapping regions. Errors in geometric reconstruction (e.g., 3D floaters enclosed by a 3D bounding primitive) and label ambiguity of the 3D coarse annotations (e.g., overlapping regions of car and road) can negatively impact the label transfer step, leading to incorrect 2D semantic and instance labels.
While several prior works have modeled 3D semantic fields~\cite{zhi2021place,vora2021nesf,xu2023jacobinerf}, we make a surprising key observation: As illustrated in~\figref{fig:naive_comparison},  na\"ive joint optimization of geometry and semantics does not necessarily yield mutual improvements in our setting.

\begin{figure}[t]
\centering
\includegraphics[width=\linewidth]{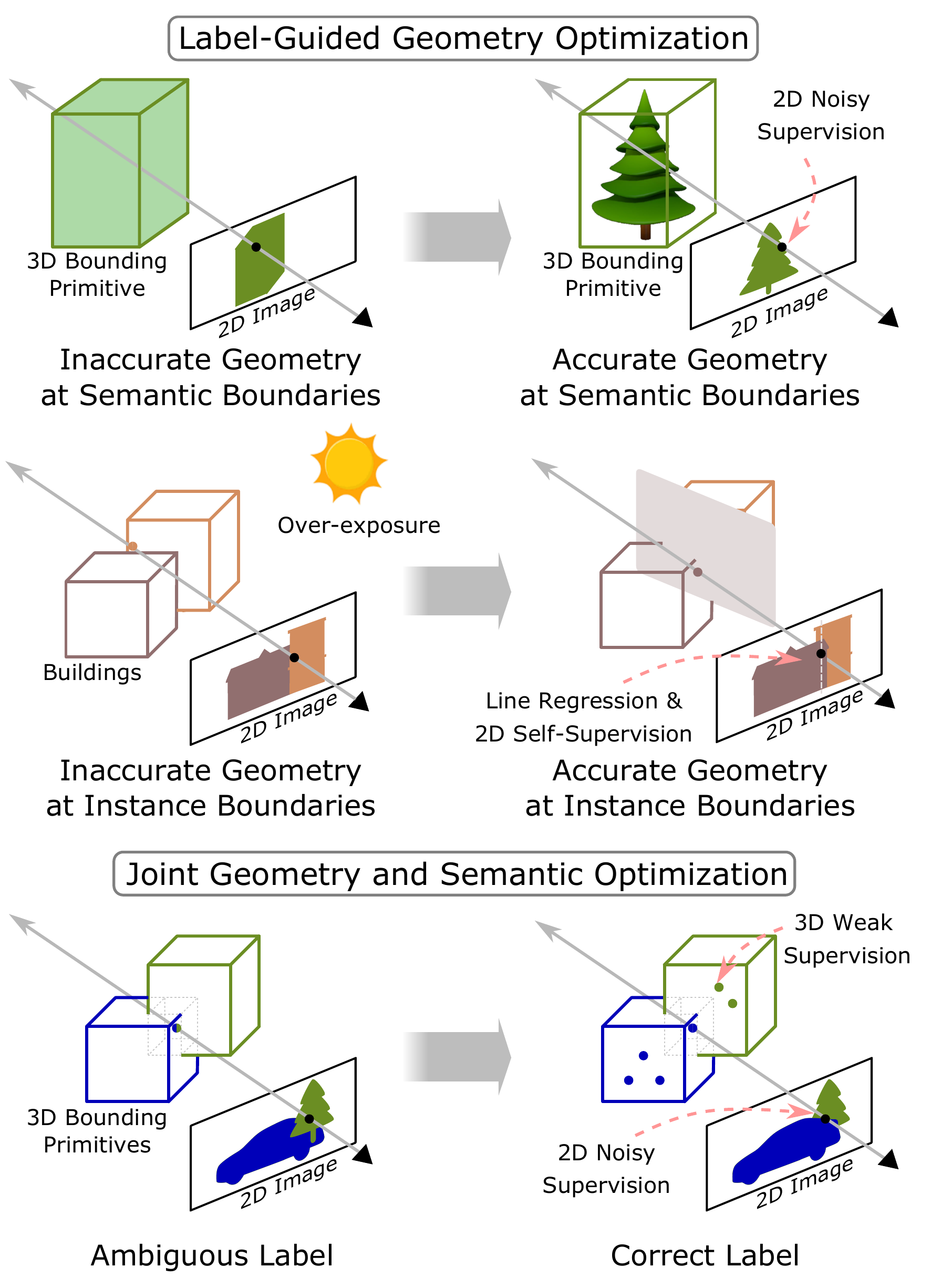}
\caption{\textbf{Challenges and Solutions.} We propose semantic and instance label-guided geometry optimization to improve the underlying geometry. This allows for rendering accurate semantic and instance labels when coarse 3D bounding primitives correctly enclose the corresponding target without ambiguity. We further resolve label ambiguity at intersection regions of the 3D bounding primitives via joint geometry and semantic optimization. 
}
\label{fig:challenges}
\end{figure}

To tackle these challenges, we aim for \textit{mutual enhancement of geometry and semantics} by introducing two auxiliary parameter-free semantic and instance fields to regularize density estimation, followed by joint optimization of geometry and semantics to resolve label ambiguity, as illustrated in \figref{fig:challenges}. The results of our model are shown at the bottom of ~\figref{fig:naive_comparison}, demonstrating higher accuracy and consistency than the baseline. Firstly, we propose label-guided geometry optimization, utilizing coarse 3D bounding primitives and 2D pseudo labels to guide geometry optimization. In particular, we render panoptic labels utilizing the deterministic 3D bounding primitives (which we refer to as a fixed semantic/instance field as it provides non-trainable semantic and instance logits) and learned density fields, and encourage the rendered labels to match a 2D panoptic pseudo ground truth. As evidenced by our experiments, this leads to significant improvements of the density field despite the pseudo ground truth being noisy. This pseudo ground truth consists of \textit{semantic} predictions from pre-trained 2D segmentation models and \textit{instance} labels derived from a simple geometric prior\footnote{For label-guided geometry optimization, we only consider instance labels of buildings, as buildings are the most frequently connected class that may yield wrong geometry at the adjacent boundary.}. Secondly, we propose a joint geometry and semantic optimization strategy to improve semantics. Specifically, conditioned on the improved geometry, we learn to predict semantic categorical 3D logits to match semantic points in 3D bounding primitives and its corresponding 2D distribution via volume rendering to match the 2D noisy predictions. This enables resolving label ambiguity of the 3D bounding primitives between different semantic classes and substantially mitigates noise in the 2D predictions. Note that depsite that 2D semantic predictions of fisheye images may be of low quality due to the lack of training data, our method resolves the noise thanks to the holistic design of utilizing the weak 3D labels and the 2D noisy predictions, enabling rendering improved panoptic labels at arbitrary viewpoints. Despite rendering satisfactory panoptic labels, attaining high-quality appearance remains a problem, as the semantic label is contiguous across the same object/stuff while the corresponding appearance can contain high-frequency details. To tackle this problem, PanopticNeRF-360 combines features of a deep MLP and multi-resolution hash grids to model semantics and appearance. This allows us to leverage the smooth inductive bias of MLPs for semantics and the expressive local hash features for appearance, enabling rendering panoptic labels and photorealistic RGB images from arbitrary novel viewpoints.  

We conduct extensive experiments on the KITTI-360 dataset and showcase that our generalization ability on the Waymo dataset. As evidenced by our experimental results, PanopticNeRF-360 showcases state-of-the-art performance and outperforms existing 3D-to-2D and 2D-to-2D label transfer methods and demonstrates a promising path toward the efficient generation of densely annotated datasets that are pivotal for the advancement of autonomous driving systems. In summary, our contributions are as follows: 
(1) We present the first model that tackles 3D-to-2D label transfer via neural rendering. Towards this goal, we unify easy-to-obtain 3D bounding primitives and noisy 2D semantic predictions in a single model, yielding high-quality panoptic labels and high-frequency imagery. 
(2) We propose novel optimization strategies to enable mutual improvement of geometry and semantics. Our label-guided geometry optimization shows that the underlying geometry can be effectively improved by leveraging noisy 3D and 2D panoptic labels.
(3) PanopticNeRF-360 achieves state-of-the-art performance compared to existing label transfer methods in terms of both semantic and instance predictions in challenging urban scenes. Further, PanopticNeRF-360 enables omnidirectional rendering of high-fidelity appearance and spatio-temporally consistent panoptic labels, providing labeled data from novel viewpoints to potentially enhance the generalization ability of perception models.

\boldparagraph{Relation to PanopticNeRF~\cite{fu2022panoptic}} This paper is an extension of our earlier conference paper PanopticNeRF~\cite{fu2022panoptic}. We improve upon~\cite{fu2022panoptic} in several aspects: We (1) extend perspective label transfer to omnidirectional 360$^{\circ}$ label transfer, (2) incorporate instance labels into label-guided geometry optimization, thus achieving panoptic label-guided geometry optimization, (3) achieve higher quality semantics (0.8 on forward-facing mIoU) and instances (2.3 on forward-facing PQ), (4) improve scene features from pure MLPs to a hybrid of MLPs and grids for improved appearance ($\sim$4dB) in less training time ($\sim$2.5x speedup), and (5) extensively enrich the experimental section by including new label transfer results on fisheye views and comparison to more recent baselines.

\section{Related Work}

\boldparagraph{Urban Scene Segmentation} Semantic instance segmentation is a critical task for autonomous vehicles~\cite{ohn2020learning,Zhou2019SR}. Learning-based algorithms have achieved compelling performance~\cite{chen2017deeplab,li2020semantic,zhao2017pyramid}, but rely on large-scale training data. Unfortunately, annotating images at the pixel level is extremely time-consuming and labor-intensive, especially for instance-level annotation. SAM~\cite{kirillov2023segment} demonstrates extraordinary generalizable ability in zero-shot object boundary segmentation, but it lacks an understanding of high-level semantics. While most urban datasets provide labels in 2D image space~\cite{brostow2009semantic,cordts2016cityscapes,neuhold2017mapillary,Kim2020CVPR,weber2021step}, autonomous vehicles are usually equipped with 3D sensors~\cite{geiger2012we,caesar2020nuscenes,Geyer2020ARXIV,huang2019apolloscape,liao2021kitti}, allowing to exploit 3D information for labeling. KITTI-360~\cite{liao2021kitti} demonstrates that annotating the scene in 3D can significantly reduce annotation time. However, transferring coarse 3D labels to 2D remains challenging. In this work, we focus on developing a novel 3D-to-2D label transfer method exploiting recent advances in neural scene representations. 

\boldparagraph{Label Transfer} Several prior works have investigated how to label individual frames more efficiently~\cite{liu2011nonparametric,Guillaumin2014IJCV,castrejon2017annotating,acuna2018efficient,ling2019fast,andriluka2018fluid}. In this paper we focus on efficient labeling of video sequences. Existing works in this area fall into two categories: 2D-to-2D and 3D-to-2D. 2D-to-2D label transfer approaches reduce the workload by propagating labels across 2D images~\cite{pathak2015constrained,hong2016learning,papandreou2015weakly,pinheiro2015image,ganeshan2021warp} or transferring labels from frontal views to Bird's-Eye-View (BEV) maps~\cite{gosala2023skyeye,gosala2022bird}, whereas 3D-to-2D methods exploit additional information in 3D for efficient labeling~\cite{huang2019apolloscape,martinovic20153d,mustafa2017semantically,xiao2009multiple,bruls2018mark,zimmer20193d,mei2022waymo}. 
To obtain dense labels in 2D image space, one line of 3D-to-2D methods requires tedious preprocessing in the 3D space~\cite{hua2016scenenn}. Another line of methods instead projects coarse 3D labels to the 2D image space and manually refines the labels in 2D~\cite{tylecek2018consistent,huang2019apolloscape}. The state-of-the-art works~\cite{liao2021kitti,xie2016semantic} perform per-frame inference jointly over the 3D point clouds and 2D pixels using a non-local multi-field CRF model, avoiding manual pre- or post-processing. However, these methods require reconstructing a 3D mesh to project 3D point clouds to 2D. The mesh reconstruction is not jointly optimized in the CRF model as it is treated as a pre-processing step, leading to inaccurate reconstruction that hinders label transfer performance. In contrast, PanopticNeRF-360 provides a novel end-to-end method for 3D-to-2D label transfer where geometry and semantics are jointly optimized. 

\boldparagraph{Semantic-informed NeRFs} Recently, NeRF~\cite{mildenhall2020nerf} emerged as a novel powerful representation for novel view synthesis. Semantic NeRF~\cite{zhi2021place} initially augments NeRF~\cite{mildenhall2020nerf} with a semantic branch to encode multi-view consistent semantics from noisy 2D semantic segmentations. However, Semantic NeRF takes as input ground truth 2D labels or synthetic noisy labels (test denoising ability), and faces difficulties generating accurate labels given real-world predictions from pre-trained 2D models. Additionally, Semantic NeRF operates in indoor scenes with dense RGB inputs and degenerates in challenging outdoor driving scenarios with sparse input views~\cite{KunduCVPR2022PNF}, as also shown in our experiments. Jacobian NeRF~\cite{xu2023jacobinerf} further enhances semantic synergies of correlated entities via contrastive learning on self-supervised visual features for more effective label propagation. In contrast, NeSF~\cite{vora2021nesf} emphasizes generalizable semantic field learning via a feed-forward 3D U-Net supervised by 2D GT labels, whereas we concentrate on the 3D-to-2D label transfer task without access to 2D GT. Another line of work distills~\cite{tschernezki2022neural,kobayashi2022decomposing,turki2023suds}  abstract visual features from zero-shot vision encoders~(\eg, \cite{caron2021emerging,li2022language}) into 3D space for scene editing but does not render precise semantic labels. While semantic-enabled NeRFs~\cite{zhi2021place, zarzar2022segnerf,mirzaei2022spin} are limited to the semantic domain, a natural extension is to explore fine-grained instance information. PNF~\cite{KunduCVPR2022PNF} extracts things with the aid of an off-the-shelf object detector. DM-NeRF~\cite{wang2022dm} requires ground truth instance annotations for 3D geometry decomposition. Panoptic Lifting~\cite{siddiqui2022panoptic}, Nerflets~\cite{Zhang_2023_CVPR}, PCFF~\cite{Cheng_2023_CVPR}, and Instance NeRF~\cite{hu2023instance} leverage predicted instance/panoptic labels for a compositional panoptic scene representation. SUDS~\cite{turki2023suds} uses geometric clustering to label instances and assign semantic labels based on DINO features. However, this simple clustering framework leads to unsatisfying instance results. While these works mainly focus on scene parsing in close-domain classes~(\eg, Cityscapes~\cite{cordts2016cityscapes} encompasses only 30 classes), our goal is to transfer 3D annotations of arbitrary classes to 2D image space to foster the development of new datasets, \eg, providing instance labels for buildings that are not available in Cityscapes~\cite{cordts2016cityscapes}. Furthermore, we are the first to study $360^{\circ}$ outward panoptic urban scene understanding through the powerful lens of neural rendering.

\section{Methodology}
\begin{figure*}[tb]
\centerline{\includegraphics[width=\textwidth]{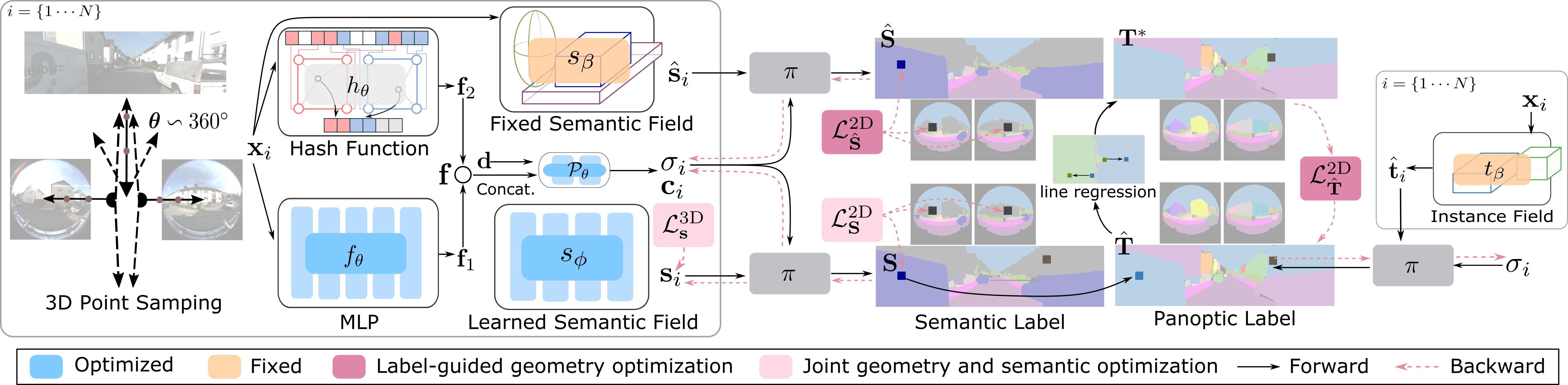}}
\caption{\textbf{Method Overview}. At each 3D location $\bx_i$, $i = \{1\cdots N\}$, we combine scene features obtained from a deep MLP $f_{\theta}$ and multi-resolution hash grids $h_{\theta}$ to jointly model geometry, appearance and semantics. We leverage dual semantic fields to obtain two semantic categorical logits: $\hat{\bs}_i$ is obtained from a fixed semantic field $s_{\beta}$ determined by the bounding primitives and ${\bs}_i$ is predicted by a learned semantic field $s_{\phi}$. The 3D semantic logits are accumulated along the ray and projected to 2D image space via volume rendering $\pi$, resulting in $\hat{\bS}$ and ${\bS}$. Our method allows for rendering panoptic labels $\hat{\bT}$ by combining the learned semantic field $s_{\phi}$ and a fixed instance field $t_{\beta}$ determined by the 3D bounding primitives. During training, we leverage label-guided geometry optimization. Here, $\cL_{\hat{\bS}}^{\text{2D}}$ and $\cL_{\hat{\bT}}^{\text{2D}}$ serve as semantic and instance guidance to improve volume density $\sigma_i$ guided by $\hat{\bS}$ and $\bT^{*}$, respectively, where $\bT^{*}$ is obtained by performing line regression on the building class in $\hat{\bT}$. Furthermore, we use $\cL_{\hat{\bs}}^{\text{3D}}$ and $\cL_{\bS}^{\text{2D}}$ to train the learned semantic field, yielding joint optimization of geometry and semantics. This allows for resolving label ambiguity at the intersections of the fixed semantic field. The final panoptic segmentation result of our method is denoted by $\hat{\bT}$.}
\label{fig:pipeline}
\end{figure*}

\subsection{Problem Formulation}
As shown in \figref{fig:teaser}, PanopticNeRF-360 aims to transfer coarse 3D bounding primitives to dense panoramic 2D semantic and instance labels, utilizing reconstructed geometry for 3D-2D transfer guidance. We follow the hardware and annotation setup of KITTI-360~\cite{liao2021kitti}: In addition to a pair of perspective stereo images and two-sided fisheye images sparsely collected at urban scenes, we assume a set of coarse 3D bounding primitives $\beta=\left\{B_{k}\right\}_{k=1}^{K}$ to be available. These 3D bounding primitives cover the full scene in the form of cuboids, ellipsoids and extruded polygons. Each 3D bounding primitive $B_{k}$ has a ``stuff'' or ``thing'' label. For ``thing'' classes (\eg, ``building'' and ``car''), $B_{k}$ is additionally associated with a unique instance ID.  We further apply a pre-trained semantic segmentation model to the RGB images to obtain 2D semantic predictions for each image. Given this input, our goal is to generate multi-view consistent panoptic labels and high-fidelity appearance for all input frames. Moreover, our method allows for rendering RGB images and panoptic labels from a wide range of novel viewpoints, even including $360^{\circ}$ omnidirectional views.

PanopticNeRF-360 provides a novel method for label transfer from 3D to 2D. \figref{fig:pipeline} provides an overview of our method. We first build a 360$^{\circ}$ scene representation (\secref{sec:360_scene_representation}) using both perspective and fisheye views, mapping a 3D point $\bx$ to a density $\sigma$ and a color value $\bc$ along with two semantic categorical logits $\hat{\bs}$ and $\bs$ based on our dual semantic fields. Here, $\hat{\bs}$ is a deterministic semantic logit derived from a \textit{fixed semantic field} $s_{\beta}$ defined by the 3D bounding primitives, and $\bs$ is learned semantic logit queried from a \textit{learned semantic field} $s_{\boldsymbol{\phi}}$. Accordingly, for each camera ray, two semantic categorical distributions $\hat{\bS}$ and $\bS$ in 2D image space are obtained via volume rendering $\pi$. Using the labeled 3D bounding primitives, we further define a deterministic \textit{fixed instance field} $t_\beta$ which divides ``thing'' classes in the fixed semantic field into distinct instances, allowing for rendering panoptic labels $\hat{\bT}$ when combined with the learned semantic field. The fixed semantic field $s_{\beta}$ and the fixed instance field $t_{\beta}$ together serve to improve the underlying scene geometry (\secref{sec:geometry_optimization}). Furthermore, by using semantic losses in 3D and 2D space (\secref{sec:semantic_optimization}), the learned semantic field $s_{\phi}$ results in improved semantics in overlapping regions based on refined underlying geometry.

\begin{figure}
\centerline{\includegraphics[width=.48\textwidth]{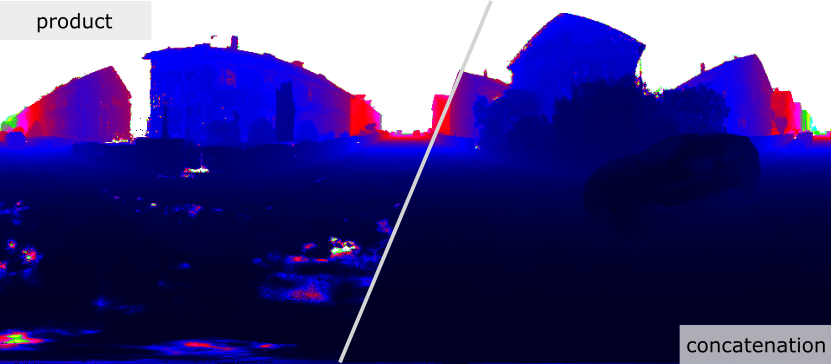}}
\caption{\textbf{Qualitative Geometric Comparison} of feature aggregation operators, ``concatenation'' and ``product''~\cite{chen2023factor}. The latter introduces jagged geometric errors on the road.} 
\label{fig:hybrid_concat_product}
\end{figure}

\subsection{360$^{\circ}$ Scene Representation} \label{sec:360_scene_representation}
\boldparagraph{Hybrid Scene Feature}
We seek to represent the 360$^{\circ}$ scene with both high-fidelity semantics and appearance. We empirically observe that the inductive bias of the MLP is a desirable property for smoothing noise in the semantic prediction, yet it is inferior in reconstructing high-fidelity urban scene appearance. On the other hand, existing local feature-based methods like Instant-NGP result in superior appearance in a short time but noisy semantics. Therefore, we fuse features of a deep MLP $f_{\boldsymbol{\theta}}$~\cite{mildenhall2020nerf} and multi-resolution hash grids $h_{\boldsymbol{\theta}}$~\cite{muller2022instant}  to form a joint representation. This simple representation allows us to combine the inductive smoothness bias of the MLP and the expressive representation power of local features, yielding high quality appearance and semantics jointly. Formally, a 3D coordinate $\bx$ is mapped to two feature vectors as follows:
\begin{equation}
f_{\boldsymbol{\theta}}: \gamma(\bx) \in\nR^{L_{\bx}} \mapsto \textbf{f}_{1}\in\nR^{D}, \quad h_{\boldsymbol{\theta}}: \bx\in\nR^3 \mapsto \textbf{f}_{2}\in\nR^{D} 
\end{equation}
where $\gamma(\cdot)$ denotes the positional encoding, $\textbf{f}_{1}$ and $\textbf{f}_{2}$ denote MLP-based and grid-based output features respectively. We obtain a unified scene feature $\textbf{f}$ via concatenation, $\textbf{f} = [\textbf{f}_{1}; \textbf{f}_{2}], \textbf{f}\in\nR^{2D}$. DiF~\cite{chen2023factor} introduces the inner product as a general form to unify different basis functions. However, we find that this element-wise ``product''  results in erroneous geometry, and we instead choose ``concatenation'' as shown in \figref{fig:hybrid_concat_product}. We hypothesize this is due to the fact that the MLP and grid branches operate less entangled during the gradient backpropagation, making it simpler for optimization. Along with a viewing direction $\bd$ and a frame-based trainable embedding $\bz$ for modeling varying exposure, we map $\textbf{f}$ to a volume density $\sigma$ and an RGB color value $\bc$ with a projection function $\mathcal{P}_{\theta}$ that consists of shallow MLPs:
\begin{equation}
\mathcal{P}_{\theta}: (\textbf{f}\in\nR^{2D}, \textbf{z}\in\nR^{n}, \gamma(\bd)\in\nR^{L_{\bd}}) \mapsto (\sigma\in\nR^+,\bc\in\nR^3)
\end{equation}
Let \mdi{$\mathbf{r}(k)=\mathbf{o}+ \mdi{k} \mathbf{d}$} denote a camera ray. The color at the corresponding pixel can be obtained by volume rendering~\cite{mildenhall2020nerf}
\begin{equation}
\mdi{\bC(\br)=\sum_{i=1}^{N} \mathcal{T}_i (1-\exp (-\sigma_i\delta_i)) \bc_i \ , \  \mathcal{T}_i = \exp}
\mdi{\left(-\sum_{j=1}^{i-1}  \sigma_j \delta_j \right)}
\label{eq:volume_render_c}
\end{equation}
where $\sigma_i$ and $\bc_i$ are density and color value at point $i$ sampled along the ray,  $\mdi{\mathcal{T}_i}$ denotes the transmittance at the sample point, and~\mdi{$\delta_{j}=k_{i+1}-k_{i}$} is the distance between adjacent samples. Let $\pi$ denote the volume rendering process of a ray $\br$, Eq.~\ref{eq:volume_render_c} can be rewritten as $\bC(\br|\boldsymbol{\theta})=\pi(\bc)$.

While the appearance embedding $\bz$ is shared across all points of the same frame during training, we use a different $\tilde{\bz}$ for each ray when rendering labels from omnidirectional novel viewpoints. This allows us to find the optimal $\bz$ for each ray and avoids stitching artifacts. Specifically, $\tilde{\bz}$ is determined based on its relative viewing direction $\bd$ with respect to the left and right fisheye cameras :
\begin{equation}
\tilde{\bz} = \alpha * \bz_{l}+ (1-\alpha) * \bz_{r}
\end{equation}
where $\alpha=\cos ^{-1}\left(\mathbf{d} \cdot \mathbf{d}_r\right) /\left(\cos ^{-1}\left(\mathbf{d} \cdot \mathbf{d}_l\right)+\cos ^{-1}\left(\mathbf{d} \cdot \mathbf{d}_r\right)\right)$ with $\bd_{l}$ and $\bd_{r}$ denoting unit viewing directions in the center of the left $\&$ right fisheye cameras, respectively, and $\bz_{l}$ and $\bz_{r}$ are corresponding latent codes.

\boldparagraph{Dual Semantic Fields}
To jointly optimize the underlying geometry and  semantics for mutual improvement, we define dual semantic fields, one is determined by the 3D bounding primitives $\beta$ and the other is learned by a semantic head $\boldsymbol{\phi}$
\begin{equation}
s_{\beta}: \bx\in\nR^3 \mapsto \hat{\bs}\in\nR^{M_s}, \quad~~ s_{\boldsymbol{\phi}}: \textbf{f}\in\nR^{2D}  \mapsto{\bs}\in\nR^{M_s}
\end{equation}
where $M_s$ denotes the number of semantic classes. 
In combination with the volume density, two semantic distributions $\hat{\bS}({\br})$ and  ${\bS}({\br})$ can be obtained at each camera ray $\br$ via accumulating the pre-softmax logits $\hat{\bs} (\bx)$ and $\bs(\bx)$ through the volume rendering operation $\pi$: %
\begin{equation}
\hat{\bS}(\br|\boldsymbol{\theta}, \beta)=  \pi (\hat{\bs} ), \quad {\bS}(\br|\boldsymbol{\theta}, \boldsymbol{\phi})=\pi ({\bs})
\label{eq:semantic_rendering}
\end{equation}

Note that both $\hat{\bS}(\br)$ and ${\bS}(\br)$ are multi-class normalized distributions through an extra softmax layer. We apply losses to both semantic distributions for training. During inference, the semantic label is determined as the class of maximum probability in $\hat{\bS}(\br)$ or ${\bS}(\br)$. 

\noindent \textit{Fixed Semantic Field}: If $\bx$ is uniquely enclosed by a 3D bounding primitive $B_k$, $\hat{\bs}$ is a fixed one-hot categorical logit vector of the category of $B_k$. For a point $\bx$ enclosed by multiple 3D bounding boxes of different semantic categories, we assign equal probability to all enclosed categories and $0$ to the others. As explained in \secref{sec:semantic_optimization}, the fixed semantic field $s_{\beta}$ is able to improve geometry but cannot resolve label ambiguity in overlapping regions.

\noindent \textit{Learned Semantic Field}: We add a semantic head parameterized by $\boldsymbol{\phi}$ to learn the semantic logits ${\bs (\bx)}$. Following~\cite{siddiqui2022panoptic}, we choose to perform softmax on 2D class logits after alpha compositing to obtain the class distribution ${\bS}(\br)$. We empirically observe that this leads to better performance than performing softmax on all 3D logits.

\boldparagraph{Fixed Instance Field}
Based on our learned semantic field $s_{\boldsymbol{\phi}}$ and the 3D bounding primitives $\beta$ with instance IDs, we can easily render a panoptic segmentation mask. Specifically, for a camera ray $\br$, the panoptic label directly takes the class with maximum probability in ${\bS}(\br)$ if it is a ``stuff'' class. For ``thing'' classes, we render an instance distribution $\hat{\bT}(\br)$ based on the bounding primitives $\beta$ to replace ${\bS}$ with $\hat{\bT}$.
Our instance field is defined as follow
\begin{equation}
t_{\beta}: \bx\in\nR^3 \mapsto \hat{\bt}\in\nR^{M_t}
\label{eq:instance field}
\end{equation}
where ${M_t}$ is the number of the things in the scene and $\hat{\bt}$ denotes categorical logits indicating which thing it belongs to. Here, $\hat{\bt}$ is determined by the bounding primitives and is a one-hot vector if $\bx$ is uniquely enclosed by a bounding primitive of a thing. In case $\bx$ is enclosed by multiple bounding primitives of different things, equal probabilities are assigned to each of them. To ensure that the instance label of this ray is consistent with the semantic class defined by ${\bS}$, we mask out instances belonging to other semantic classes by setting their probabilities to $0$ in $\hat{\bT}$.

\begin{figure}
\centerline{\includegraphics[width=.48\textwidth]{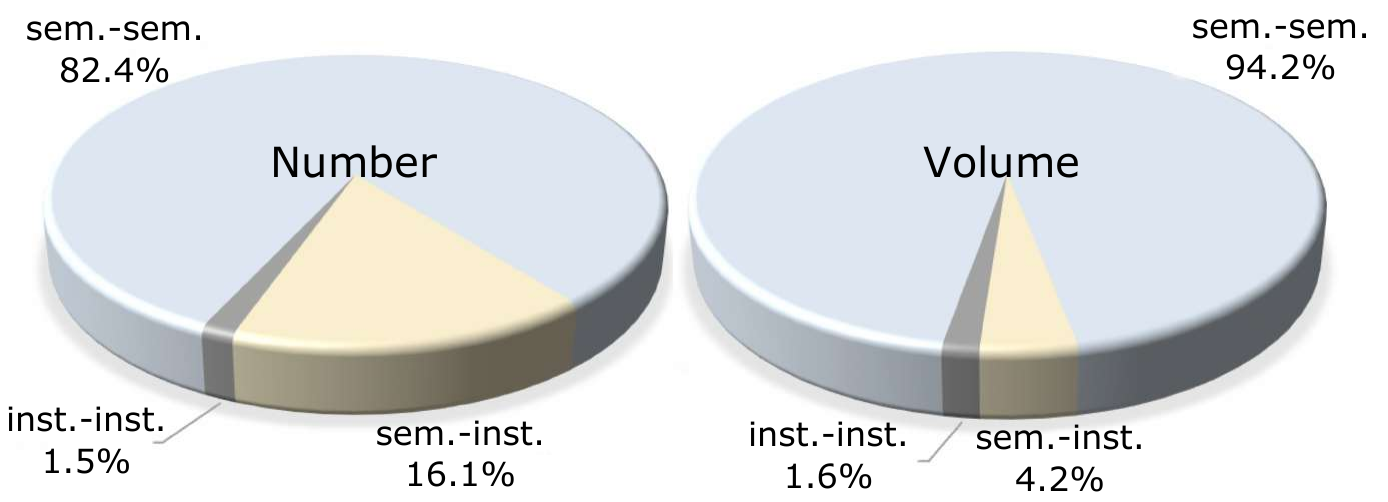}}
\caption{\textbf{Cross-semantics Bounding Primitve Intersection.} We calculate the number of intersected bounding primitives (left) and their corresponding intersected volume (right) in the entire driving sequence. Here, `sem.-sem.' denotes semantic-semantic intersection, `sem.-inst.' denotes semantic-instance intersection, and `inst.-inst.' denotes instance-instance intersection, respectively.}
\label{fig:bbox_sect}
\end{figure}

We observe that optimizing an additional learned instance field is not required. As shown in~\figref{fig:bbox_sect}, overlap often occurs at the intersection regions of stuff and thing regions, and the bounding primitives of things rarely overlap with each other~(the number of overlapping regions accounts for only 1.5$\%$ of the total number of pixels and their combined volume is only 1.6$\%$). Thus, the deterministic instance field can lead to reliable performance when the underlying geometry is correctly estimated.

\subsection{Label-Guided Geometry Optimization} \label{sec:geometry_optimization}
In the driving scenario considered in our setting, the RGB images are sparsely captured with many textureless and overexposed regions. We observe that a vanilla NeRF model fails to recover reliable geometry in this setting. Therefore, we propose to leverage a fixed semantic field and a fixed instance field to guide the optimization of scene geometry.

\begin{figure}[tb]
\centerline{\includegraphics[width=.5\textwidth]{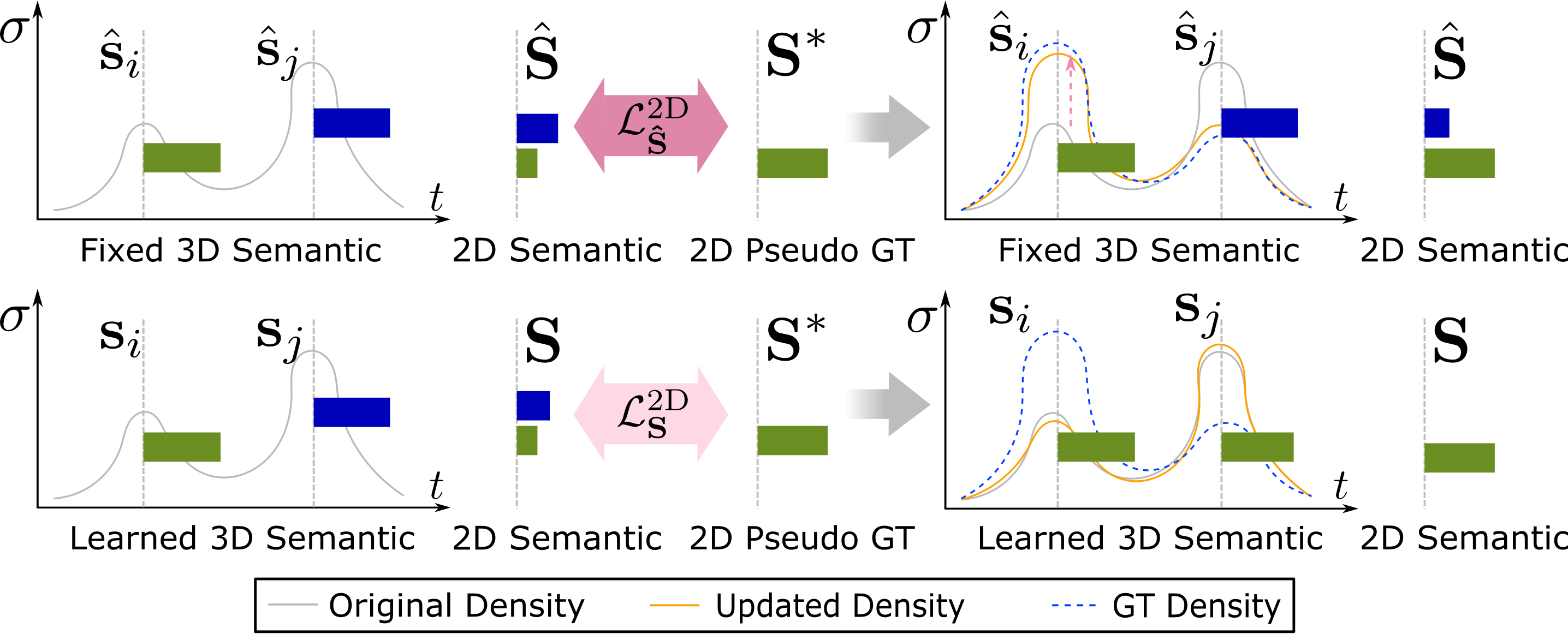}}
\caption{\textbf{Semantic Label Guided Geometry Optimization}. The top row illustrates a single ray of the fixed semantic field $s_\beta$, where $\cL^{\text{2D}}_{\hat{\bS}}$ can only update the underlying geometry as the semantic logits $\hat{\bs}$ is fixed. The second row shows a single ray of the learned semantic field $s_{\boldsymbol{\phi}}$. In this case, the network can ``cheat'' by adjusting the semantic prediction ${\bs}$ to satisfy $\cL^{\text{2D}}_{\bS}$ instead of updating the density $\sigma$.}
\label{fig:loss}
\end{figure}

\boldparagraph{Semantic Label-Guided Geometry Optimization} 
We find that leveraging noisy 2D semantic predictions as pseudo ground truth can substantially boost density prediction when applied to the fixed semantic fields $s_\beta$
\begin{equation}
\cL^{\text{2D}}_{\hat{\bS}}(\boldsymbol{\theta}, \beta)=-\frac{1}{|\mathcal{R}|}\sum_{\mathbf{r} \in \mathcal{R}}\sum_{k=1}^{M_{s}} \bS^*_{k}(\mathbf{r}) \log \hat{\bS}_{k}(\mathbf{r})
\label{eq:loss_baseline1}
\end{equation}
where $\hat{\bS}_{k}(\br)$ denotes the probability of the camera ray $\br$ belonging to the class $k$, and $\bS^*_k(\br)$ denotes the corresponding 2D pseudo ground truth. As illustrated in \figref{fig:loss}, the key to improving density is to directly apply the semantic loss to the fixed semantic field $s_{\beta}$, where $\cL^{\text{2D}}_{\hat{\bS}}$ can only be minimized by updating the density $\sigma$. 
\figref{fig:loss} shows that a correct $\bS^*$ increases the volume density of 3D points inside the respective bounding primitive and suppresses the density of others. When $\bS^*$ is wrong, the negative impact can be mitigated:
1) If $\bS^*$ does not match any bounding primitive along the ray, it has no impact on the radiance field $f_{\theta}$. 
2) If $\bS^*$ exists in one of the bounding primitives along the ray, this indicates that $\bS^*$ corresponds to an occluding/occluded bounding primitive with the wrong depth. To compensate for this, we introduce a weak depth loss $\cL_{d}$ based on stereo matching to alleviate the misguidance of $\cL^{\text{2D}}_{\hat{\bS}}$. Although  $\cL_{d}$ improves the overall geometry as shown in our ablation study, it fails to produce accurate object boundaries when used alone (see supplementary). In contrast, adding our semantic label-guided geometry optimization yields more accurate density estimation as pre-trained segmentation models usually perform well on frequently occurring classes, \eg, cars and roads. 

\boldparagraph{Instance Label-Guided Geometry Finetuning}
We further develop a simple but effective approach to improve the geometry of adjacent instances. We observe that the class building is the only ``thing'' class where two instances of the same semantic class are frequently spatially adjacent, which implies that their boundaries remain unchanged in the semantic label-guided geometry optimization. Further, the geometry of buildings is often negatively impacted by overexposure and low texturedness. Therefore, we propose to use pseudo-GT derived from a simple geometric prior to guide the density field optimization, assuming that neighboring buildings have a line-shaped boundary. Here, we do not use pre-trained segmentation models for two reasons: 1) existing pre-trained instance segmentation models are only applicable to a set of known classes, thus are not suited for unseen thing classes, e.g., buildings; 2) open-set segmentation models such as SAM~\cite{kirillov2023segment} lead to unsatisfactory boundaries in our challenging setting (see Supplementary). 

Let us consider a simple case where two neighboring buildings have a 2D boundary $l$. As buildings usually have cuboid-like shapes, $l$ can be approximated by \textit{a straight line}. However, the rendered boundary is typically noisy and non-straight when the density field predicts wrong geometry. Thus, we construct pseudo-GTs by applying line regression to rendered 2D boundaries and utilize the pseudo-GTs to optimize the density field. While the pseudo-GTs independently generated across different views may not be multi-view consistent, we fuse them in 3D space to render consistent labels. Specifically, we first render an initial instance boundary based on the fixed instance field and the learned imprecise density. As shown in~\figref{fig:ft_instance} (b), the naive boundaries can be jagged. Given the initial noisy boundary, we sample a set of points on the boundary of two adjacent buildings, ignoring the upper and lower regions which may contain unexpected irregular objects (\eg, eave, fence, vegetation) that violate the \textit{line} prior. Next, we apply linear regression on this set of sampled points to obtain a pseudo-GT of the boundary. Based on the regressed boundary formulation, we remap the instance labels on both sides to obtain a pseudo-GT panoptic label of two neighboring buildings. After refining all boundaries, we obtain pseudo panoptic label ${\bT}^{*}$:
\begin{equation}
{\bT}^{*} = f_{\chi} (\hat{\bT})
\end{equation}
where $f_{\chi}(\cdot)$ maps the initial instance label to a refined pseudo instance GT and $\chi$ denotes all the fitted lines between the buildings.

Then, we leverage ${\bT}^{*}$ to guide the underlying scene optimization. Similar to \eqref{eq:loss_baseline1}, we apply an instance loss based on the fixed instance field $t_{\beta}$ on ${\bT}^{*}$ to improve boundary geometry:
\begin{equation}
\cL^{\text{2D}}_{\hat{\bT}}(\boldsymbol{\theta}, \beta)=-\frac{1}{|\mathcal{R}_{k}|}\sum_{\mathbf{r} \in \mathcal{R}_{k} } \bT^*_{k}(\mathbf{r}) \log \hat{\bT}_{k}(\mathbf{r})
\label{eq:loss_fix_instance}
\end{equation}
where $k$ denotes the ``building'' class. The instance loss is weighted by $\lambda_{\hat{\bT}}$.

\subsection{Joint Geometry and Semantic Optimization} \label{sec:semantic_optimization}
While enabling improved geometry, the 3D label of overlapping regions remains ambiguous in the fixed semantic field. We leverage $s_{\boldsymbol{\phi}}$ to address this problem by jointly learning the semantic and the radiance fields. Towards this goal, we apply a modified cross-entropy loss $\cL^{\text{2D}}_{\bS}$ to each camera ray based on the filtered 2D pseudo ground truth, where $\mathbbm{1}(\br)$ is set to 1 if $\bS^*(\mathbf{r})$ matches the semantic class of any bounding primitive along the ray and otherwise $0$. 
Due to the imbalanced categorical distribution nature of pseudo semantic labels, we modify the softmax cross-entropy operator by introducing a weight $w(k)$ for each class k, which is distributed between [0,1] determined on the semantic class frequency~\cite{menon2020long}. We do not use $w(k)$ in $\cL^{\text{2D}}$ as we experimentally observe that the category distribution prior does not additionally help to improve geometry. To further suppress noise in the 2D predictions, we add a per-point semantic loss $\cL^{\text{3D}}_{\bs}$ based on the 3D bounding primitives
\begin{equation}
\begin{aligned}
\cL^{\text{2D}}_{\bS}(\boldsymbol{\theta},\boldsymbol{\phi}) &=-\frac{1}{|\mathcal{R}|}\sum_{\mathbf{r} \in \mathcal{R}} \mathbbm{1}(\br) \sum_{k=1}^{M_{s}} w(k) \bS^*_{k}(\mathbf{r}) \log {\bS}_{k}(\mathbf{r}) \\
\cL^{\text{3D}}_{\bs}(\boldsymbol{\theta}, \boldsymbol{\phi}, \beta) &= - \frac{1}{|\mathcal{R}|}\sum_{\mathbf{r} \in \mathcal{R}}\sum_{i=1}^{N} \mathbbm{1} (\bx) \sum_{k=1}^{M_s} \hat{\bs}_i^k \log {\bs}_i^k
\end{aligned}
\end{equation}
where $\mathbbm{1}(\bx)$ is a per-point binary mask. $\mathbbm{1}(\bx)$ is set to $1$ if (1) $\bx_i$ has a unique 3D semantic label and (2) the density $\sigma$ is above a threshold $\sigma_{th}$ to focus on the object surface.
As illustrated in \figref{fig:loss}, $\cL^{\text{2D}}_{\bS}(\boldsymbol{\theta},\boldsymbol{\phi})$ does not necessarily improve the underlying geometry as the network can learn a simple shortcut and adjust the semantic head $s_\phi$ to satisfy the loss. This behavior is also observed in novel view synthesis where NeRF does not necessarily recover good geometry when optimized for image reconstruction alone, specifically given sparse input views~\cite{Deng2021ARXIV,niemeyer2021regnerf}.

\begin{figure*}[!ht]
 \centering
 \newcommand{\mywidth}{.95\textwidth}
 \setlength\tabcolsep{0.05em}
 \newcolumntype{P}[1]{>{\centering\arraybackslash}m{#1}}
 \def\arraystretch{0.50}
  \begin{tabular}{P{0.5em}P{0.5em}P{\mywidth}}
    \rot{\scriptsize{Input}}&& \includegraphics[width=\mywidth]{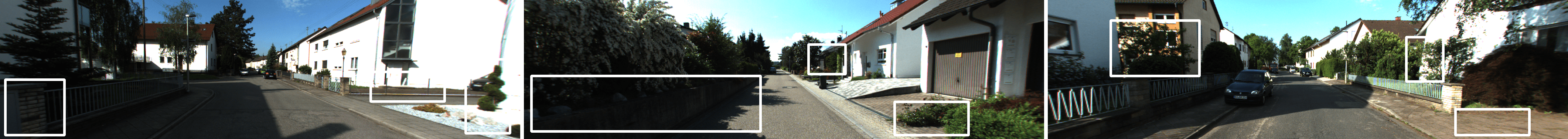} \\
     \rot{\scriptsize{J-NeRF}}&\rot{\tiny{~\cite{xu2023jacobinerf}}}&  \includegraphics[width=\mywidth]{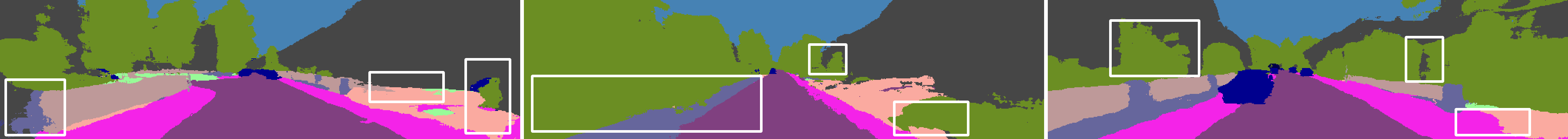}\\
     \rot{\scriptsize{PSPNet*}}&\rot{\tiny{~\cite{zhao2017pyramid}}}& \includegraphics[width=\mywidth]{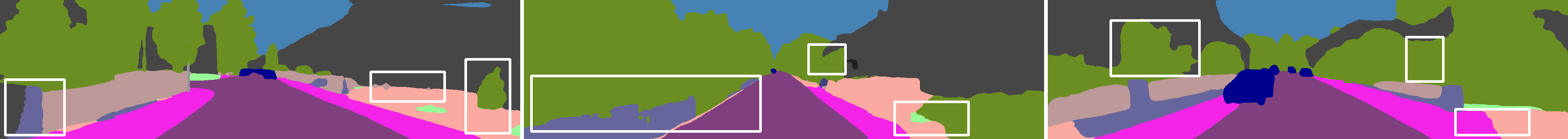}\\
     \md{\rot{\scriptsize{Panoptic Lifting}}}&\md{\rot{\tiny{~\cite{siddiqui2022panoptic}}}}&  \includegraphics[width=\mywidth]{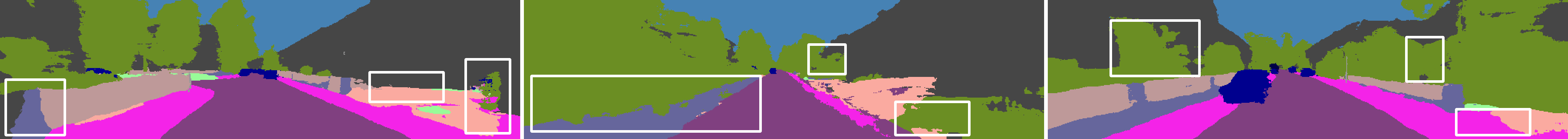}\\
     \rot{\scriptsize{3D-2D CRF}}&\rot{\tiny{~\cite{liao2021kitti}}}& \includegraphics[width=\mywidth]{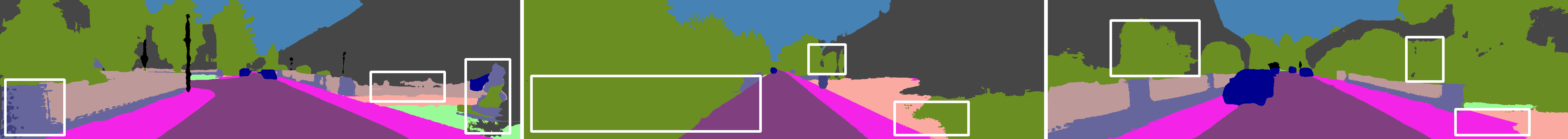} \\
    \rot{\scriptsize{Ours}}&& \includegraphics[width=\mywidth]{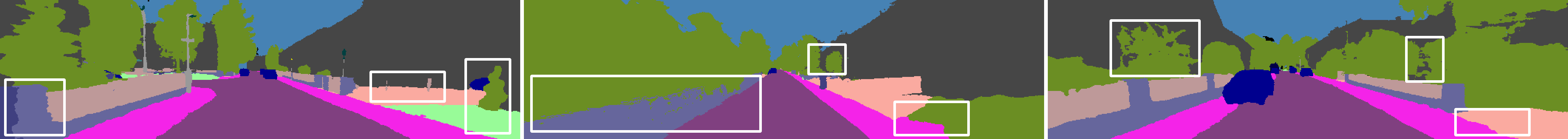}\\
   \rot{\scriptsize{GT}}&& \includegraphics[width=\mywidth]{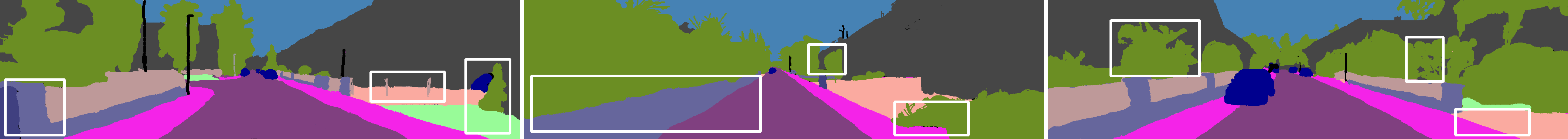}\\
 \end{tabular}
 \caption{
  \textbf{Qualitative Comparison of Perspective Semantic Label Transfer.}  Our method achieves superior performance in challenging regions compared to the baselines, e.g. in under- or over-exposed regions, by recovering the underlying 3D geometry, see wall (left, middle) and regions where vegetation and building intersect (right).}
 \label{fig:semantic_perspective}
\end{figure*}

\subsection{Implementation Details} 

\boldparagraph{Loss}
We train our model in two stages. In the first stage without instance finetuning, the total loss takes the following form
\begin{equation}
\cL =  \lambda_{\hat{\bS}} \cL^{\text{2D}}_{\hat{\bS}}  + \lambda_{{\bS}}\cL^{\text{2D}}_{\bS} + \lambda_{{\bs}}\cL^{\text{3D}}_{\bs}  + \lambda_{\bC}\cL_{p} + \lambda_{d}\cL_{d}
\label{eq:total loss}
\end{equation}
where $\cL_{p}=\frac{1}{|\mathcal{R}|}\sum_{\mathbf{r} \in \mathcal{R}}\left\|\bC^*(\br)-\bC(\br)\right\|_{2}^{2}$ and $\cL_{d}=\frac{1}{|\mathcal{R}|}\sum_{\mathbf{r} \in \mathcal{R}}\left\|\bD^*(\br)-\bD(\br)\right\|_{2}^{2}$ denote the photometric loss and the depth loss, respectively. $\lambda_{\hat{\bS}}$, $\lambda_{{\bS}}$, $\lambda_{{\bs}}$, $\lambda_{\bC}$, and $\lambda_{d}$ are constant weighting parameters. $\bC^*(\br)$ and $\bC(\br)$ are the ground truth and rendered RGB colors for ray $\br$. $\bD^*(\br)$ and $\bD(\br)$ are pseudo ground truth depth generated by stereo matching and rendered depth, respectively. Please refer to the supplementary for more details about $\bD^*$.

We activate instance label-guided geometry optimization in the second stage, leading to the overall loss $\cL_f$:
\begin{equation}
\cL_f = \cL + \lambda_{\hat{\bT}} \cL^{\text{2D}}_{\hat{\bT}}
\label{eq:loss_ft}
\end{equation}

As depicted in~\figref{fig:ft_instance}, the fine-tuning stage significantly improves the network's ability to produce smoother and more accurate geometry for adjacent buildings. 

\boldparagraph{Training}
We optimize one PanopticNeRF-360 model per scene, using a single NVIDIA RTX 3090. For each scene, we set the origin to the center of the scene. We use Adam~\cite{kingma2014adam} with a learning rate of 5e-4 to train our models. We set the latent appearance code length to $n=12$, and loss weights to $\lambda_{\hat{\bS}}=2, \lambda_{{\bS}}=1, \lambda_{{\bs}}=1, \lambda_{\hat{\bT}}=2, \lambda_{\bC}=1, \lambda_{d}=0.1$, and the density threshold to $\sigma_{th}=0.1$. We optimize the total loss $\cL$ for 30,000 iterations. For the stage of refining instances, we further fine-tune the model for 4,000 iterations.

\boldparagraph{Sampling Strategy and Sky Modeling} \label{samplingstrategy} With the 3D bounding primitives covering the full scene, we sample points inside the bounding primitives to skip empty space. For each ray, we optionally sample a set of points to model the sky after the furthest bounding primitive. More details regarding our sampling strategy can be found in the supplementary. Our sampling strategy allows the network to focus on non-empty regions. As evidenced by our experiments, this is particularly beneficial for unbounded outdoor environments.

\section{Experiments} \label{experiment}

\begin{figure*}[!ht]
 \centering
 \newcommand{\mywidth}{0.15\textwidth}
 \setlength\tabcolsep{0.2em}
 \begin{tabular}{cccccc}
  \includegraphics[width=\mywidth]{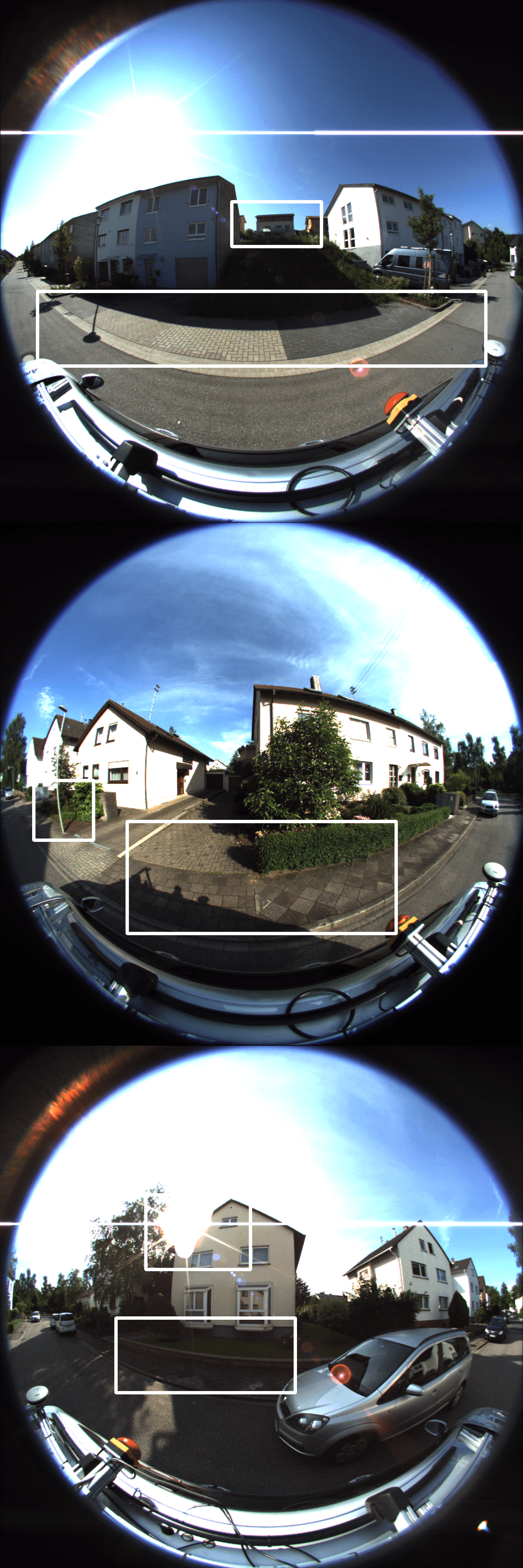}&
  \includegraphics[width=\mywidth]{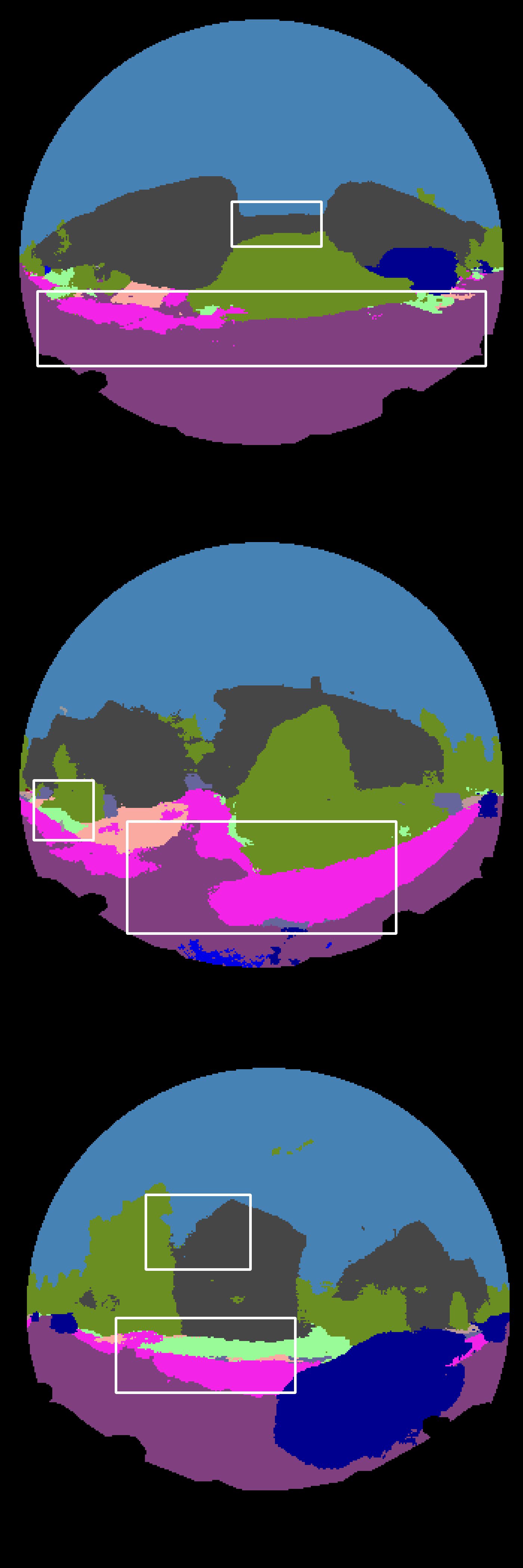} &
  \includegraphics[width=\mywidth]{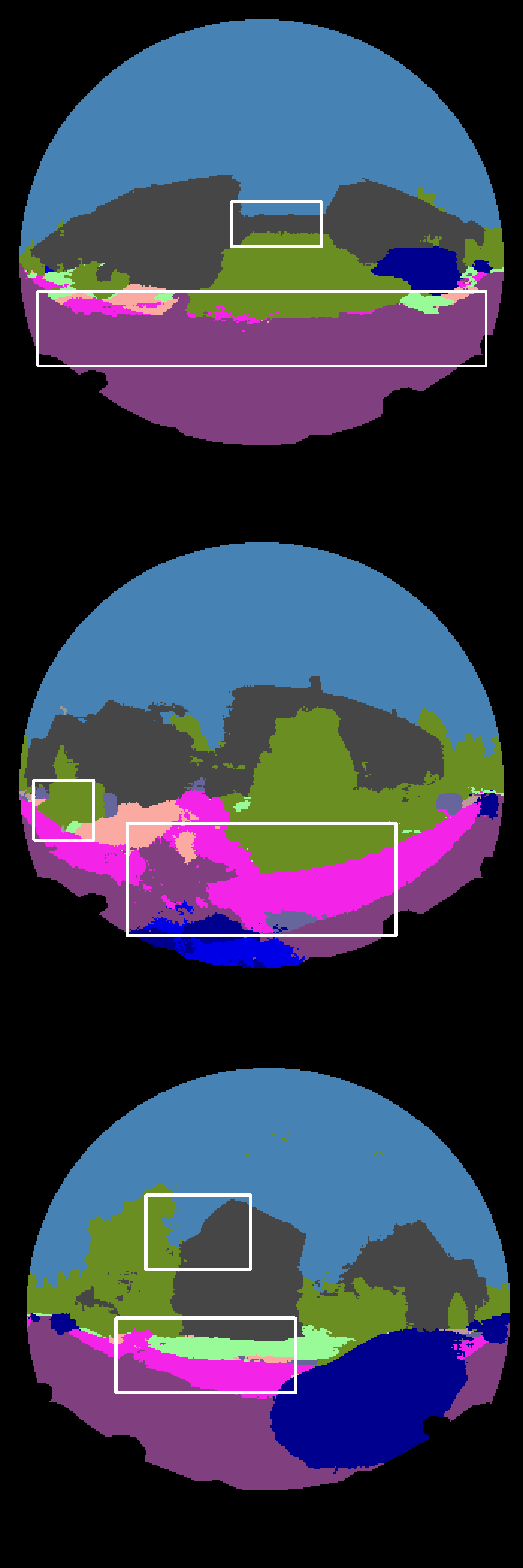} &
  \includegraphics[width=\mywidth]{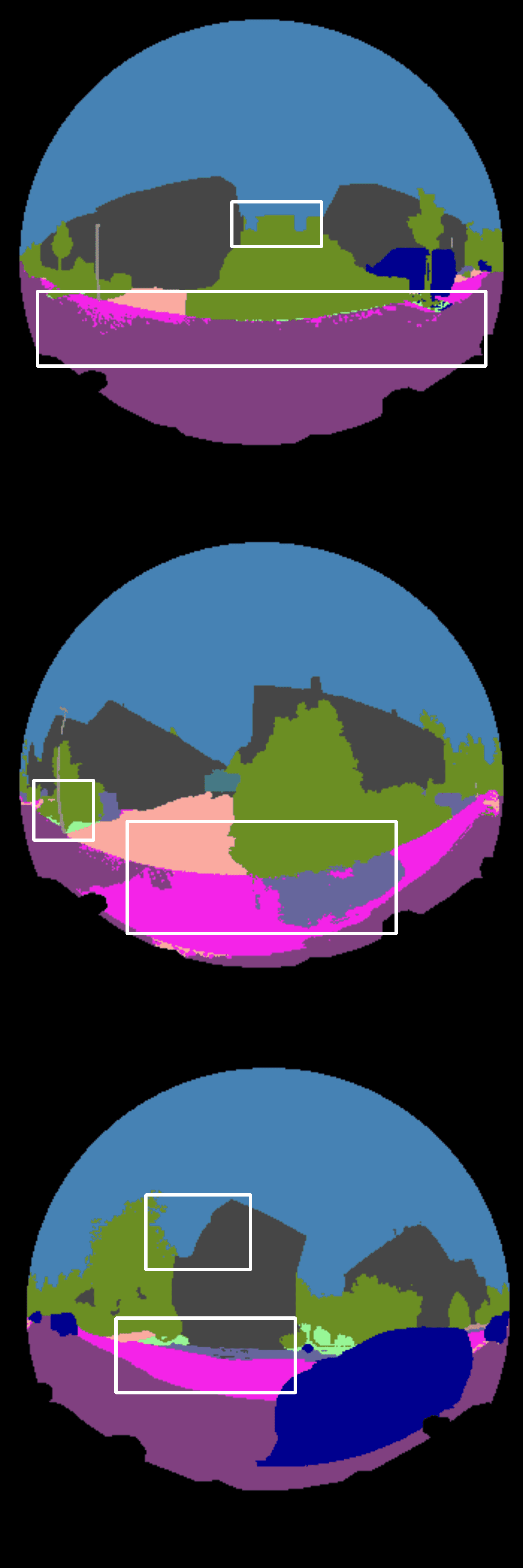} &
  \includegraphics[width=\mywidth]{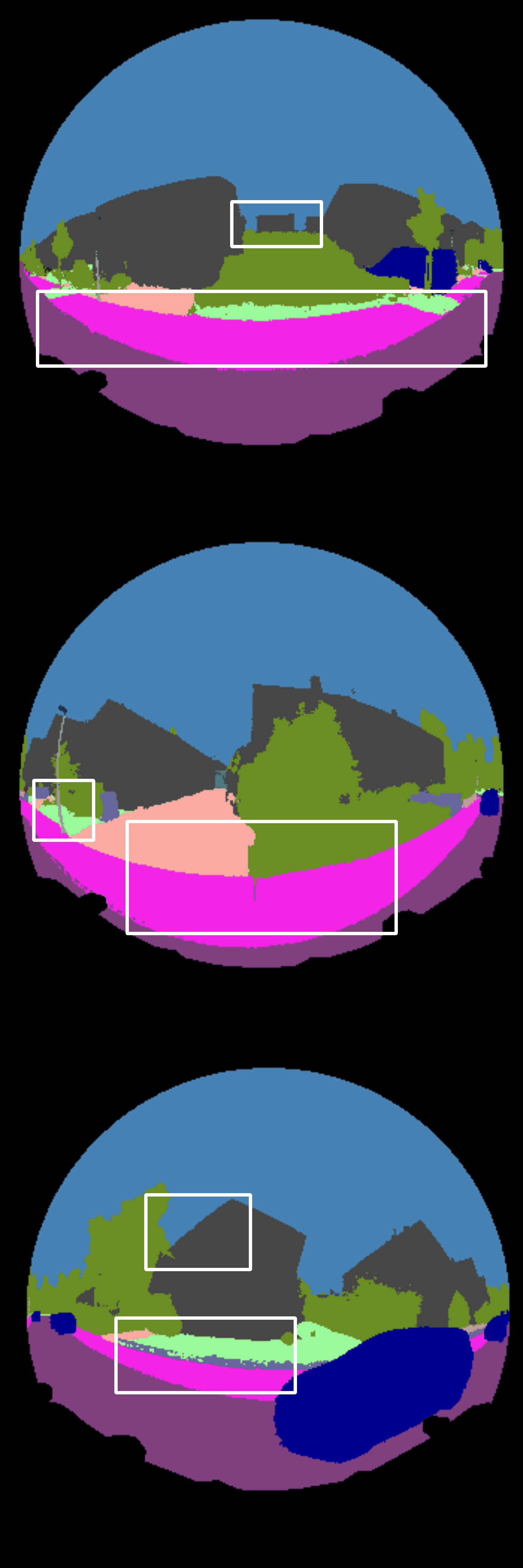}&
  \includegraphics[width=\mywidth]{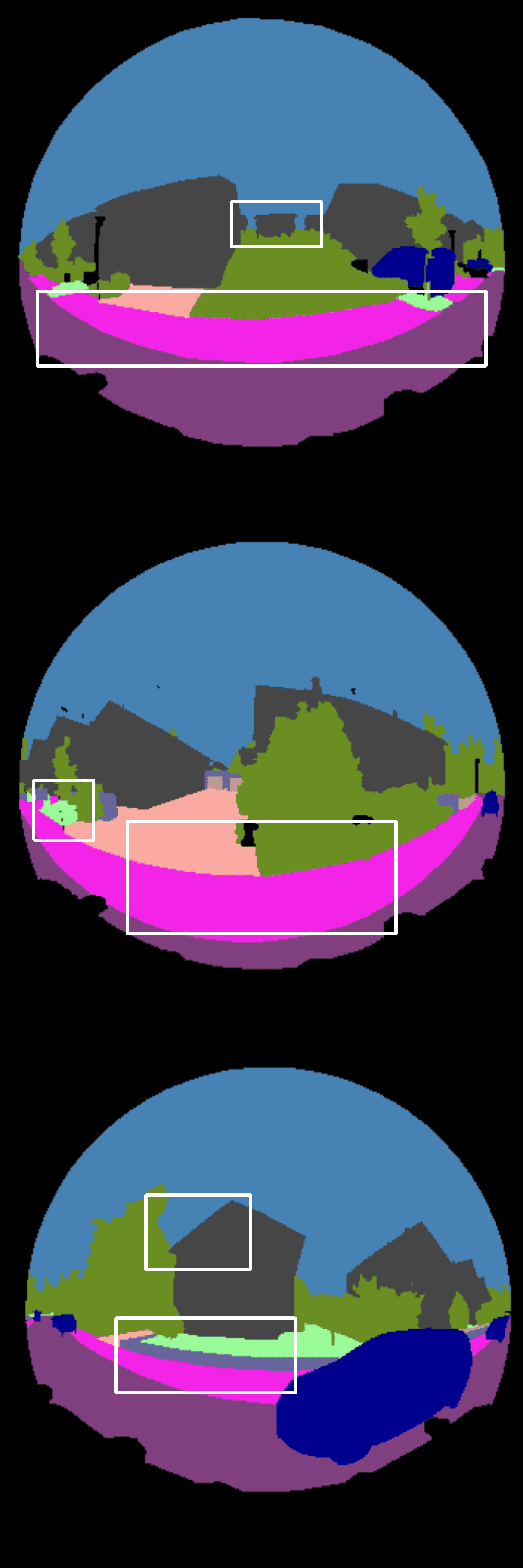}
    \\
    \small{Input} &  
    \small{J-NeRF~\cite{xu2023jacobinerf}} & 
    \small{\md{Panoptic Lifting~\cite{siddiqui2022panoptic}}} & 
    \small{3D-2D CRF~\cite{liao2021kitti}} &  
    \small{Ours} & 
    \small{GT Label}
 \end{tabular}
 \caption{
  \textbf{Qualitative Comparison of Fisheye Semantic Label Transfer.} Our method consistently achieves the best results. In the third row, ours can generate robust construction semantics under unexpected over-exposure, while the others can not.
 }
 \label{fig:semantic_fe}
\end{figure*}

\begin{table*}[!ht]
\centering
\resizebox{\textwidth}{!} {
\begin{tabular}{c||cccccccccccccc|c|c|c} \hline  
\toprule
Method &Road &Park &Sdwlk &Terr &Bldg &Vegt &Car &Trler &Crvn &Gate &Wall &Fence &Box &Sky &mIoU &Acc &MC \\  
\midrule
\multicolumn{18}{c}{Forward-facing} \\
\midrule
FC CRF + Manual GT~\cite{krahenbuhl2011efficient} & 90.3 & 49.9 & 67.7 & 62.5 & 88.3 & 79.2 & 85.6 & 48.9 & 78.1 & 23.4 & 35.3 & 46.5 & 42.0 & 92.7 & 63.6 & 89.1 & 85.41 \\ 
S-NeRF + Manual GT~\cite{zhi2021place} & 87.0 & 35.8 & 64.7 & 58.2 & 83.4 & 76.3 & 70.3 & 93.5 & 76.5 & 41.4 & 44.0 & 52.6 & 29.0 & 92.0 & 64.6 & 86.8 & 88.98 \\ 
Pseudo GT (PSPNet*)~\cite{zhao2017pyramid} & 95.5 & 49.7 & 77.5 & 66.7 & 88.9 & 82.4 & 91.6 & 46.5 & 83.1 & 24.2 & 43.3 & 51.3 & 51.1 & 89.3 & 67.2 & 90.7 & 91.79 \\ 
S-NeRF + Pseudo GT~\cite{zhi2021place} & 94.5 & 52.7 & 78.0 & 64.8 & 88.5 & 81.7 & 89.0 & 43.9 & 81.1 & 35.2 & 45.6 & 57.2 & 43.8 & 91.0 & 67.7 & 90.4 & 92.76 \\
J-NeRF + Pseudo GT~\cite{xu2023jacobinerf} & 94.8 & 49.7 & 80.2 & 65.3 & 86.2 & 83.1 & 91.0 & 45.4 & 80.6 & 32.5 & 48.1 & 58.1 & 52.3 & 91.2 & 68.5 & 90.8 & 92.88 \\
\md{Panoptic Lifting~\cite{siddiqui2022panoptic}} & \md{94.8} & \md{57.6} & \md{75.0} & \md{58.6} & \md{83.1} & \md{82.9} & \md{89.4} & \md{68.7} & \md{84.1} & \md{38.4} & \md{54.4} & \md{49.4} & \md{48.8} & \md{88.6} & \md{69.6} & \md{90.5} & \md{93.12}   \\
3D Primitives + GC & 81.7 & 31.0 & 45.6 & 22.5 & 59.6 & 56.7 & 63.0 & 61.7 & 37.3 & 61.6 & 28.8 & 50.6 & 39.5 & 50.3 & 49.3 & 73.4 & 86.56 \\ 
3D Mesh + GC & 91.7 & 53.1 & 67.2 & 31.4 & 81.3 & 72.1 & 85.2 & 93.5 & 86.0 & 65.2 & 40.7 & 59.7 & 54.4 & 65.6 & 67.7 & 86.0 & 94.99  \\ 
3D Point + GC & 93.5 & 59.0 & 76.1 & 37.2 & 82.0 & 74.1 & 87.5 & 94.7 & 85.7 & 66.7 & 59.4 & 65.9 & 58.6 & 68.0 & 72.0 & 87.9 & \textbf{96.51} \\ 
3D-2D CRF~\cite{liao2021kitti} & 95.2 & 64.2 & \textbf{83.8} & 67.9 & 90.3 & 84.2 & 92.2 & 93.4 & 90.8 & 68.2 & 64.5 & 70.0 & 55.8 & \textbf{92.8} & 79.5 & 92.8& 94.98 \\
Ours & \textbf{95.4} & \textbf{70.0} & 83.6 & \textbf{70.9} & \textbf{91.5} & \textbf{85.3} & \textbf{94.2} & \textbf{95.0} & \textbf{94.4} & \textbf{69.4} & \textbf{65.2} & \textbf{72.2} & \textbf{68.2} & 91.7 & \textbf{81.9} & \textbf{93.4} & 94.82\\ 
\midrule
\multicolumn{18}{c}{Fisheye} \\
\midrule
Pseudo GT (Tao~\textit{et al.})~\cite{tao2020hierarchical} & 84.8 & 0.0 & 56.2 & 57.5 & 89.5 & \textbf{84.5} & 77.7 & 0.0 & 0.0 & 0.0 & 39.3 & 60.0 & 0.0 & 98.0 & 46.3 & 91.0 & - \\
S-NeRF + Pseudo GT~\cite{zhi2021place} & 86.2 & 18.2 & 59.4 & 57.6 & 88.0 & 81.0 & 76.5 & 14.0 & 33.8 & 0.0 & 42.1 & 53.7 & 13.4 & 97.0 & 51.5 & 90.7 & - \\
J-NeRF + Pseudo GT~\cite{xu2023jacobinerf} & 86.3 & 17.2 & 61.2 & 58.4 & 87.1 & 81.9 & 78.1 & 17.3 & 30.6 & 0.0 & 44.3 & 55.3 & 15.2 & 96.8 & 52.1 & 91.1 & - \\
\md{Panoptic Lifting~\cite{siddiqui2022panoptic}} & \md{88.2} & \md{20.9} & \md{64.3} & \md{61.9} & \md{85.0} & \md{79.1} & \md{80.8} & \md{13.1} & \md{32.6} & \md{0.0} & \md{46.4} & \md{53.3} & \md{15.0} & \md{95.0} & \md{52.5} & \md{91.3} & \md{-} \\
3D-2D CRF~\cite{liao2021kitti} & 87.1 & 56.2 & 66,4 & 39.0 & 88.5 & 80.8 & 86.3 & 77.1 & 75.8 & 44.3 & 41.8 & 61.1 & 0.0 & 97.7 & 64.4 & 92.2 & - \\
Ours & \textbf{93.7} & \textbf{73.7} & \textbf{81.9} & \textbf{66.0} & \textbf{90.0} & 83.6 & \textbf{88.2} & \textbf{84.0} & \textbf{89.7} & \textbf{51.3} & \textbf{61.1} & \textbf{66.8} & \textbf{40.1} & \textbf{98.3} & \textbf{76.3} & \textbf{95.0} & - \\ 
\bottomrule
\end{tabular}
}
\caption{\textbf{Quantitative Comparison of Semantic Label Transfer} on 10 scenes of KITTI-360.}
\label{tab:semantic_perspective_fe}
\end{table*}

\begin{figure*}[!t]
 \centering
 \newcommand{\mywidth}{.95\textwidth}
 \setlength\tabcolsep{0.05em}
 \newcolumntype{P}[1]{>{\centering\arraybackslash}m{#1}}
 \def\arraystretch{0.50}
  \begin{tabular}{P{0.5em}P{0.5em}P{\mywidth}}
    \rot{\scriptsize{Input}}&& \includegraphics[width=\mywidth]{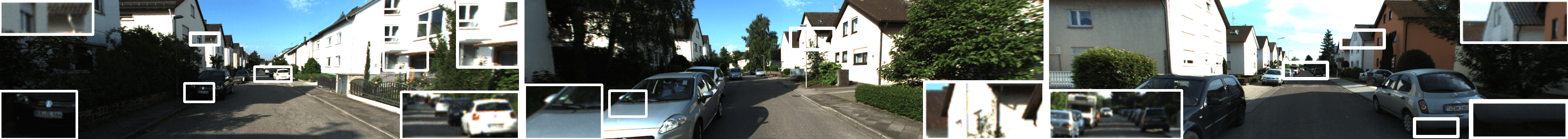} \\
     \md{\rot{\scriptsize{Panoptic Lifting}}}&\md{\rot{\tiny{~\cite{siddiqui2022panoptic}}}}&  \includegraphics[width=\mywidth]{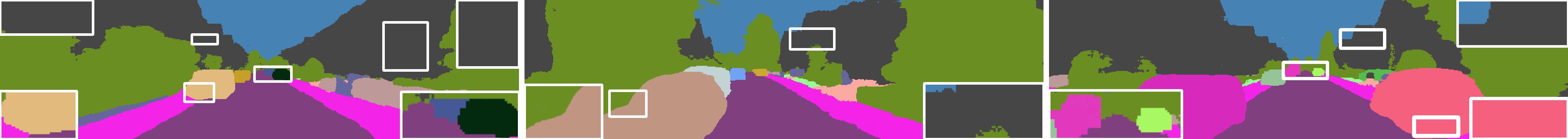}\\
     \rot{\scriptsize{3D-2D CRF}}&\rot{\tiny{~\cite{liao2021kitti}}}& \includegraphics[width=\mywidth]{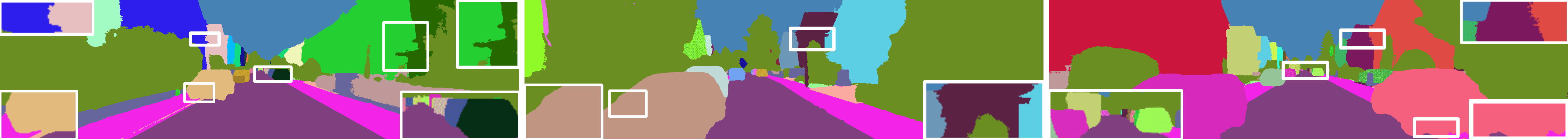} \\
    \rot{\scriptsize{Ours}}&& \includegraphics[width=\mywidth]{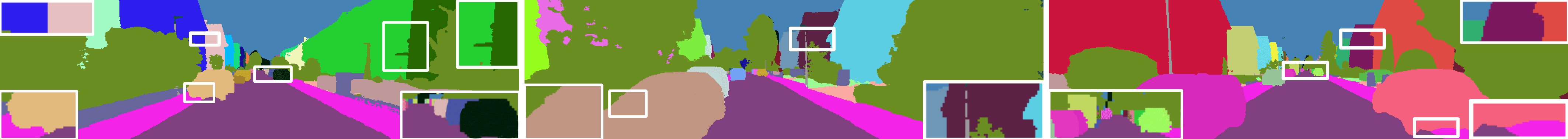}\\
   \rot{\scriptsize{GT}}&& \includegraphics[width=\mywidth]{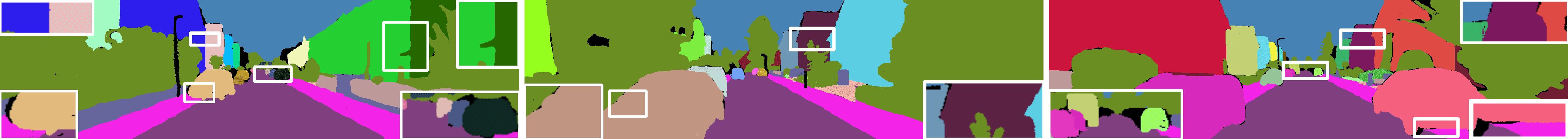}\\
 \end{tabular}
 \caption{
  \textbf{Qualitative Comparison of Perspective Panoptic Label Transfer.} Our method is capable of distinguishing instances as it infers in 3D space. In contrast, 3D-2D CRF struggles in far and overexposed regions. \md{Panoptic Lifting falls short in rendering 1) car instances in far regions and 2) building instances.}
 }
 \label{fig:instance_perspective}
\end{figure*}

\begin{figure*}[tb]
 \centering
 \newcommand{\mywidth}{0.194\textwidth}
 \setlength\tabcolsep{0.2em}
 \begin{tabular}{ccccc}
  \includegraphics[width=\mywidth]{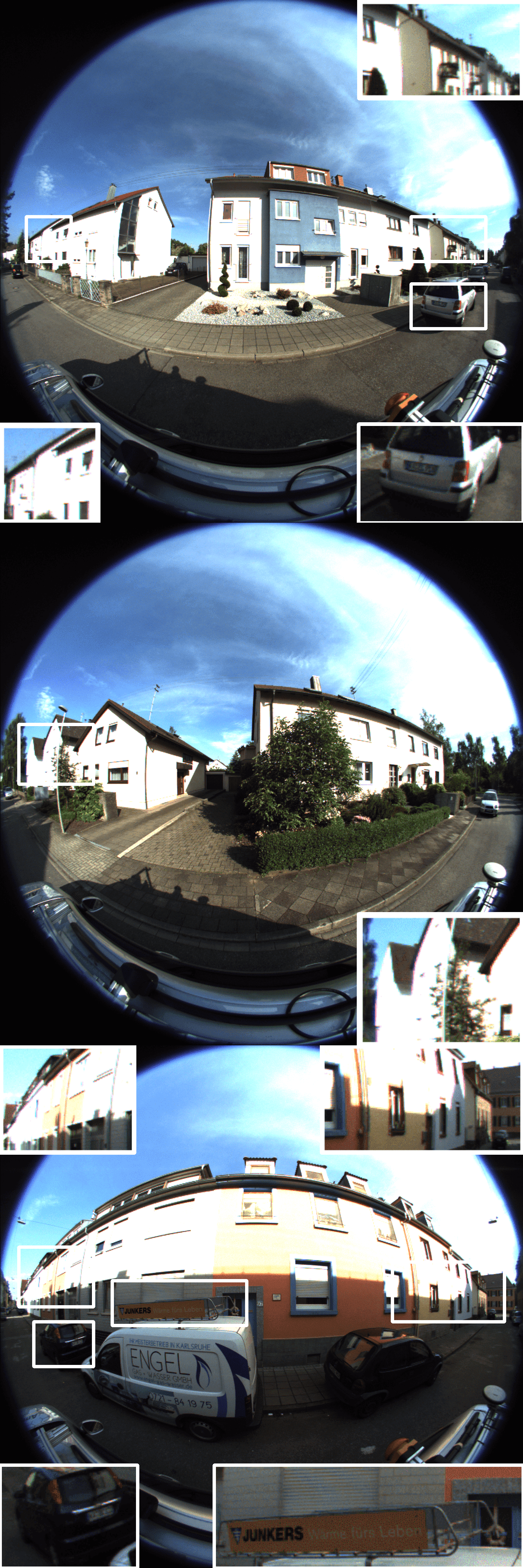}&
  \includegraphics[width=\mywidth]{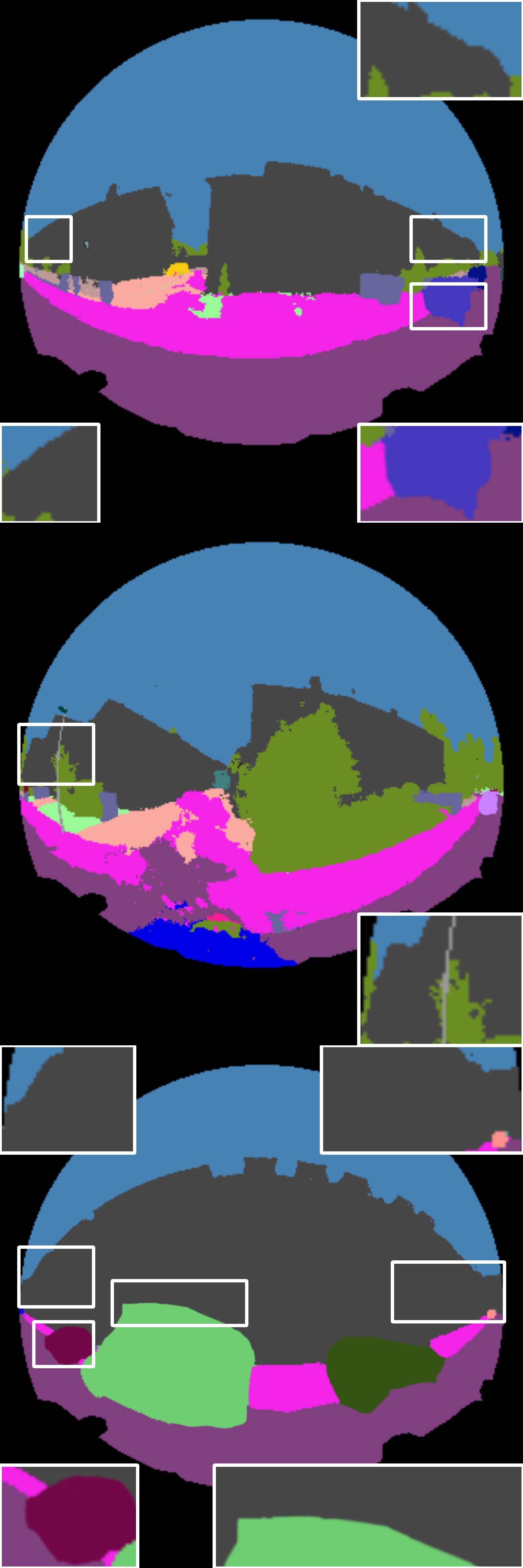} &
  \includegraphics[width=\mywidth]{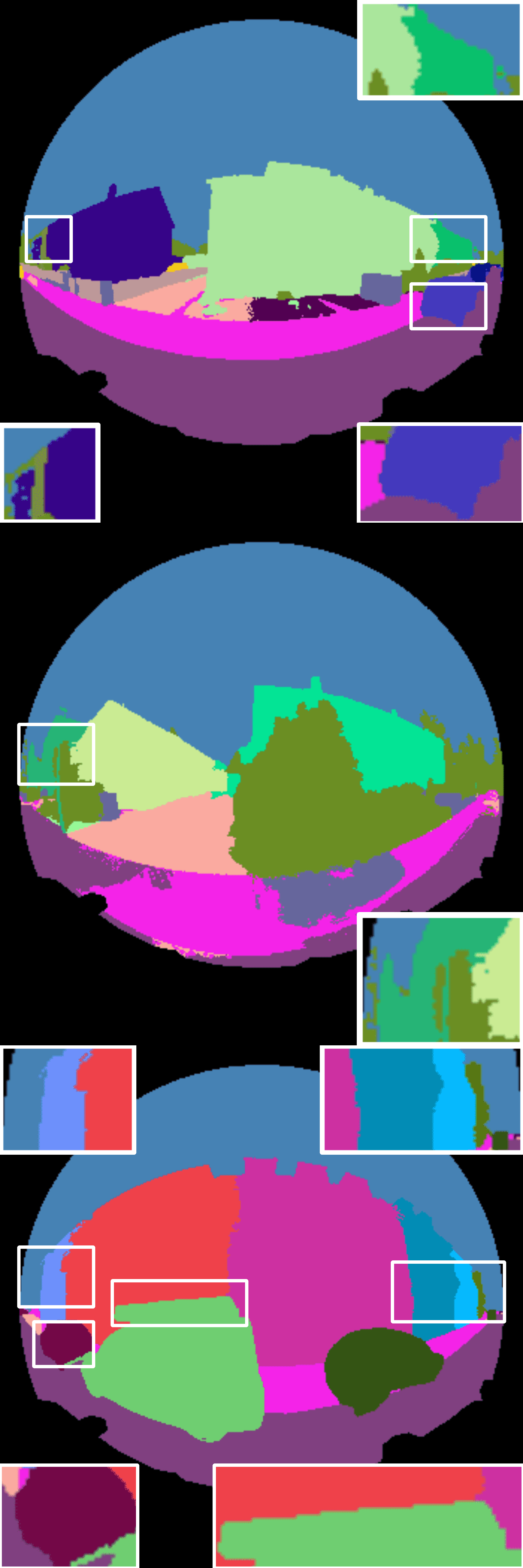} &
  \includegraphics[width=\mywidth]{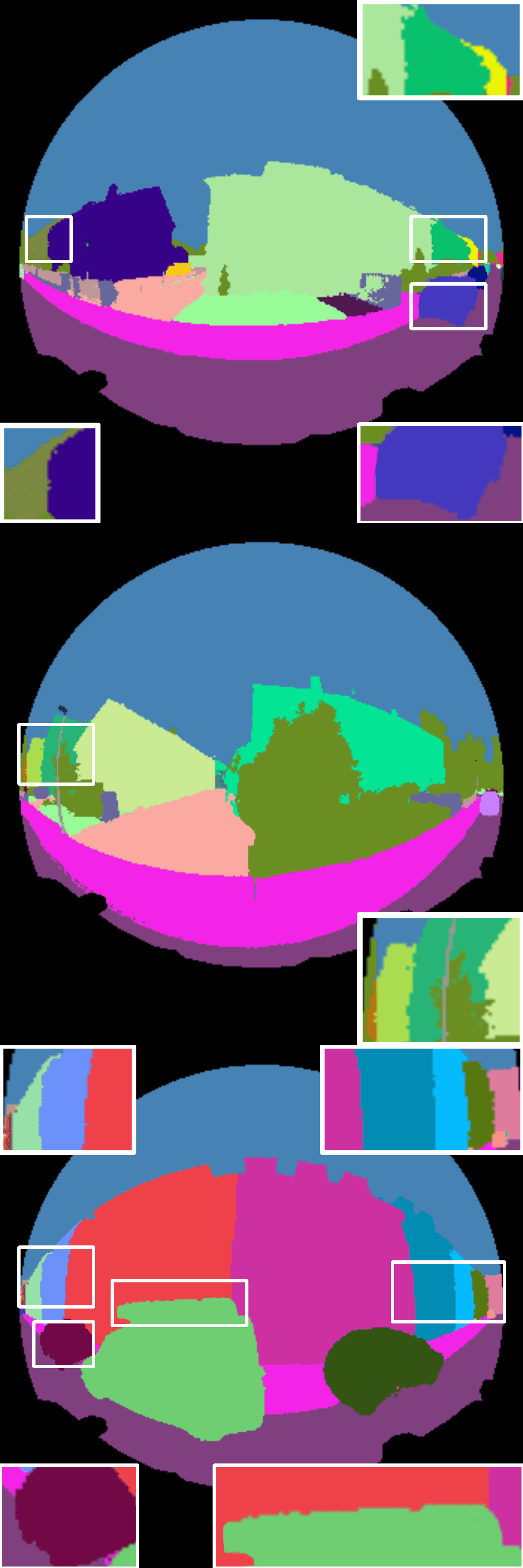}&
  \includegraphics[width=\mywidth]{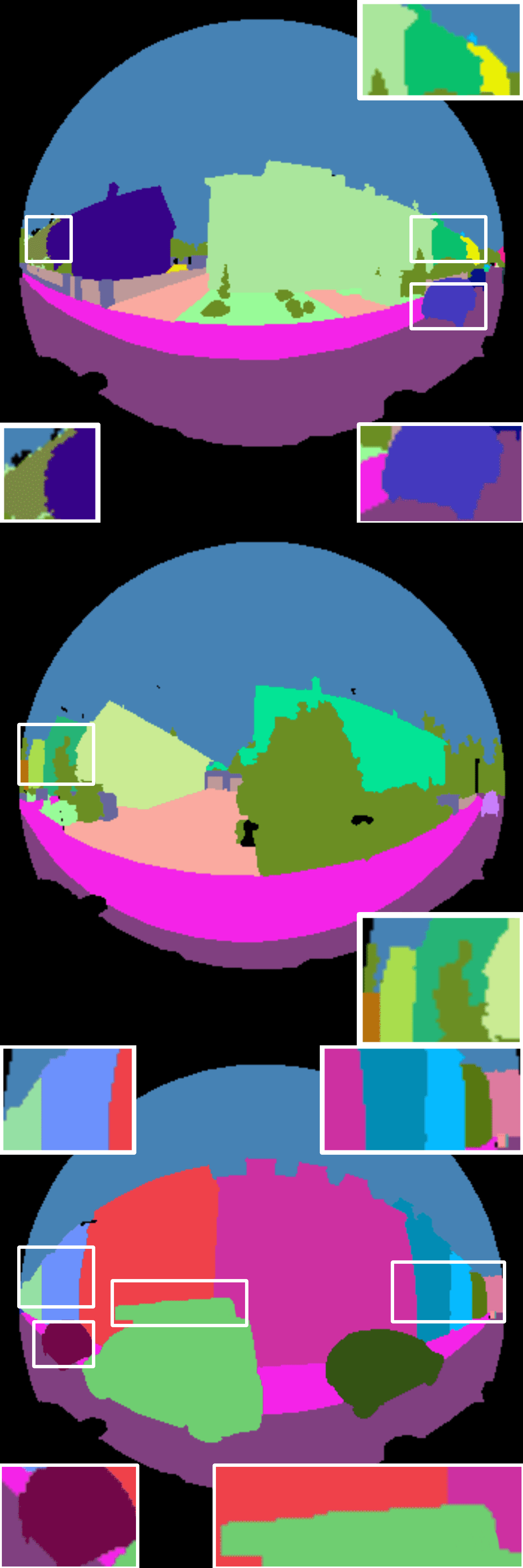}
    \\
    \small{Input} &  
    \small{Panoptic Lifting~\cite{siddiqui2022panoptic}} &  
    \small{3D-2D CRF~\cite{liao2021kitti}} &  
    \small{Ours} & 
    \small{GT Label}
 \end{tabular}
 \caption{
  \textbf{Qualitative Comparison of Fisheye Panoptic Label Transfer.} Our method outperforms 3D-2D CRF not only in near-field regions but also in far regions where instances are harder to distinguish. Compared to perspective views, the performance on panoptic labels of 3D-2D CRF and Panoptic Lifting declines more on fisheye views.
 }
 \label{fig:instance_fe}
\end{figure*}

\begin{table*}[h]
\centering
\resizebox{.8\textwidth}{!} {

\begin{tabular}{c|cc|cccc||c|cccc}
\toprule
& & \multicolumn{1}{c}{Forward-facing} & PQ & SQ & RQ & PQ† & \multicolumn{1}{c}{Fisheye} & PQ & SQ & RQ & PQ†\\

\midrule
{\multirow{6}{*}{\md{Building as stuff}}}  &
\multicolumn{1}{c|}{\multirow{3}{*}{\md{Panoptic Lifting~\cite{siddiqui2022panoptic}}}} & \md{All} & \md{52.3} & \md{69.4} & \md{68.2} & \md{54.9} & \md{All} & \md{42.6} & \md{68.5} & \md{56.6} & \md{44.7} \\
& \multicolumn{1}{c|}{}                           & \md{Things} & \md{56.5} & \md{76.3} & \md{70.4} & \md{57.8} & \md{Things} & \md{37.2} & \md{64.7} & \md{55.8} & \md{38.5} \\
& \multicolumn{1}{c|}{}                           & \md{Stuff}  & \md{53.9} & \md{68.9} & \md{75.3} & \md{55.7} & \md{Stuff} & \md{49.3} & \md{70.5} & \md{61.6} & \md{51.6} \\
\cmidrule{2-12}
& \multicolumn{1}{c|}{\multirow{3}{*}{\md{Ours*}}}      & \md{All} & \md{\textbf{66.7}} & \md{\textbf{80.6}} & \md{\textbf{81.7}} & \md{\textbf{68.4}} & \md{All} & \md{\textbf{54.2}} & \md{\textbf{77.6}} & \md{\textbf{67.6}} & \md{\textbf{56.5}} \\
& \multicolumn{1}{c|}{} & \md{Things} & \md{\textbf{65.3}} & \md{\textbf{83.6}} & \md{\textbf{77.2}} & \md{\textbf{65.3}} & \md{Things} & \md{\textbf{44.6}} & \md{\textbf{75.4}} & \md{\textbf{56.7}} & \md{\textbf{44.6}} \\
& \multicolumn{1}{c|}{} & \md{Stuff}  & \md{\textbf{67.8}} & \md{\textbf{79.6}} & \md{\textbf{84.0}} & \md{\textbf{69.3}} & \md{Stuff}  & \md{\textbf{58.9}} & \md{\textbf{78.2}} & \md{\textbf{72.8}} & \md{\textbf{61.9}} \\

\midrule
\midrule
{\multirow{6}{*}{\md{Building as thing}}}  & \multicolumn{1}{c|}{\multirow{3}{*}{3D-2D CRF~\cite{liao2021kitti}}} & All  & 62.2 & 79.1 & 76.9 & 64.9 & All & 47.5 & 72.2 & 60.1 & 50.9 \\
& \multicolumn{1}{c|}{} & Things & 60.7 & 79.5 & 75.2 & 60.7 & Things & 42.9 & 68.3 & 54.3 & 42.9 \\
& \multicolumn{1}{c|}{}                           & Stuff  & 63.0 & 78.9 & 77.9 & 67.3 & Stuff  & 50.1 & 74.6 & 63.5 & 55.3 \\
\cmidrule{2-12}
& \multicolumn{1}{c|}{\multirow{3}{*}{Ours}}      & All   & \textbf{66.7} & \textbf{80.6} & \textbf{81.7} & \textbf{69.3} & All  & \textbf{54.2} & \textbf{77.6} & \textbf{67.6} & \textbf{58.3} \\
& \multicolumn{1}{c|}{}                           & Things & \textbf{64.9} & \textbf{83.7} & \textbf{76.9} & \textbf{64.9} & Things & \textbf{46.5} & \textbf{77.2} & \textbf{59.3} & \textbf{46.5} \\
& \multicolumn{1}{c|}{}                           & Stuff  & \textbf{67.7} & \textbf{78.9} & \textbf{84.3} & \textbf{71.8} & Stuff  & \textbf{58.5} & \textbf{77.8} & \textbf{72.2} & \textbf{64.9} \\

\bottomrule
\end{tabular}

}
\caption{\textbf{Quantitative Comparison of Panoptic Label} over all 10 test scenes on KITTI-360. We consider building as \textit{stuff} when comparing our method (Ours*) to Panoptic Lifting which can not render construction instances, where building is considered as \textit{thing} in our default setting.}

\label{tab:instance_perspective_fe}
\end{table*}

\begin{figure*}[tb]
 \centering
 \newcommand{\mywidth}{0.48\textwidth}
 \setlength\tabcolsep{0.2em}
 \begin{tabular}{cccc}
   \includegraphics[width=\mywidth]{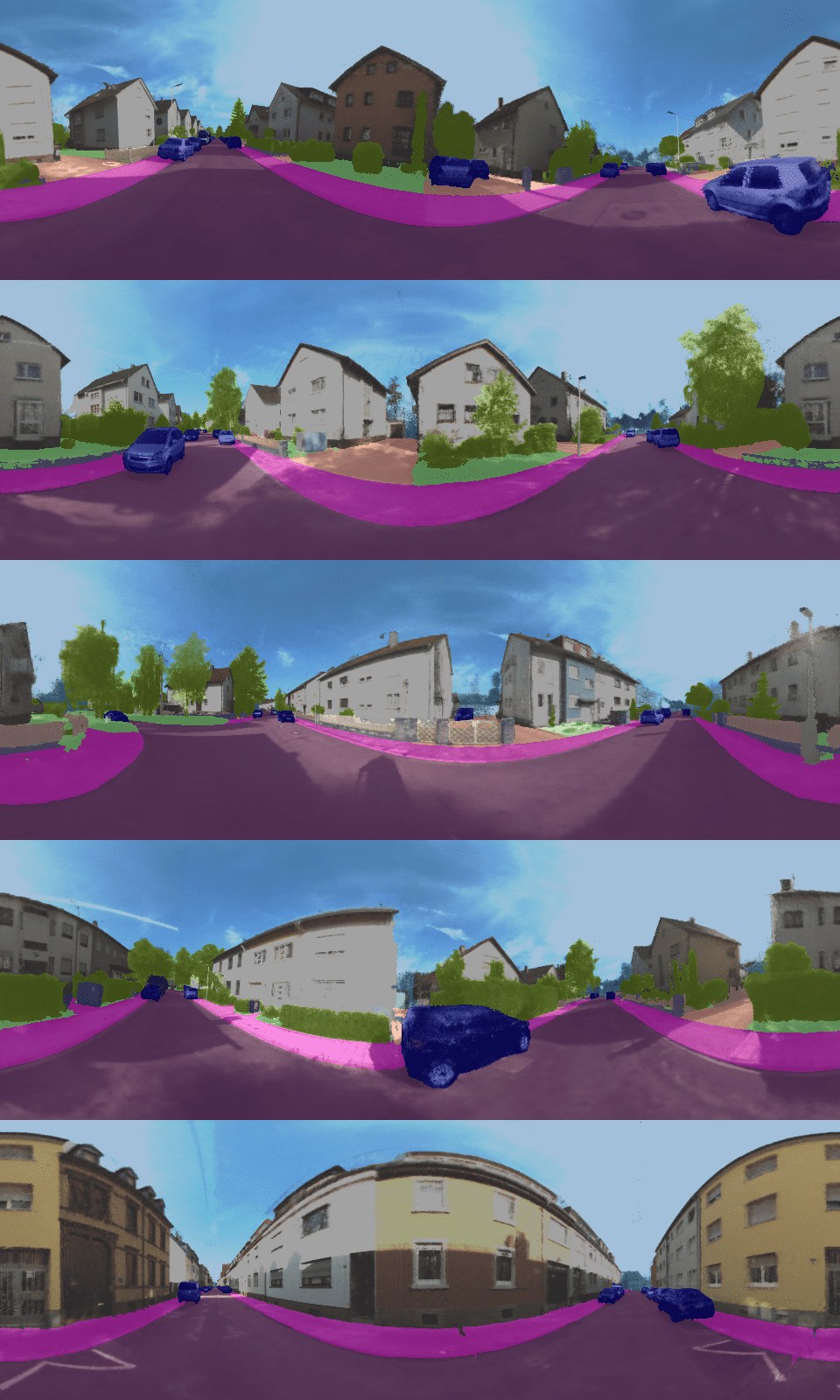}&
  \includegraphics[width=\mywidth]{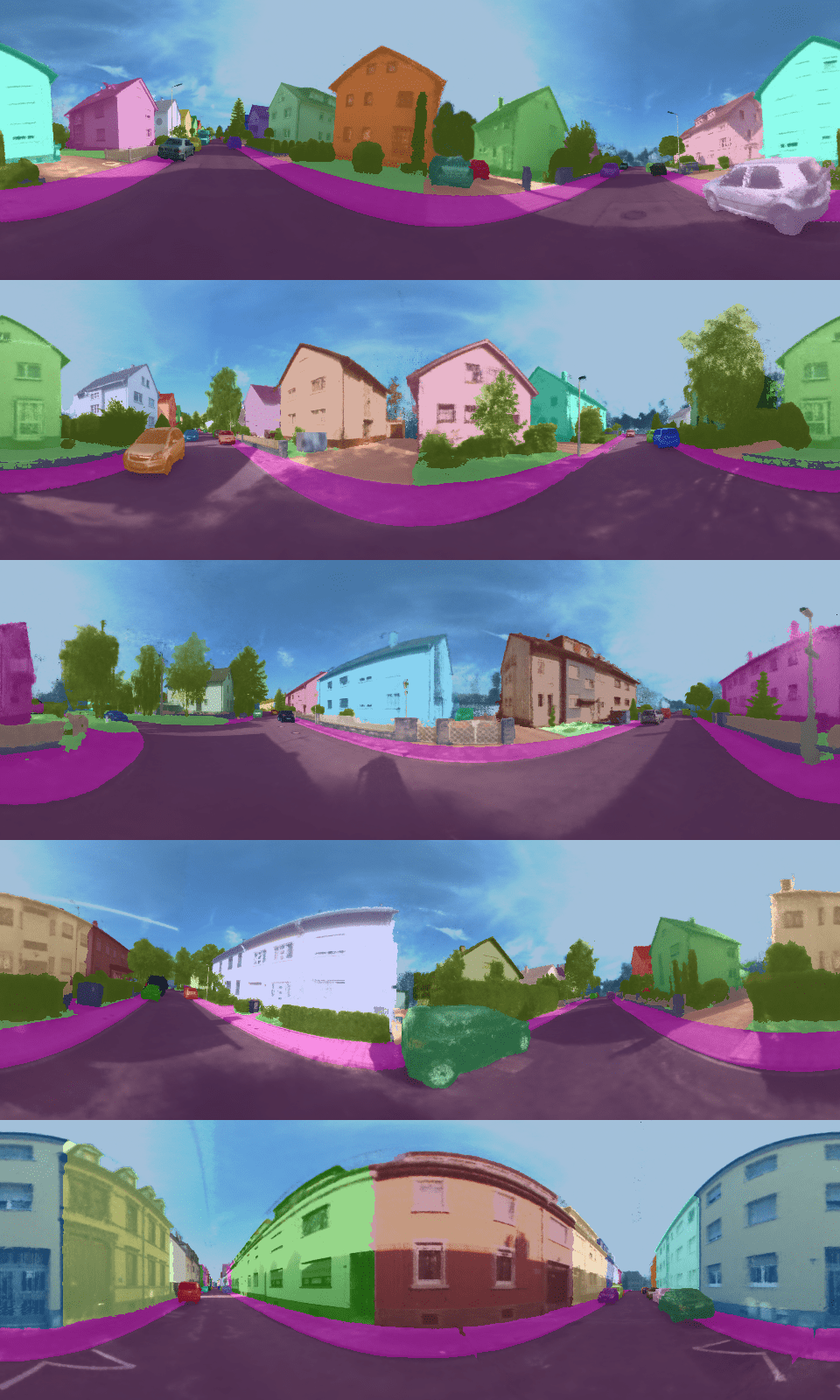}
  \\
     \small{Panoramic Semantic Maps}&
     \small{Panoramic Panoptic Maps} 
 \end{tabular}
 \vspace{-0.3cm}
 \caption{
  \textbf{Panoramic Label Rendering.} We enable omnidirectional panoptic label inference in real-world driving scenarios.
 }
 \label{fig:more_panorama}
 \vspace{-2em}
\end{figure*}

\begin{figure*}[!ht]
 \centering
\includegraphics[width=0.95\textwidth]{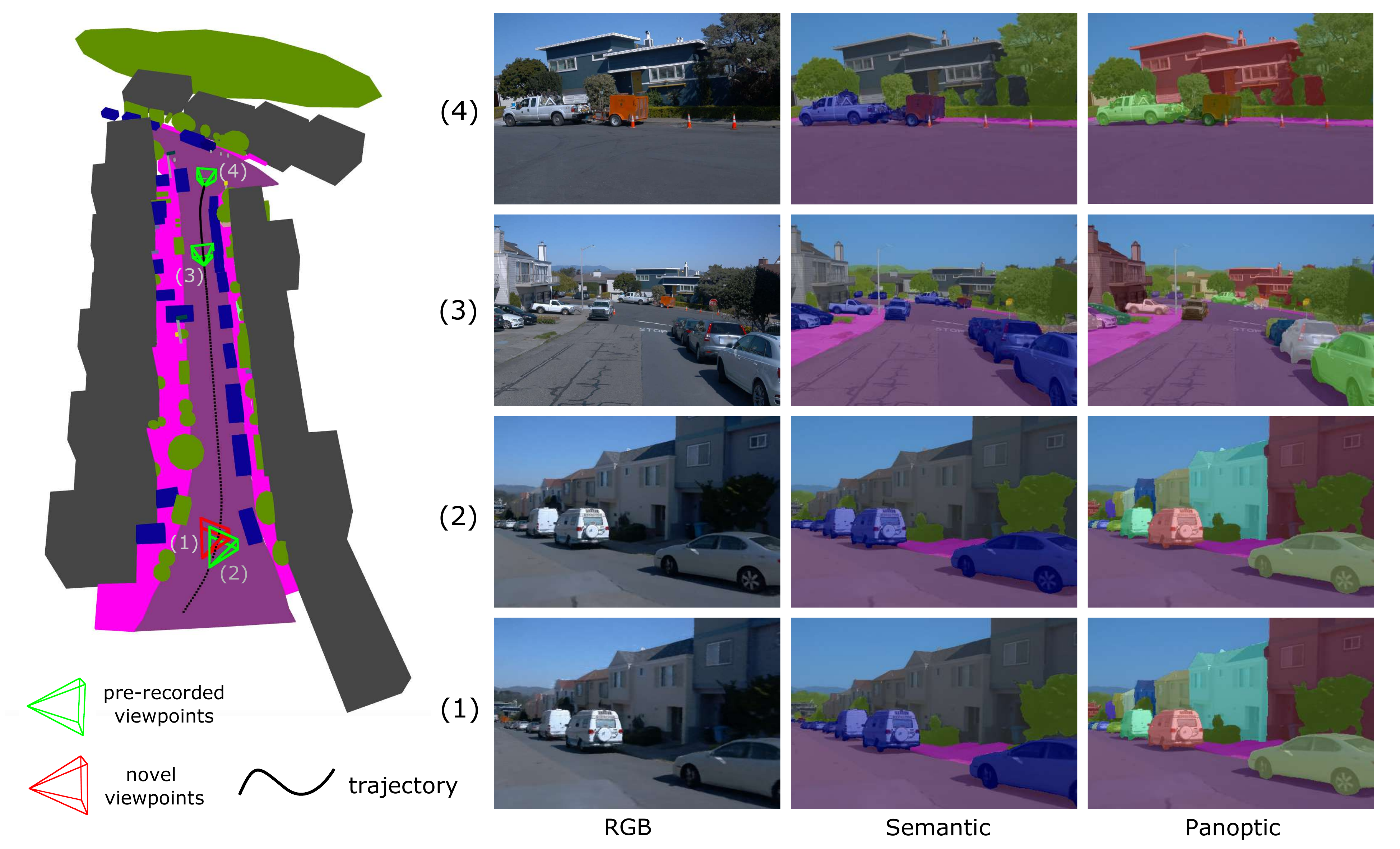}
 \caption{
 \textbf{Qualitative Label Transfer Results on Waymo.} Our method can render high-quality panoptic label maps as well as high-frequency appearance on the Waymo dataset.}
 \label{fig:waymo_panopticnerf}
\end{figure*}

\boldparagraph{Dataset} We conduct our main experiments on the KITTI-360~\cite{liao2021kitti} dataset, which is collected in suburban areas and provides 3D bounding primitives covering the full scene. Following~\cite{liao2021kitti}, we evaluate PanopticNeRF-360 on manually annotated frames from 5 static suburbs. We split these 5 suburbs into 10 scenes, each comprising 64 consecutive frames with 4 cameras each with an average travel distance of 0.8m between frames. We leverage these 64 pairs of posed stereo perspective and fisheye images for training. KITTI-360 provides a set of manually labeled frames sampled in equidistant steps of 5 frames on perspective views. Following~\cite{liao2021kitti}, we use half of the manually labeled frames for evaluation and provide the other half as input to 2D-to-2D label transfer baselines. We further improve the quality of the manually labeled ground truth which is inaccurate in ambiguous regions, see supplementary for details. In order to quantitatively evaluate our label transfer performance on side-facing viewpoints, we manually annotate fisheye images sampled in equidistant steps of 10 frames and use all of them for evaluation. We additionally showcase the generalization ability of our method on the Waymo dataset.

\boldparagraph{Baselines} We compare against several competitive baselines in two categories: (1) \textit{2D-to-2D label transfer baselines}, including Fully Connected CRF (\textit{FC CRF})~\cite{krahenbuhl2011efficient}, Semantic NeRF (\textit{S-NeRF})~\cite{zhi2021place}, JacobiNeRF (\textit{J-NeRF})~\cite{xu2023jacobinerf}, and Panoptic Lifting~\cite{siddiqui2022panoptic}. We provide manually annotated 2D frames as input, sparsely sampled at equidistant steps of 10 frames. Note that labeling these 2D frames takes similar or longer compared to annotating 3D bounding primitives~\cite{xie2016semantic}. As these 2D annotations are extremely sparse, we further provide the same pseudo-2D labels used by our method to Semantic NeRF, JacobiNeRF and Panoptic Lifting. For Panoptic Lifting, We additionally run
Mask2Former~\cite{cheng2022masked} to generate instance masks, and then fuse them with 2D semantic maps to obtain the final input panoptic masks. For JacobiNeRF, we follow~\cite{xu2023jacobinerf} to extract DINO features from the images for its similarity prior. (2) \textit{3D-to-2D label transfer baselines}, including PSPNet*,  3D Primitives/Meshs/Points+GC~\cite{liao2021kitti}, and 3D-2D CRF~\cite{liao2021kitti}. All these baselines leverage the same 3D bounding primitives to transfer labels to 2D. Here, PSPNet* is considered 3D-to-2D as it is pre-trained on Cityscapes and fine-tuned on KITTI-360 based on the 3D sparse label projections. The second set of baselines first project 3D primitives/meshes/points to 2D and then apply Graph Cut to densify the label. The 3D-2D CRF densely connects 2D image pixels and 3D LiDAR points, performing inference jointly on these two fields with a set of consistency constraints. 

\boldparagraph{Pseudo 2D GT} For pseudo ground truth, we use PSPNet* for perspective views and Tao \etal~\cite{tao2020hierarchical} that has a stronger generalization ability for fisheye views to supervise our dual semantic fields. This ensures a fair comparison to the 3D-2D CRF, which takes the predictions of PSPNet* and Tao \etal as unary terms on perspective views and fisheye views, respectively. Note that PSPNet* is fine-tuned on KITTI-360. In our ablation study we investigate the performance of our method using pre-trained models on Cityscape without any fine-tuning, including PSPNet~\cite{zhao2017pyramid}, Deeplab~\cite{chen2017deeplab} and Tao \textit{et al.}~\cite{tao2020hierarchical}. We also involve SSA~\cite{chen2023semantic}, a variant of SAM for zero-shot semantic prediction.

\boldparagraph{Metrics} Following ~\cite{fu2022panoptic}, we evaluate semantic labels via the mean Intersection over Union (mIoU) and the average pixel accuracy (Acc) metrics. To quantitatively evaluate multi-view consistency (MC), we utilize LiDAR points to retrieve corresponding pixel pairs between two consecutive evaluation frames. The MC metric is then calculated as the ratio of pixel pairs with consistent semantic labels over all pairs. For evaluating panoptic segmentation, we report Panoptic Quality (PQ)~\cite{kirillov2019panoptic}, which can be decomposed into Segmentation Quality (SQ) and Recognition Quality (RQ). We additionally adopt PQ†~\cite{porzi2019seamless} as PQ over-penalizes errors of stuff classes. To evaluate perspective and fisheye labels in a unified manner, we design \textit{${metric}^{\ast}$} (\eg, mIoU* and PQ*) that reports the average of the perspective metric and fisheye metric. To verify that PanopticNeRF-360 is able to improve the underlying geometry, we further evaluate on perspective views the rendered depth compared to sparse depth maps obtained from LiDAR using Root Mean Squared Error (RMSE) and the ratio of accurate predictions ($\delta_{1.25}$)~\cite{bhoi2019monocular,fu2018deep}.

\subsection{Label Transfer~\mdi{on KITTI-360 Dataset}}
As most baselines are not designed for panoptic label transfer, we first compare the semantic predictions, and then compare the panoptic predictions to 3D-2D CRF.

\boldparagraph{Semantic Label Transfer} 
As shown in \tabref{tab:semantic_perspective_fe}, \figref{fig:semantic_perspective} and \figref{fig:semantic_fe}, our method achieves the highest mIoU and Acc quantitatively and qualitatively. Specifically, compared to 3D-2D CRF, we obtain an absolute mIoU improvement of $2.4\%$ on perspective views and $11.9\%$ on fisheye views. The performance gap widens on the fisheye view as the pseudo-GT is less accurate on fisheye imagery. This observation also highlights the value of our method which is able to provide RGB images and semantic labels at novel viewpoints to enhance the generalization ability of 2D perception models. Despite PSPNet* being finetuned on KITTI-360 which reduces the performance gap, our method outperforms PSPNet* by a large margin. Supervised by the extremely sparse manually annotated GT, Semantic NeRF struggles to produce reliable performance. Using pseudo labels of PSPNet* and Tao \etal, JacobiNeRF and Panoptic Lifting are capable of denoising and thus improving performance $(67.2\% \rightarrow 68.5\%/69.6\%, 46.3\% \rightarrow 52.1\%/52.5\%)$. However, both JacobiNeRF and Panoptic Lifting are inferior in urban scenarios when the input views are sparse. In terms of perspective MC, ours is comparable to 3D-2D CRF and significantly surpasses 2D-to-2D label transfer methods. While our method slightly lags behind 3D Point + GC in terms of MC, this can be explained as label consistency is evaluated on sparsely projected 3D points which GC takes as input to generate a dense label map.

\boldparagraph{Panoptic Label Transfer}
We compare against Panoptic Lifting and 3D-2D CRF for panoptic label transfer, as the other baselines do not support distinguishing instances. As Panoptic Lifting is limited to segment instance classes known to pre-trained 2D panoptic segmentation models, it is not capable of segmenting the ``building'' class. Therefore, we consider two settings for quantitatively measuring the panoptic segmentation metrics: ``building'' as \textit{stuff} when comparing with Panoptic Lifting, and ``building'' as \textit{thing} when comparing with 3D-2D CRF.  \tabref{tab:instance_perspective_fe} shows that PanopticNeRF-360 outperforms the baselines in both cases. As shown in \figref{fig:instance_perspective}, Panoptic Lifting can not reconstruct instances well in far regions where Mask2Former and their instance assignment strategy perform poorly, yielding a larger gap in terms of the panoptic segmentation metrics. In contrast, PanopticNeRF-360 avoids this issue by leveraging weak 3D labels, consistently producing good performance in both near and far regions.
Our proposed method also outperforms the 3D-2D CRF in both things and stuff classes, despite that 3D-2D CRF leverages the same weak 3D labels as input. Visual comparisons are shown in \figref{fig:instance_perspective} and \figref{fig:instance_fe}. As can be seen, our method gracefully handles over- and under-exposure at buildings, which is a challenge for the 3D-2D CRF, as it projects LiDAR points to reconstruct the intermediate meshes whose quality suffers in these scenarios (see Supplementary). 

\boldparagraph{Panoramic Label Synthesis}
PanopticNeRF-360 can render RGB images and panoptic labels at omnidirectional novel viewpoints, whereas 3D-2D CRF is not capable of doing so. We therefore show qualitative panoramic results of our methods in~\figref{fig:more_panorama}.

\vspace{-2em}
\mdi{\subsection{Label Transfer on Waymo Dataset}}

We additionally conduct experiments on the Waymo dataset~\cite{mei2022waymo} to showcase PanopticNeRF-360's generalization ability. We choose two scenes, each consisting of 198 consecutive frames with the front-view camera. During data preparation, we use the KITTI-360 annotation toolkit to label 3D bounding primitives and generate 2D pseudo semantic labels with Mask2Former~\cite{cheng2022masked}. This annotation takes around 2 hours for each scene, i.e., the average per-frame labeling time takes only 0.6 minutes. We illustrate the entire set of labeled bounding boxes and camera trajectory in~\figref{fig:waymo_panopticnerf}. Ours not only renders high-fidelity appearance (28.28 dB on test views), but also generates high-quality semantic and instance masks on both pre-recorded and novel viewpoints. This demonstrates the applicability of our method.

\begin{table*}[h]
\centering
\resizebox{.97\textwidth}{!} {
\begin{tabular}{lc|l||ccccccccc|c|c} \hline  
\toprule
\md{Method}  & \md{Test View} & \md{F.T. Data} & \md{Road}  & \md{Sdwlk} & \md{Terr} & \md{Bldg} & \md{Vegt} & \md{Car} & \md{Wall} & \md{Fence} & \md{Sky} & \md{mIoU} & \md{Acc} \\  
\midrule
\midrule
\md{Mask2Former} & {\multirow{4}{*}{\md{Persp.}}} & \mdi{-} & \md{93.1} & \md{61.3} & \md{37.0} & \md{88.7} & \md{87.7} & \md{92.3} & \md{31.8} & \md{61.4} & \md{\textbf{94.8}} & \md{72.0} & \md{89.3}  \\
\mdi{Mask2Former-ft (coarse)} & & \mdi{(a)} & \mdi{94.2} & \mdi{66.2} & \mdi{37.5} & \mdi{90.1} & \mdi{\textbf{89.6}} & \mdi{92.8} & \mdi{35.7} & \mdi{63.1} & \mdi{94.3} & \mdi{73.7} & \mdi{90.2} \\ 
\mdi{Mask2Former-ft (coarse+ours)} & & \mdi{(a)+(b)} & \mdi{94.5} & \mdi{71.3} & \mdi{37.9} & \mdi{90.3} & \mdi{89.5} & \mdi{93.3} & \mdi{38.2} & \mdi{64.7} & \mdi{94.5} & \mdi{74.9} & \mdi{90.7} \\ 
Mask2Former-ft~\mdi{(ours)} & & \mdi{(b)} & \md{\textbf{94.6}} & \md{\textbf{72.1}} & \md{\textbf{38.4}} & \md{\textbf{90.9}} & \md{89.3} & \md{\textbf{93.4}} & \md{\textbf{40.9}} & \md{\textbf{66.8}} & \md{94.6} & \md{\textbf{75.7}} & \md{\textbf{91.1}} \\ 
\midrule
\md{Mask2Former} & {\multirow{4}{*}{\md{Fisheye}}} & - & \md{84.5}  & \md{60.7} & \md{20.5} & \md{80.9} & \md{77.1} & \md{60.9} & \md{26.5} & \md{42.3} & \md{\textbf{98.4}} & \md{61.3} & \md{89.1}  \\
\mdi{Mask2Former-ft (coarse)} & & \mdi{(a)} & \mdi{87.9} & \mdi{64.7} & \mdi{25.3} & \mdi{83.7} & \mdi{\textbf{89.1}} & \mdi{\textbf{62.3}} & \mdi{33.9} & \mdi{43.7} & \mdi{98.1} & \mdi{65.4} & \mdi{89.8} \\ 
\mdi{Mask2Former-ft (coarse+ours)} & & \mdi{(a)+(b)} &\mdi{88.3} & \mdi{68.4} & \mdi{26.5} & \mdi{83.5} & \mdi{88.9} & \mdi{61.9} & \mdi{40.3} & \mdi{44.5} & \mdi{97.7} & \mdi{66.7} & \mdi{90.4} \\ 
Mask2Former-ft~\mdi{(ours)} & & \mdi{(b)} & \md{\textbf{88.4}} & \md{\textbf{70.3}} & \md{\textbf{27.2}} & \md{\textbf{84.9}} & \md{88.4} & \md{\textbf{62.3}} & \md{\textbf{43.1}} & \md{\textbf{45.8}} & \md{98.3} & \md{\textbf{67.6}} & \md{\textbf{90.8}} \\ 
\bottomrule
\end{tabular}
}
\caption{\md{\textbf{Quantitative Comparison of Semantic Label} on Test Scenes of KITTI-360.} \mdi{(a): 512 coarse pseudo 2D labels (PSPNet*); (b): 1,536 synthesized labels using our method.}}
\label{tab:semantic_ft_rebuttal}
\end{table*}

\begin{table}[tb]
\centering
\resizebox{.49\textwidth}{!} {
\begin{tabular}{l||ccc|ccc} 
\toprule
& \multicolumn{3}{c|}{Forward-facing} & \multicolumn{3}{c}{Fisheye} \\
\midrule
Method & mIoU $\uparrow$ & PQ $\uparrow$ & PSNR $\uparrow$ & mIoU $\uparrow$ & PQ $\uparrow$ & PSNR$\uparrow$ \\
\midrule
MLP~\cite{mildenhall2020nerf} & \textbf{81.2} & \textbf{69.5} & 23.78 & \textbf{68.2} & \textbf{55.8} & 24.47 \\
iNGP~\cite{muller2022instant} & 79.0 & 67.9 & 27.22 & 67.0 & 54.6 & \textbf{28.14} \\
Tri-planes~\cite{chen2022tensorf}  & 75.1 & 62.1 & 23.25 & 60.2 & 50.2 & 23.32 \\
Ours & \underline{80.9} & \underline{68.7} & \textbf{27.25} & \underline{67.6} & \underline{55.4} & \underline{28.05} \\
\bottomrule
\end{tabular}
}
\caption{\textbf{Neural Scene Representation Study} over 4 scenes. We bold and underline the two best performing methods, respectively.} 
\label{tab:representation}
\end{table}

\begin{figure*}[!ht]
 \centering
\includegraphics[width=0.99\textwidth]{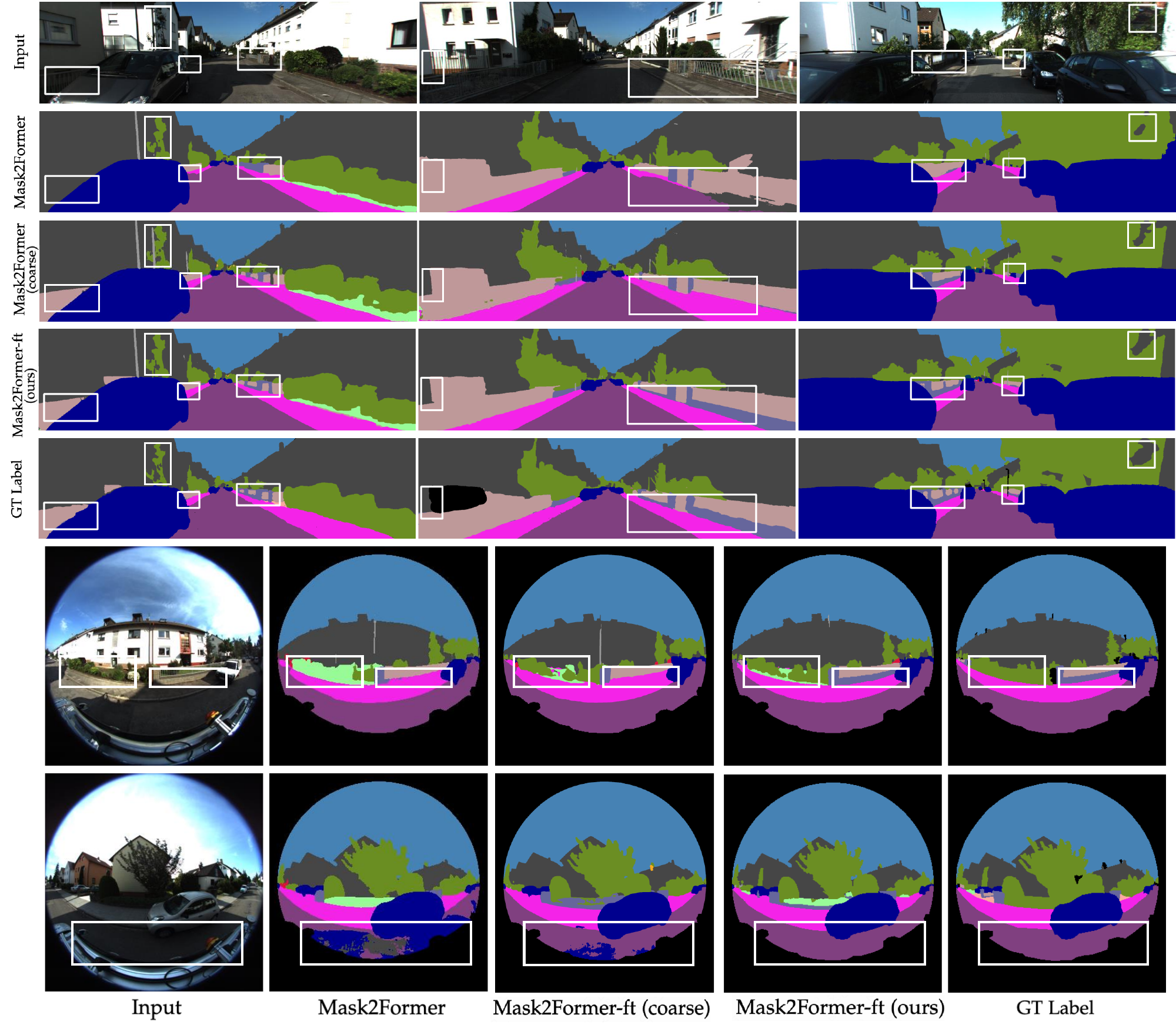}
 \caption{
  \textbf{Qualitative Comparison of Semantic Prediction on Perspective and Fisheye Views.} }
 \label{fig:perception_ft_rebuttal}
\end{figure*}

\subsection{Neural Scene Representation}
\begin{figure*}[tb]
 \centering
 \newcommand{\mywidth}{0.185\textwidth}
 \setlength\tabcolsep{0.2em}
 \begin{tabular}{ccccc}
  \includegraphics[width=\mywidth]{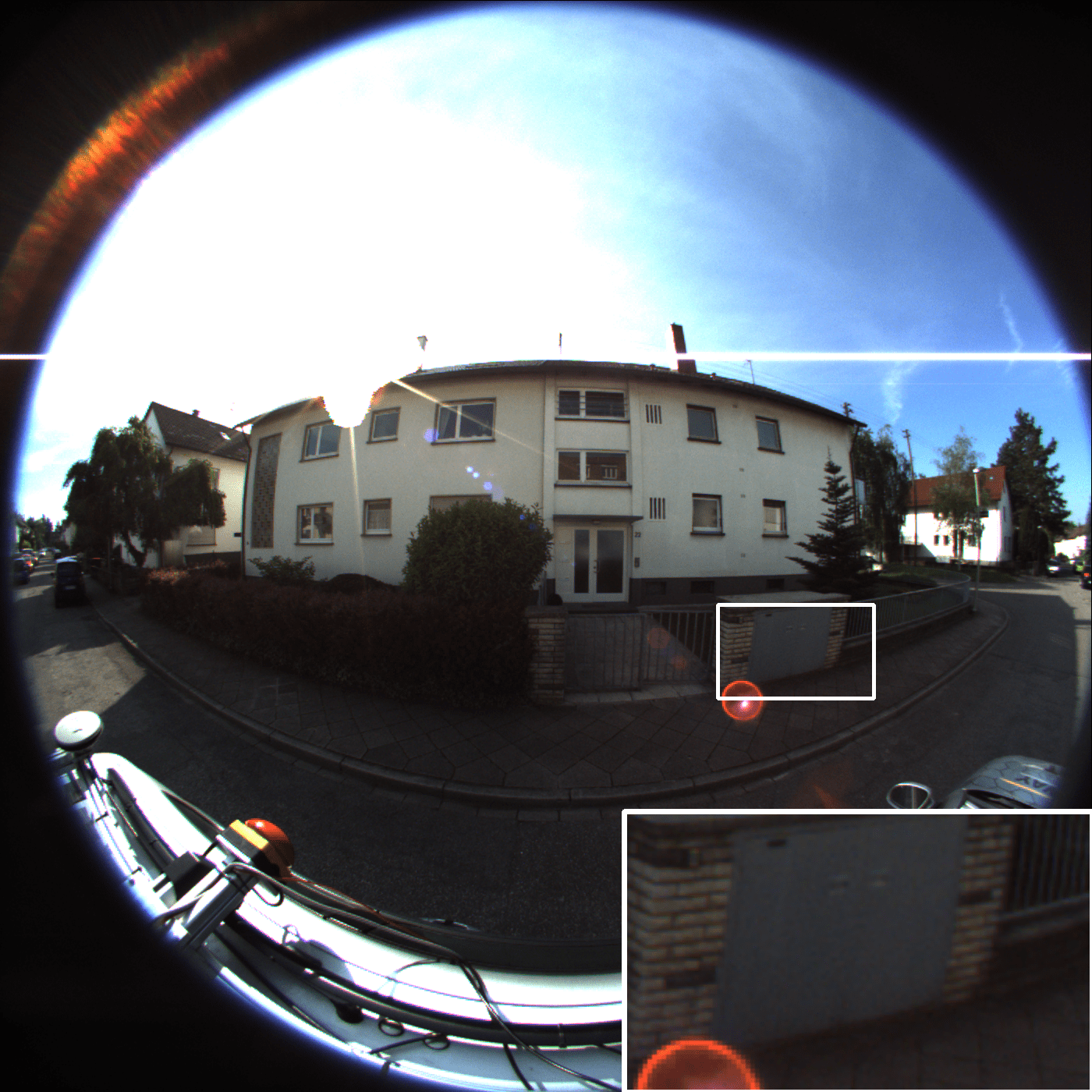} &
  \includegraphics[width=\mywidth]{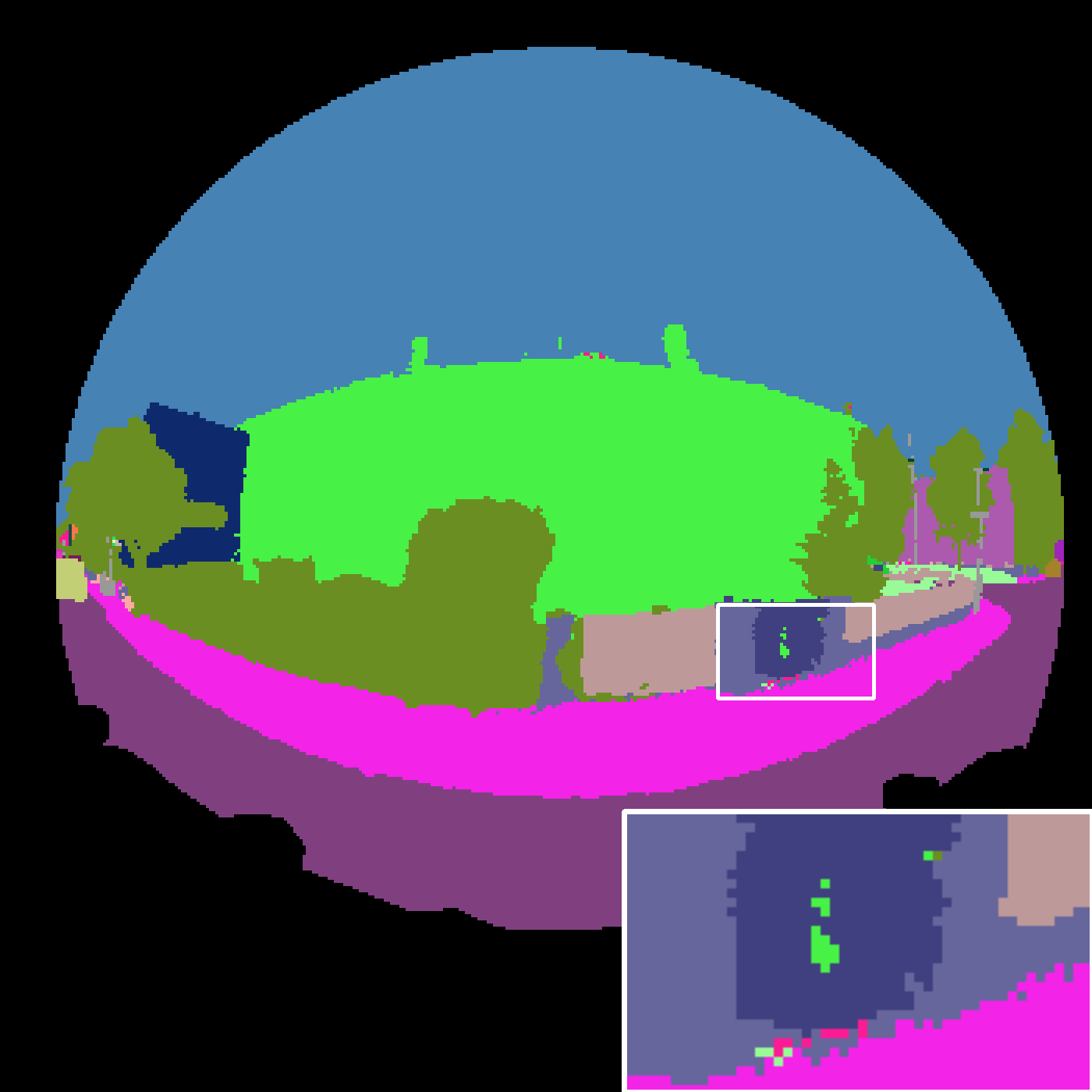} &
  \includegraphics[width=\mywidth]{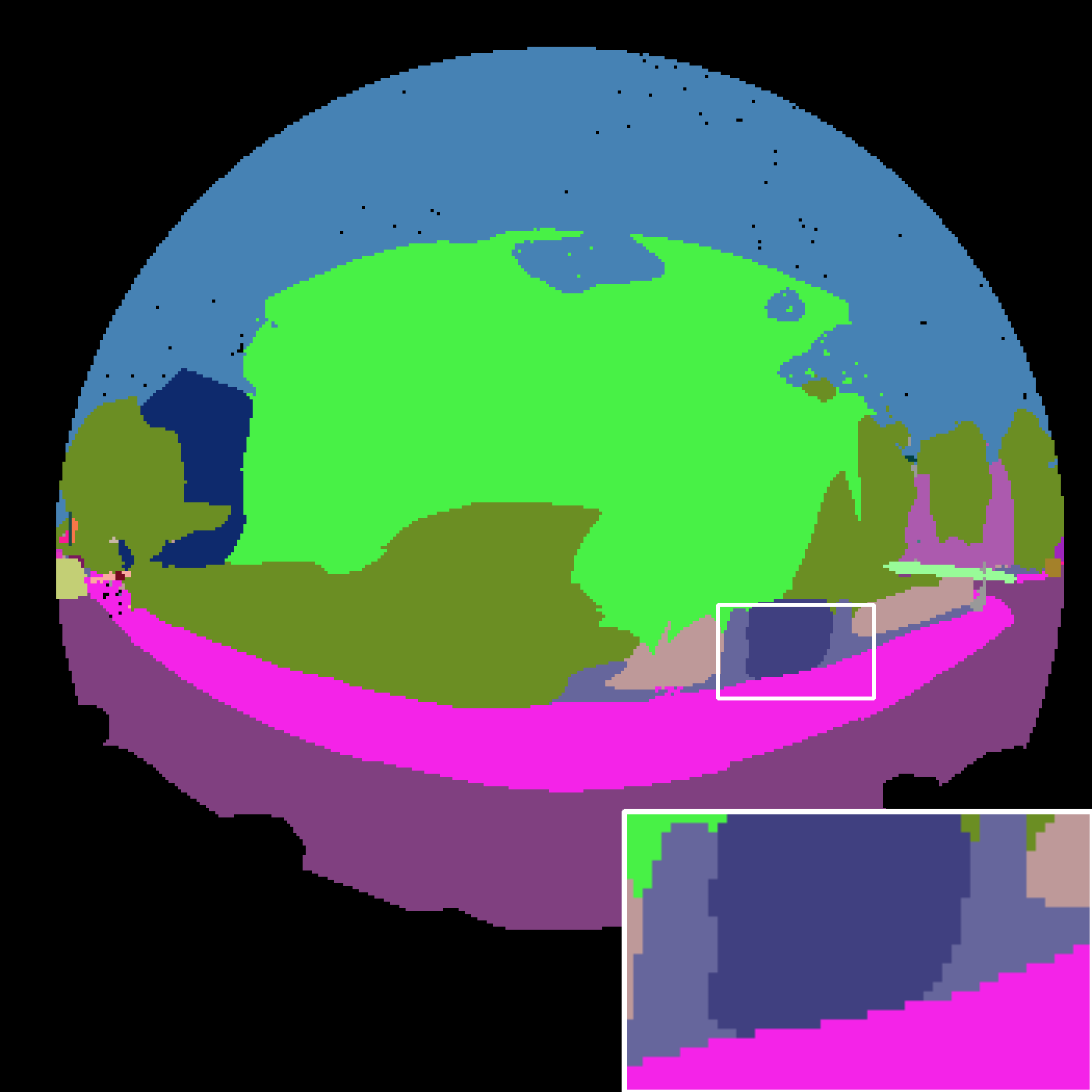} &
  \includegraphics[width=\mywidth]{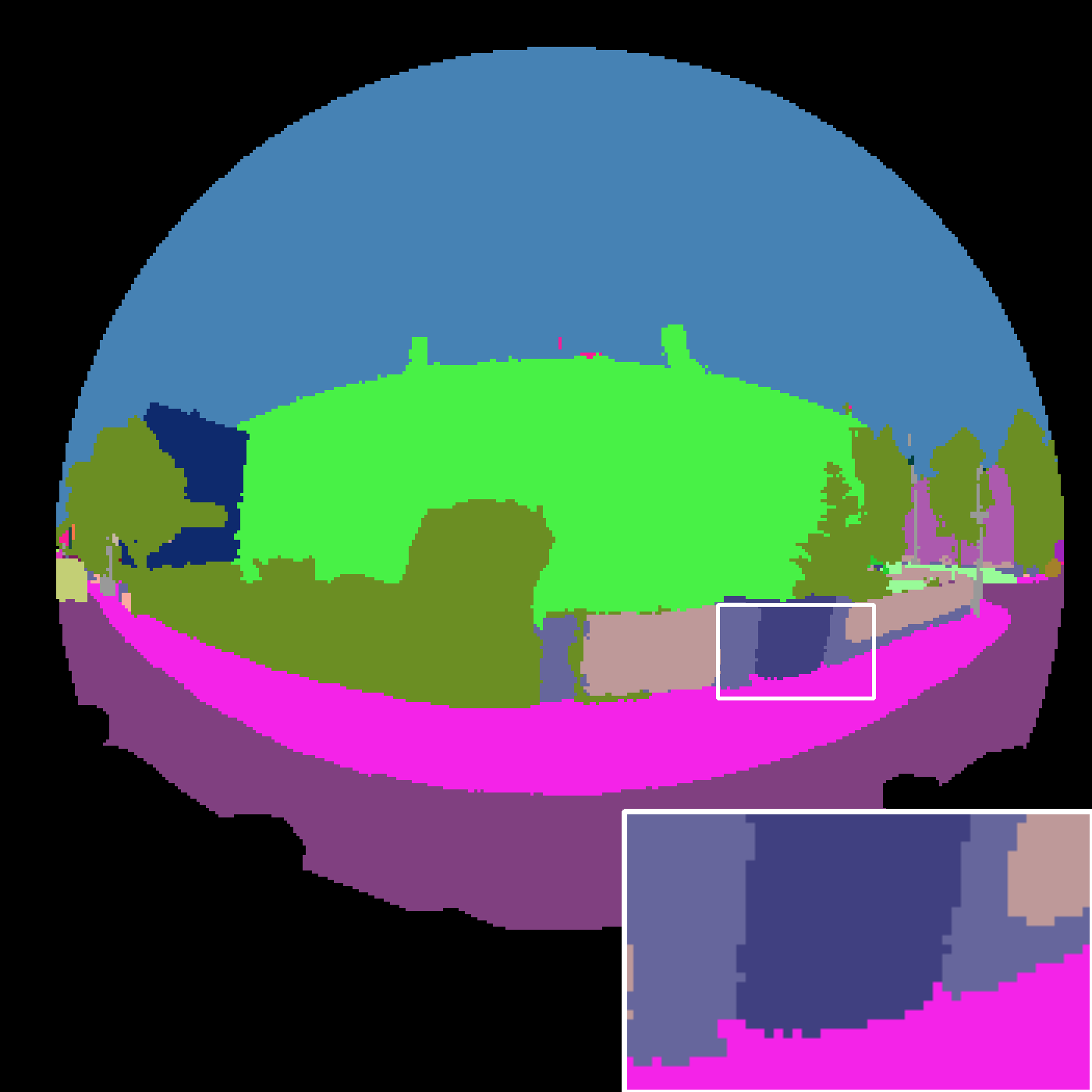} &
  \includegraphics[width=\mywidth]{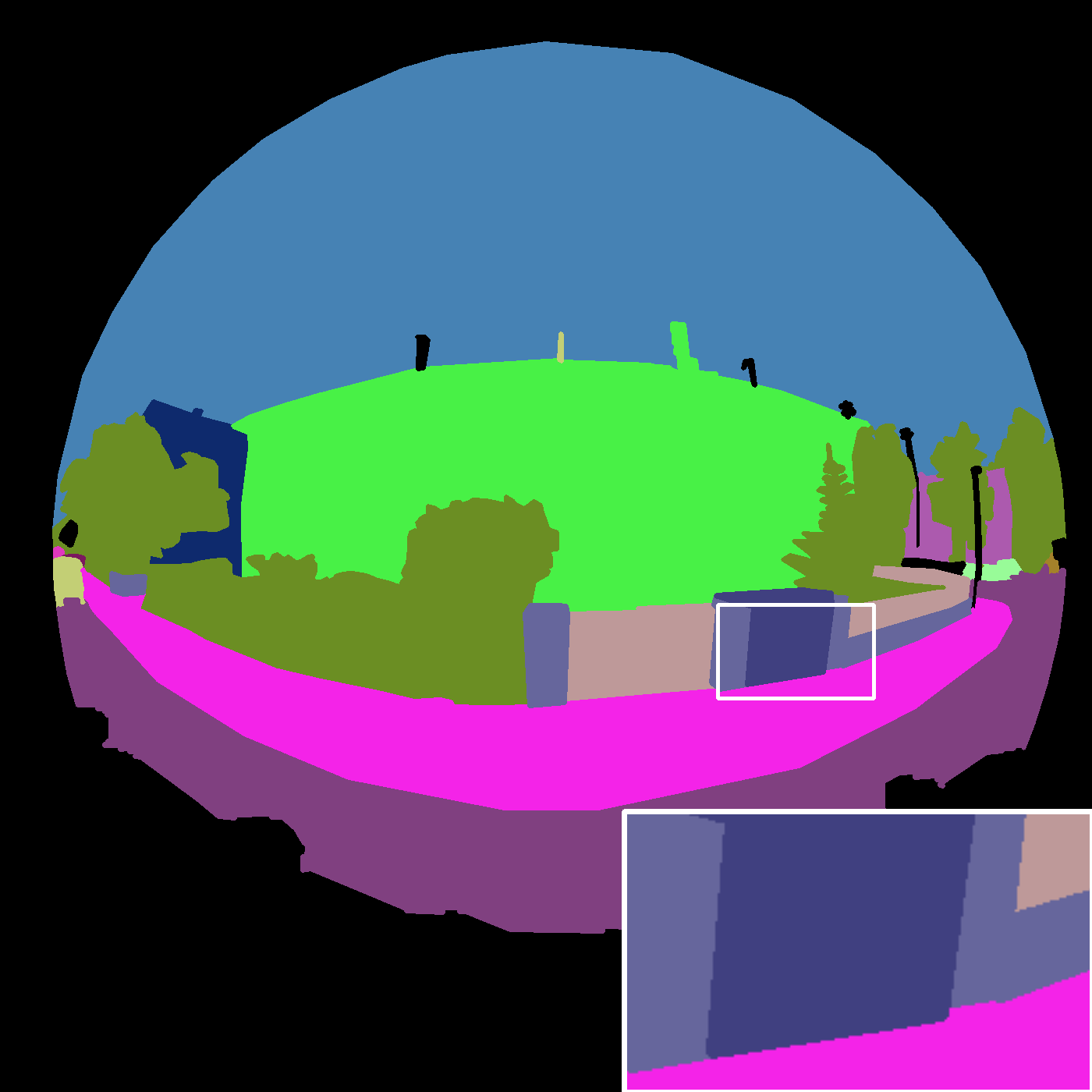}
    \\
    \small{Input} &  
    \small{iNGP~\cite{muller2022instant}} &  
    \small{Tri-planes~\cite{chen2022tensorf}} &  
    \small{Ours} & 
    \small{GT Label}
 \end{tabular}
 \caption{
  \textbf{Qualitative Comparison of Neural Scene Representations on Fisheye.} Ours generates smoother labels than others.
  }
 \label{fig:representation}
\end{figure*}

We further compare iNGP~\cite{muller2022instant}, Tri-planes~\cite{chen2022tensorf} (TensoRF-VM), and MLP~\cite{mildenhall2020nerf} as scene representation. Towards this goal, we analyze the results over 4 scenes where ambient conditions vary between each other. For a fair comparison, we run each scene for 30,000 iterations. Please refer to the supplementary for implementation details. \tabref{tab:representation} indicates that while the pure MLP demonstrates superior label quality, it falls short in appearance reconstruction. On the other hand, iNGP excels in novel view synthesis ($\sim$3.5dB higher than MLP) and convergence, albeit at the cost of subpar semantic label quality in comparison to the MLP. Our proposed scene feature, a hybrid of MLP and hash grids, achieves a tradeoff between label quality and appearance. Tri-planes fail to represent high-fidelity scene structure both in terms of semantic labels and appearance, especially at boundaries between foreground and background (sky) and in over-exposed regions as grid-based methods are not robust to filter noise and tend to suffer from local minimum. While iNGP is also grid-based, it conditions features on multi-scale levels, thereby potentially aggregating global information, and employs a hash function for non-periodic signal embedding. As depicted in \figref{fig:representation}, iNGP contains more noise than ours, and Tri-planes fail to reconstruct the overall label, especially at buildings over-exposed by sunlight.

\subsection{Synthesized Labels for Perception Models}

To validate that the synthesized novel view labels are important for autonomous driving and robotics applications, 
we fine-tune Mask2Former~\cite{cheng2022masked} pre-trained on CityScape~\cite{cordts2016cityscapes} using our \textit{synthesized} semantic labels on the KITTI-360. We fine-tune it using our synthesized semantic labels on the KITTI-360. Specifically, we divide 10 test scenes into 8 for training and 2 for test. For each scene, we render 3 semantic maps at each of 64 trajectory points using the left perspective camera: one in the original direction, one rotated horizontally 30° to the left, and another rotated horizontally 30° to the right, resulting in a total of 1,536 training samples. \mdi{We further compare Mask2Former-ft (ours) with two additional variants: one fine-tuned solely on coarse semantic labels that we use as pseudo 2D GTs (coarse), and another trained on a combination of coarse and synthesized images (coarse + ours).} During fine-tuning, we format the data to match CityScape's and exclude categories not present in CityScape.  We fine-tune~\mdi{all the models} for 5 epochs with a batch size of 8, using a learning rate of 0.0001 and AdamW optimizer. As illustrated in \tabref{tab:semantic_ft_rebuttal} and \figref{fig:perception_ft_rebuttal}, Mask2Former-ft (ours) significantly outperforms the original (+3.7 mIoU on perspective views, and +6.3 mIoU on fisheye views), especially on Sidewalk and Wall. \mdi{It is also reasonable that Mask2Former-ft (ours) outperforms the other two finetuning baselines as our synthesized labels significantly surpass the base 2D pseudo labels.} This improvement demonstrates the effectiveness of our synthesized labels in enhancing the perception model's performance.  Moreover, the larger improvement on the fisheye views further verifies that existing perception models degenerates on unseen viewpoints and demonstrates the importance of being able to synthesizing labels with large viewpoint changes.

\subsection{Ablation Study} \label{sec:ablation}
We validate our pipeline's design modules with extensive ablations in \tabref{ablation} by removing one component at a time. We perform label evaluation on joint perspective $\&$ fisheye views and estimate scene geometry on perspective views on one scene. 

\begin{table}[tb!]
\centering
\resizebox{.49\textwidth}{!} 
{
\begin{tabular}{c||cc|c|ccc} 
\toprule
& \multicolumn{2}{c|}{Depth~(0-100m)} & Eval. & \multicolumn{3}{c}{Perspective~$\&$~Fisheye} \\
&RMSE$\downarrow$ &$\delta_{1.25}\uparrow$ & Label & mIoU* &Acc* &PQ* \\ 
\midrule
3D-2D CRF& - & - & - & 69.7 & 94.1 & 57.6 \\
\midrule
NeRF-360* & 19.11 & 75.5 & $\hat{\bS}(s_\beta)$ & 64.6 & 88.9 & 54.6 \\
w/o $\cL^{\text{2D}}_{\hat{\bS}}$& 16.90 & 78.4 & $\bS(s_\phi)$ &71.3 & 93.8& 61.6 \\
w/o $\cL^{\text{2D}}_{\bS}$& 7.57 & 94.7& $\bS(s_\phi)$ & 74.6 & 94.9 & 64.0 \\
w/o $\cL^{\text{2D}}_{\bS}$& 7.57 & 94.7& $\hat{\bS}(s_\beta)$ & 73.4 & 94.8 & 62.3 \\
w/o $\cL^{\text{3D}}_{\bs}$& 7.53 & 94.7 & $\bS(s_\phi)$ & 75.0 & 95.0 & 65.3 \\
w/o $\cL_{d}$& 8.36 & 93.4 & $\bS(s_\phi)$ & 74.4 & 94.9 & 63.8 \\
w/o $\mathbbm{1}(\br)$& 8.29 & 94.5& $\bS(s_\phi)$ & 75.1 & 95.0 & 65.4 \\
Uniform S. & 9.52 & 93.5 & $\bS(s_\phi)$ & 67.3 & 92.2 & 56.7 \\
S. points$\downarrow$& 7.36 & 94.6 & $\bS(s_\phi)$ & 75.3 & 95.1 & 65.4 \\
w/o $w(k)$& \textbf{7.12} & \textbf{94.8} & $\bS(s_\phi)$ & 75.6 & 95.1 & 66.2 \\
w/o fisheye*& 7.90 & 92.5 & $\bS(s_\phi)$ & 74.8 & 94.7 & 64.0 \\
\midrule
Complete & 7.27 & \textbf{94.7}& $\bS(s_\phi)$ & \textbf{76.1} & \textbf{95.2}  & \textbf{66.6} \\ 
\bottomrule
\end{tabular}
}
\caption{\textbf{Ablation Study} over one scene.} 
\label{ablation}
\end{table}

\begin{figure*}[tb]
 \centering
 \newcommand{\mywidth}{0.235\textwidth}
 \setlength\tabcolsep{0.2em}
 \begin{tabular}{cccc}

   \includegraphics[width=\mywidth]{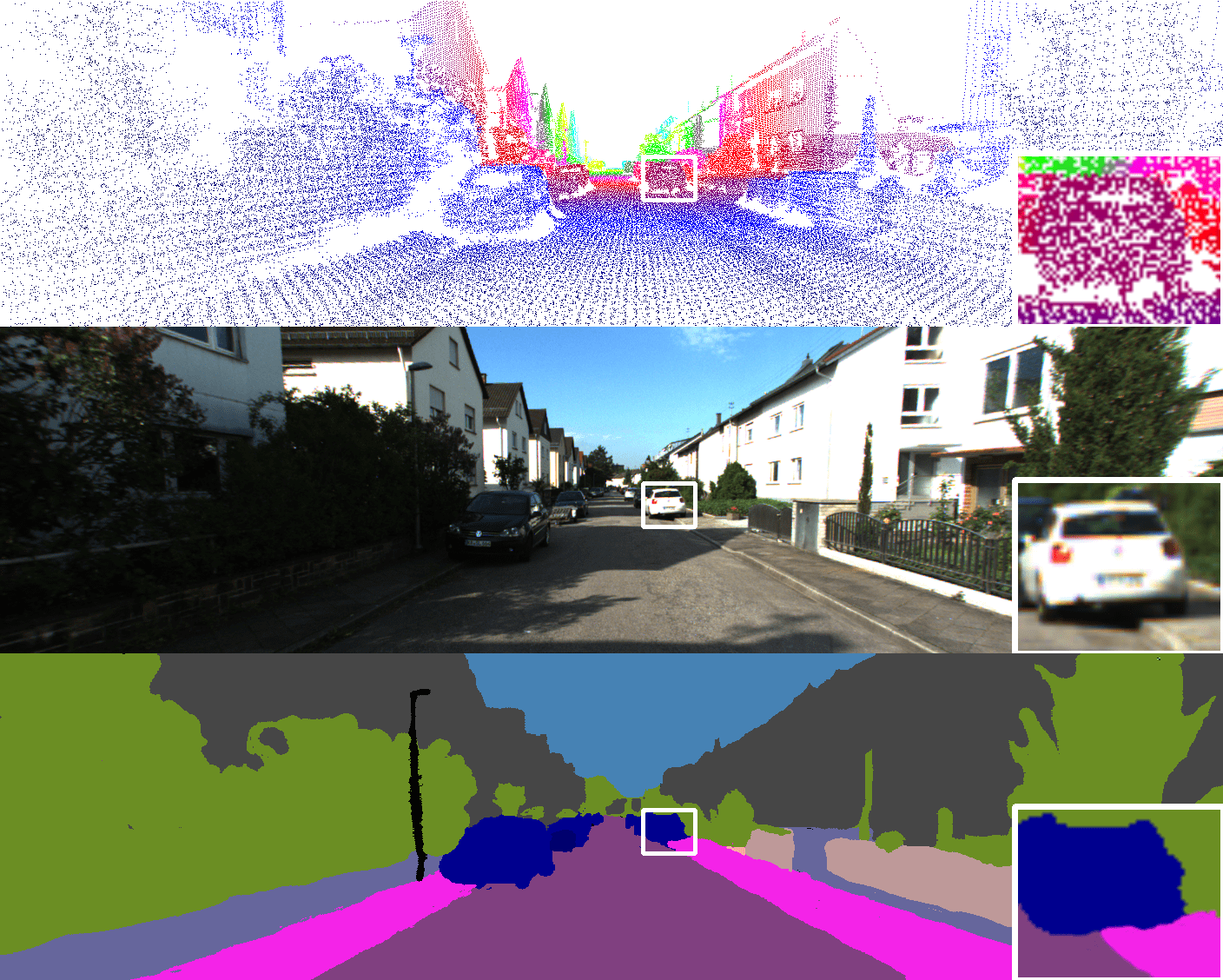}&
  \includegraphics[width=\mywidth]{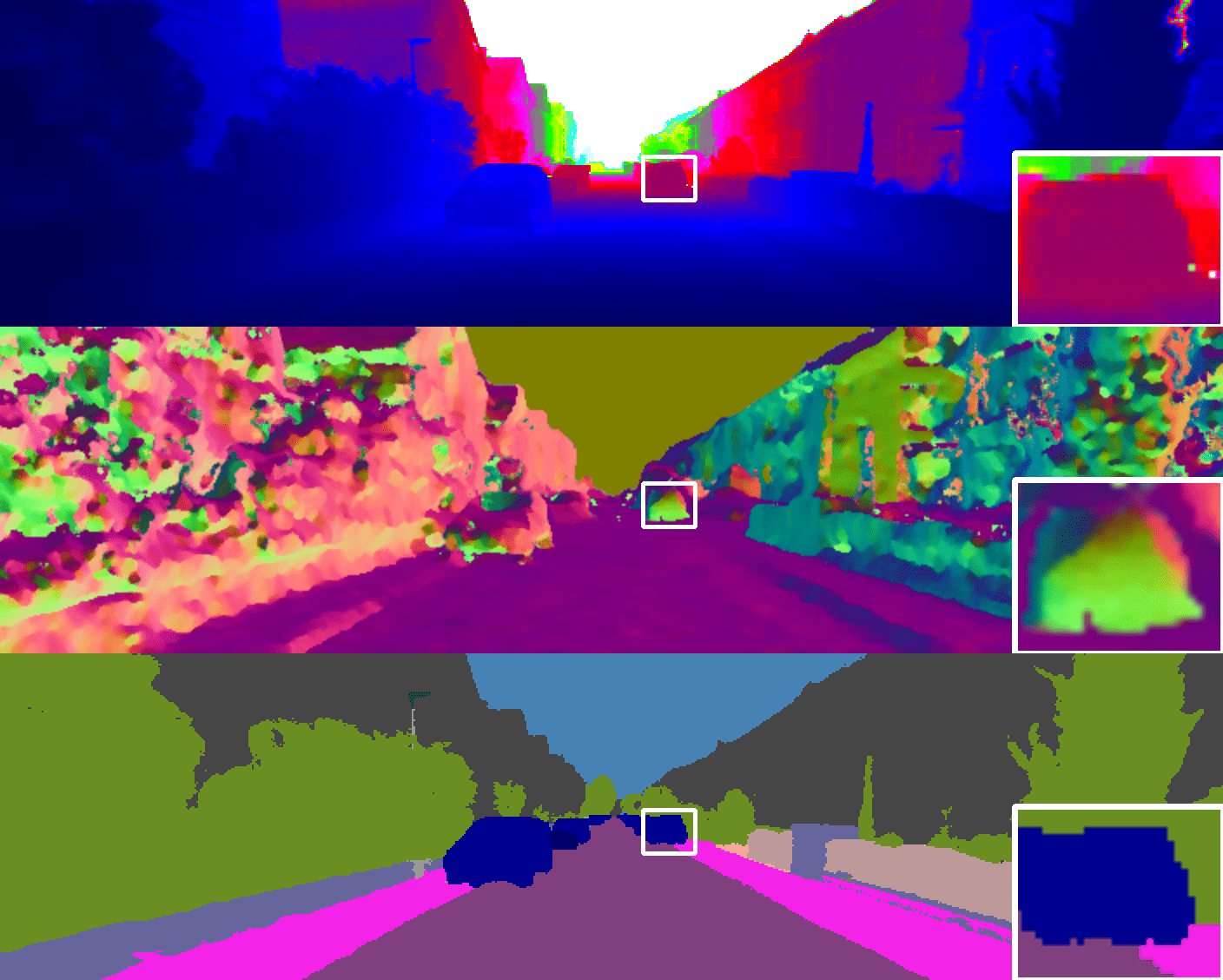}&
  \includegraphics[width=\mywidth]{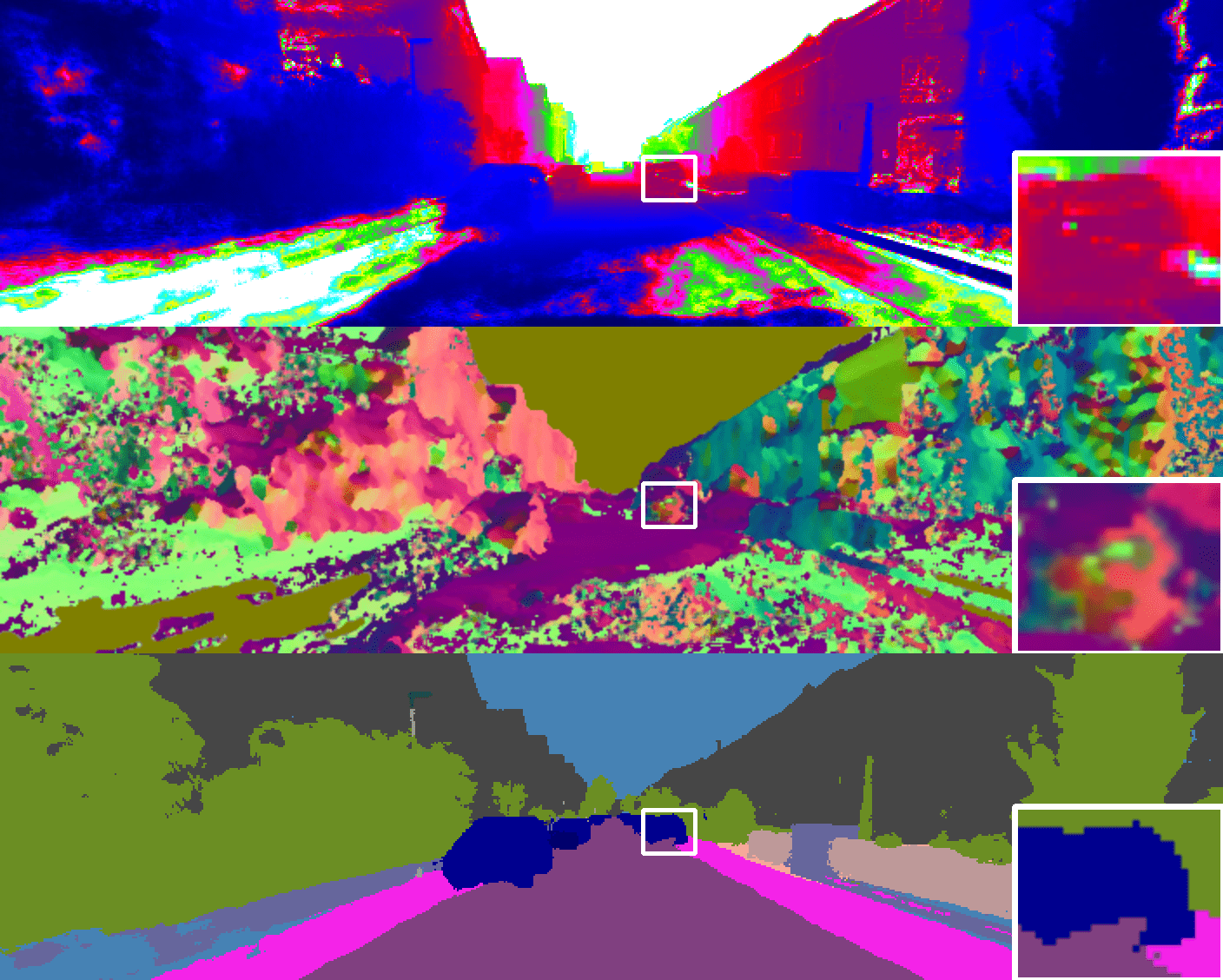}&
  \includegraphics[width=\mywidth]{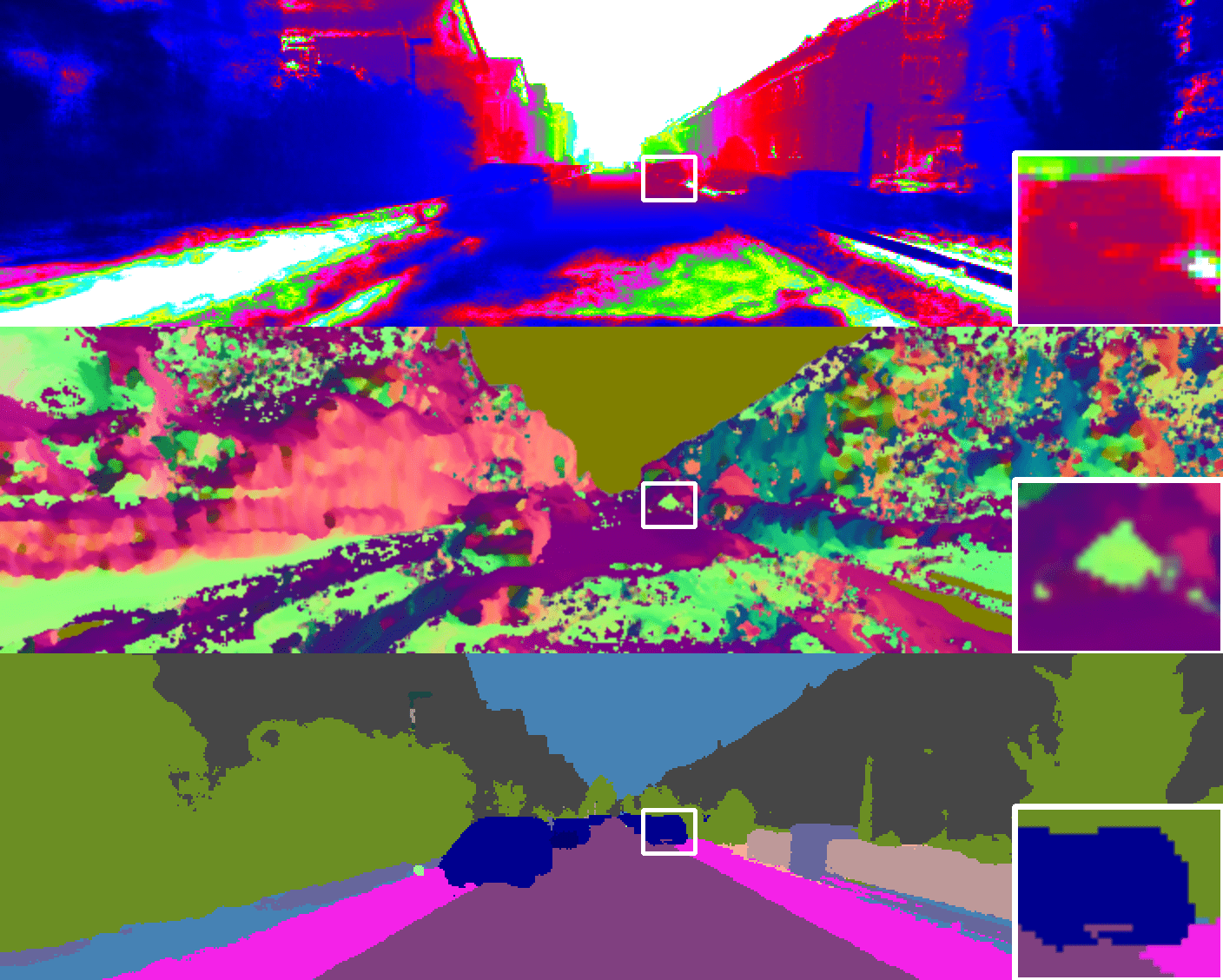}
  \\
     \small{GT}&
     \small{Complete}  &  
     \small{w/o $\cL^{\text{2D}}_{\hat{\bS}}$}& 
     \small{NeRF-360*} 
 \end{tabular}
 \vspace{-0.3cm}
 \caption{
  \textbf{Ablation Study.} Top: LiDAR depth map and rendered depth maps. Middle: RGB input and normal maps computed as the gradient of the volume density with respect to 3D position. Bottom: Semantic GT and predictions. Note that removing $\cL^{\text{2D}}_{\hat{\bS}}$ leads to severely impaired geometry, and inaccurate boundary and semantic segmentation.}
 \label{fig:depth}
\end{figure*}

\begin{figure}[!t]
 \centering
 \newcommand{\mywidth}{.46\textwidth}
 \setlength\tabcolsep{0.05em}
 \newcolumntype{P}[1]{>{\centering\arraybackslash}m{#1}}
 \def\arraystretch{0.50}
  \begin{tabular}{P{0.5em}P{0.5em}P{\mywidth}}
    \rot{\scriptsize{RGB Image}}&& \includegraphics[width=\mywidth]{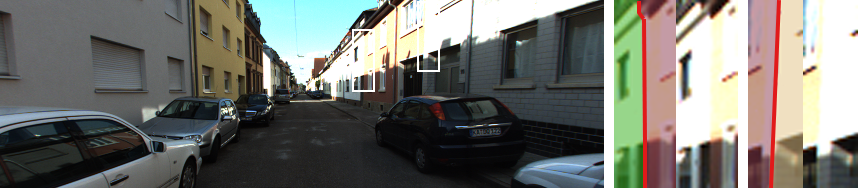} \\
     \rot{\scriptsize{(a)}}&& \includegraphics[width=\mywidth]{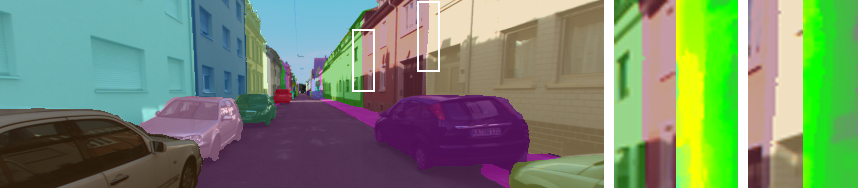} \\
     \rot{\scriptsize{(b)}}&& \includegraphics[width=\mywidth]{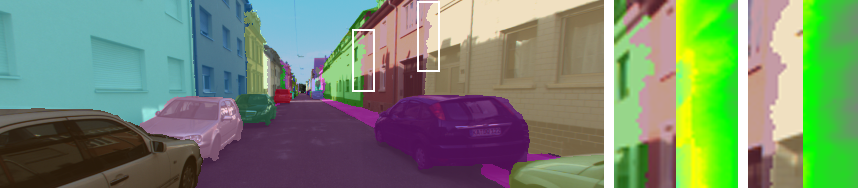}\\
     \rot{\scriptsize{(c)}}&& \includegraphics[width=\mywidth]{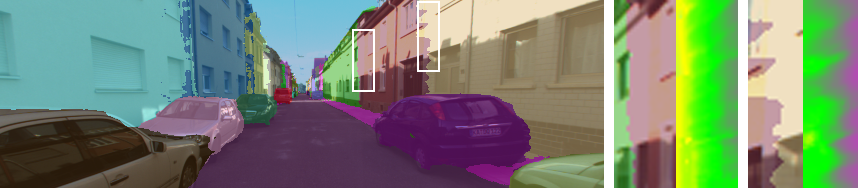}\\
 \end{tabular}
 \caption{
  \textbf{Qualitative Ablations of Fine-tuning Stage.} (a) full model (b) w/o finetuning instance (c) w/o fisheye, w/o finetuning instance. Zoom in the visualization of the overexposed area between adjacent ``building''s and the corresponding depth. The red lines mark the GT boundaries.
 }
 \label{fig:ft_instance}
\end{figure}

\boldparagraph{Geometric Reconstruction} 
1) \textit{Semantic Label Guidance}: We now verify that our method effectively improves the underlying geometry by leveraging semantic information. We first remove all the other losses except for $\cL_{p}$, leading to a baseline, NeRF-360*, which still employs a hybrid of MLP and grids and uses our proposed sampling strategy. In this case, we render a semantic map based on the fixed semantic field $s_\beta$. As can be seen from \tabref{ablation} and \figref{fig:depth}, the underlying geometry of NeRF-360* drops significantly with only using $\cL_{p}$. More importantly, the depth prediction also degrades considerably when removing $\cL^{\text{2D}}_{\hat{\bS}}$ (w/o $\cL^{\text{2D}}_{\hat{\bS}}$, RMSE: 7.27$\rightarrow$16.90), indicating the importance of the fixed semantic field in improving the underlying geometry to render accurate semantic boundaries. This can also be verified by removing fisheye semantic predictions from the input (w/o fisheye*), which results in more inaccurate geometry ($\delta_{1.25}$: 94.7$\rightarrow$92.5) and semantics (mIoU*: 76.1$\rightarrow$74.8). 

\noindent 2) \textit{Instance Label Guidance}: To illustrate the effectiveness of $\cL^{\text{2D}}_{\hat{\bT}}$ (\eqref{eq:loss_fix_instance}), we present \figref{fig:ft_instance} for qualitative comparision. In the marked region with overexposure, the full model achieves more accurate boundaries and smoother geometry than the model without the fine-tuning stage (w/o ft. instance). 

\noindent 3) \textit{Others}: For the hybrid scene features, as shown in \figref{fig:hybrid_concat_product}, ``concatenation'' performs better than ``product'' in merging MLP and hash features, leading to smoother and less erroneous geometry. We further show that eliminating $\cL_d$ (w/o $\cL_d$), replacing the sampling strategy with standard uniform sampling (Uniform S.), or removing ray masking (w/o $\mathbbm{1}(\br)$) all impair the geometric reconstruction, and consequently the semantic estimation. 

\boldparagraph{Panoptic Segmentation} When removing $\cL^{\text{2D}}_{\bS}$ (w/o $\cL^{\text{2D}}_{\bS}$), the performance also drops as the learned semantic field $s_\phi$ is only supervised by weak 3D supervision. Interestingly, this baseline still outperforms the semantic map rendered by the fixed semantic field despite that they share the same geometry. This observation suggests that the weak 3D supervision provided by $\cL^{\text{3D}}_{\bs}$ also allows us to address label ambiguity in overlapping regions to a certain extent. Therefore, it is not surprising that removing $\cL^{\text{3D}}_{\bs}$ (w/o $\cL^{\text{3D}}_{\bs}$) worsens the label prediction compared to the full model. Discarding $w(k)$ (w/o $w(k)$) causes worse label quality by sacrificing geometry (RMSE:7.27$\rightarrow$7.12) to model imbalanced label distribution. 

\boldparagraph{2D Pseudo GT}Finally, we evaluate how the quality of the pseudo 2D ground truth affects our method in \tabref{tab:pseudo_label}. As some classes are not considered during training in Cityscapes, we additionally report mIoU$_\text{sub}$ over the remaining classes. It is worth noting that using models pre-trained on Cityscapes without any fine-tuning leads to promising results, where  Ours w/ Deeplab~\cite{chen2017deeplab} and Ours w/ Tao \etal~\cite{tao2020hierarchical} are very close to Ours + PSPNet* in terms of mIoU$_\text{sub}$. More importantly, our method consistently outperforms the corresponding 2D pseudo GT by leveraging the 3D bounding primitives. We additionally incorporate SSA, a SAM-based semantic variant to demonstrate the inefficiency of SAM~\cite{kirillov2023segment} despite its unprecedented zero-shot mask generation ability.

\begin{table}[!tb]
\centering
\resizebox{.38\textwidth}{!} {
\begin{tabular}{c||cccc} 
\toprule
Method &mIoU* &mIoU$_\text{sub}$* &Acc* &PQ*\\ 
\midrule
SSA~\cite{chen2023semantic}& - & 67.4 & 86.0 & -\\
Deeplab~\cite{chen2017deeplab} &  - &73.3 & 91.0 & -\\
Tao~\textit{et al.} ~\cite{tao2020hierarchical}& - &75.9 & 92.5 & -\\
PSPNet~\cite{zhao2017pyramid} & - &71.4 & 90.4 & -\\
PSPNet*~\cite{zhao2017pyramid} & 67.0 & 74.7 & 92.0 & -\\
\midrule
Ours w/~\cite{chen2023semantic}& 65.2 & 71.6 & 90.9 & 58.6\\
Ours w/~\cite{chen2017deeplab} & 73.3 & 78.6 & 94.5 & 63.6\\
Ours w/~\cite{tao2020hierarchical} & 73.9 & 78.4 & 95.1 & 64.9\\
Ours w/~\cite{zhao2017pyramid} & 70.7 & 74.3 & 92.3 & 62.5\\
Ours & \textbf{76.1} & \textbf{79.6} & \textbf{95.2} & \textbf{66.6}\\
\bottomrule
\end{tabular}
}
\caption{\textbf{Quantitative Comparison} using different 2D Pseudo GTs on one scene. We keep the pesudo labels for the two side-view fisheye cameras and only change the pseudo labels on the forward-facing views.}
\label{tab:pseudo_label}
\end{table}

\section{Conclusion}
We present PanopticNeRF-360 that infers in 3D space and renders omnidirectional per-pixel semantic and instance labels for 3D-to-2D label transfer. By unifying coarse 3D bounding primitives and noisy 2D semantic predictions, PanopticNeRF-360 is capable of performing mutual enhancement of geometry and semantics compared to na\"ive joint optimization of geometry and semantics. Specifically, it improves the underlying scene geometry given sparse input views leveraging label-guided geometry optimization, while concurrently resolving label noise based on improved geometry. Moreover, it enables label synthesis at a large range of novel viewpoints, including panoramic perspectives. We posit that our method marks a significant step towards improving data annotation efficiency while delivering a consistent, continuous 3D panoptic representation.

\subsection{Limitations}

Our method has several limitations: (1) We perform per-scene optimization. To further shorten the time for label transfer, future works might explore how to endow the model with training-free inference abilities on novel scenarios. This can be potentially realized by pretaining on a large number of urban scenes of diverse environments. (2) In cases where there are missing bounding boxes in distant regions, our method cannot accurately recover the correct labels and typically categorizes these areas as "sky." However, the impact of this on image-based models is usually negligible since the very distant regions represent only a small portion of the 2D image space.  (3) We focus on label transfer of static scenes. It will be interesting to extend our method to dynamic scenes given annotated bounding primitives of the dynamic objects.

\boldparagraph{Potential Improvement on Backbone} 3D Gaussian Splatting~\cite{kerbl20233d} has demonstrated remarkable training and rendering efficiency in comparison to NeRF-like architectures. It is indeed a very interesting future direction to be explored such efficient representations to scaling the proposed method to large-scale scenes~\cite{lu2023scaffold,lin2024vastgaussian,liu2024citygaussian}. With the 3D bounding primitives, we can seamlessly employ the points within them for initial setup. With the 3D bounding primitives, we can seamlessly employ the points within them for initial setup. Besides, the explicit representation of gaussian splatting is more suitable for modeling and visualizing semantic occupancy compared to NeRF (implicit function).

\section*{Acknowledgments}
This work was supported by NSFC under grant 62202418, U21B2004, and the Zhejiang University Education Foundation Qizhen Scholar Foundation.
Andreas Geiger was supported by the ERC Starting Grant LEGO-3D (850533) and the DFG EXC number 2064/1 - project number 390727645.

{\small
\bibliographystyle{ieee_fullname}
\bibliography{bibliography_short,bibliography_custom}
}

\begin{IEEEbiography}[{\includegraphics[width=1in,height=1.25in,clip,keepaspectratio]{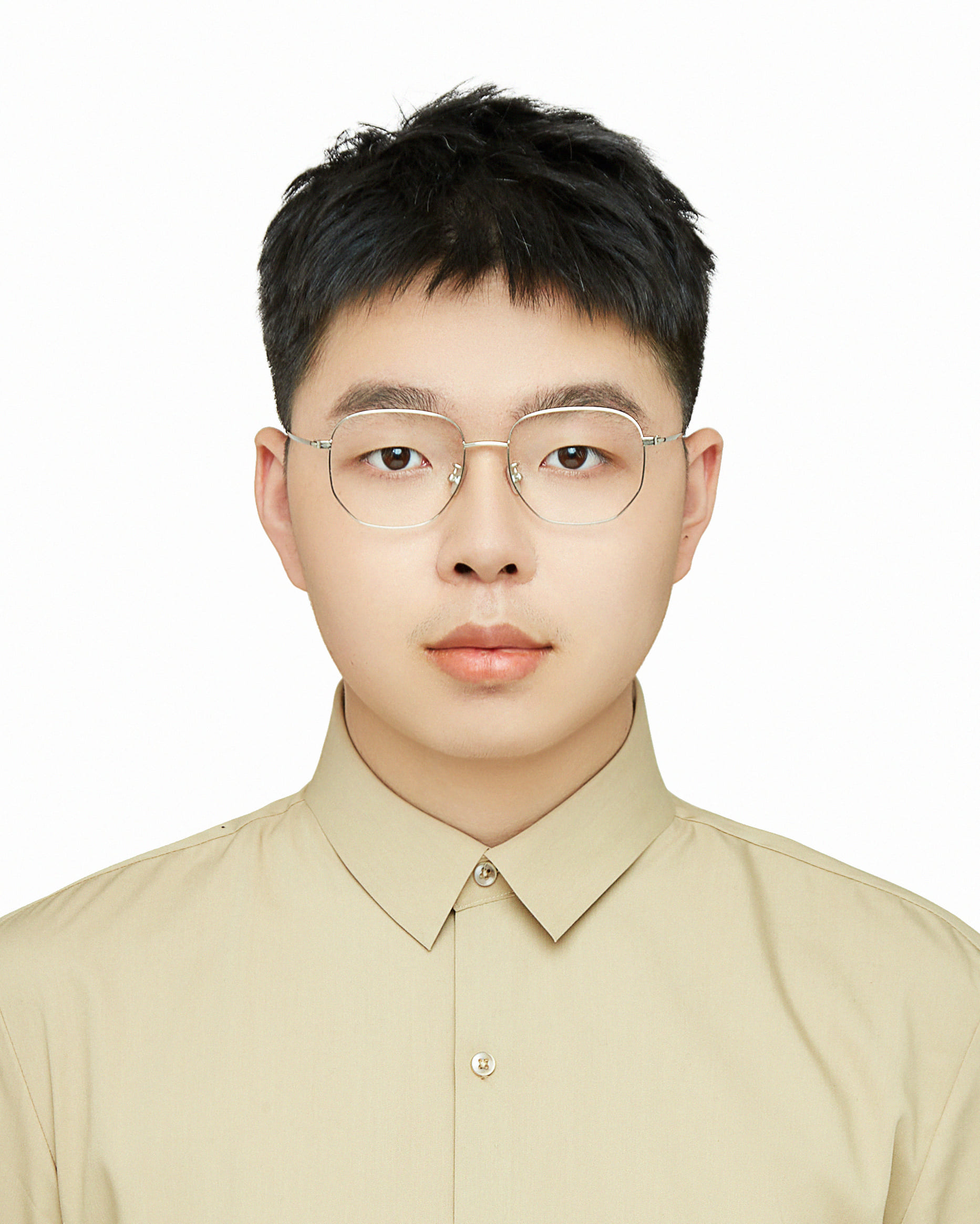}}]{Xiao Fu}
 received bachelor degree in Information Engineering at Zhejiang University in 2022, advised by Prof. Yiyi Liao. His research interest lies in 3D computer vision and machine learning, including neural rendering, 3D/4D reconstruction, generation and scene editing.
\end{IEEEbiography}

\begin{IEEEbiography}[{\includegraphics[width=1in,height=1.25in,clip,keepaspectratio]{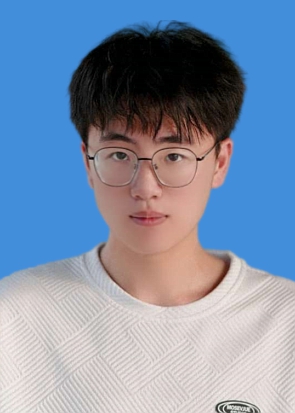}}]{Shangzhan Zhang}
is a highly motivated master's student in Computer Science at Zhejiang University, where he is advised by Professor Xiaowei Zhou. He received his bachelor's degree from the same university in 2022.
\end{IEEEbiography}

\begin{IEEEbiography}[{\includegraphics[width=1in,height=1.25in,clip,keepaspectratio]{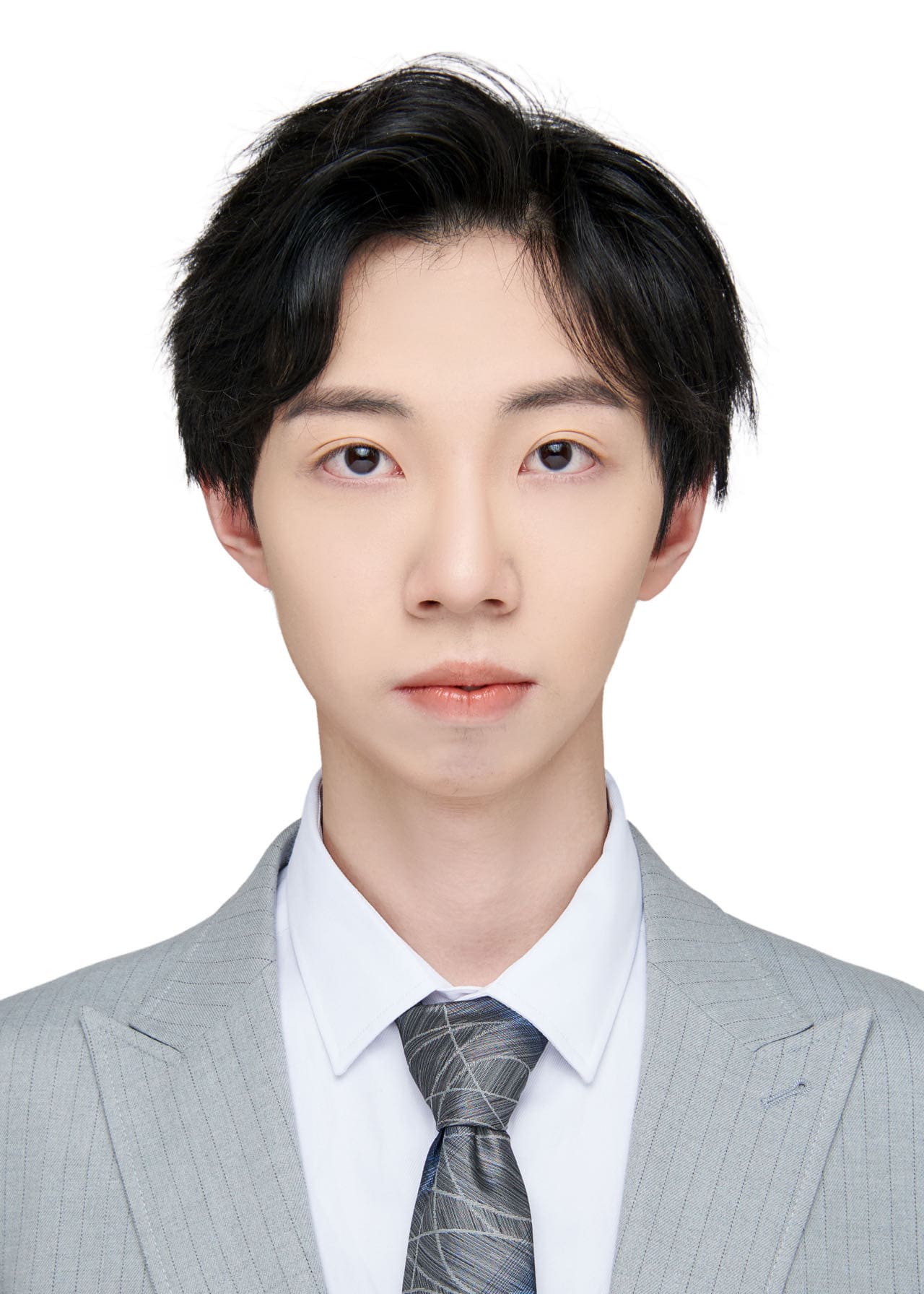}}]{Tianrun Chen}
received the bachelor's degree in College of Information Science and Electronic Engineering, Zhejiang University and is persuing Ph.D. degree in Computer Science and Technology at Zhejiang University. He is also the founder and the technical director of Moxin Technology (KOKONI) Co., LTD. His research interest includes computer vision and its enabling applications. 
\end{IEEEbiography}

\begin{IEEEbiography}[{\includegraphics[width=1in,height=1.25in,clip,keepaspectratio]{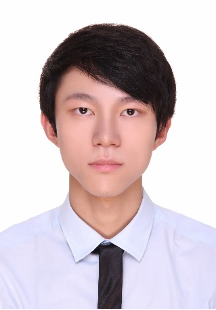}}]{Yichong Lu}
is a senior undergraduate student in College of Information Science and Electronic Engineering, Zhejiang University. His research interest includes 3D reconstruction, generation and editing. 
\end{IEEEbiography}

\begin{IEEEbiography}[{\includegraphics[width=1in,height=1.25in,clip,keepaspectratio]{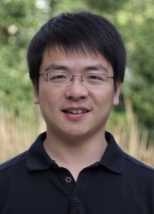}}]{Xiaowei Zhou}
is a Research Professor of Computer Science at Zhejiang University, China. He obtained his Ph.D. degree from The Hong Kong University and Science and Technology, after which he was a postdoctoral researcher at the GRASP Lab, University of Pennsylvania. His research interests include 3D reconstruction and
scene understanding.
\end{IEEEbiography}

\begin{IEEEbiography}[{\includegraphics[width=1in,height=1.25in,clip,keepaspectratio]{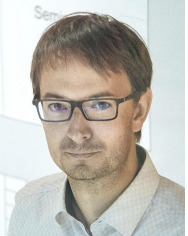}}]{Andreas Geiger}
received his Diploma in computer science and his Ph.D. degree from Karlsruhe Institute of Technology in 2008 and 2013. Currently, he is leading the Autonomous Vision Group at the University of Tubingen. He is also a core faculty member of the Tübingen AI Center. His research interests include computer vision, machine learning and scene understanding with a focus on self-driving vehicles.
\end{IEEEbiography}

\begin{IEEEbiography}[{\includegraphics[width=1in,height=1.25in,clip,keepaspectratio]{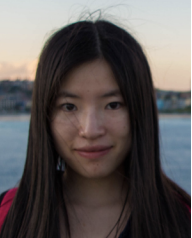}}]{Yiyi Liao}
is an assistant professor at Zhejiang University. Before that, she was a Postdoc at Autonomous Vision Group at the University of Tübingen and the MPI for Intelligent Systems. She received her Ph.D. in Control Science and Engineering from Zhejiang University in June 2018 and her B.S. degree from Xi’an Jiaotong University in 2013. Her research interests include 3D vision and scene understanding.
\end{IEEEbiography}

\clearpage

\setcounter{section}{0}
\setcounter{table}{0}
\setcounter{figure}{0}
\setcounter{equation}{0}

\renewcommand\theequation{E\arabic{equation}}
\renewcommand{\thetable}{R\arabic{table}}
\renewcommand\thefigure{S\arabic{figure}}

\section*{Appendix}

\begin{appendices}

\boldparagraph{Overview}
In this supplementary document, we first give a detailed overview of our network architecture, sampling strategy, far-class fusing strategy, evaluation metrics, and training and inference procedure in \secref{implementation}. Next, we describe our data preparation process in \secref{datapre}, including the stereo depth maps for supervision, LIDAR depth maps, semantic labels for evaluation, and \md{Acquirement of Weak 3D\&2D Labels}. Then, we provide additional experiments, including quantitative evaluation of bbox intersection, more label transfer results, novel label synthesis at 360$^\circ$ outward rotated viewpoints and panoramic viewpoints, qualitative comparison of label transfer and neural scene representations, weak depth supervision, analysis of 3D-2D CRF, and attempt in generating self-distilled pseudo instance GT using SAM in \secref{additionalexper}. Finally, we provide failure cases, including far-region label synthesis and fisheye geometric reconstruction in \secref{failure}.

\section{Implementation Details} \label{implementation}

\subsection{Network Architecture} \label{sec:network_architecture}
\begin{figure}[!h]
\centerline{\includegraphics[width=.49\textwidth]{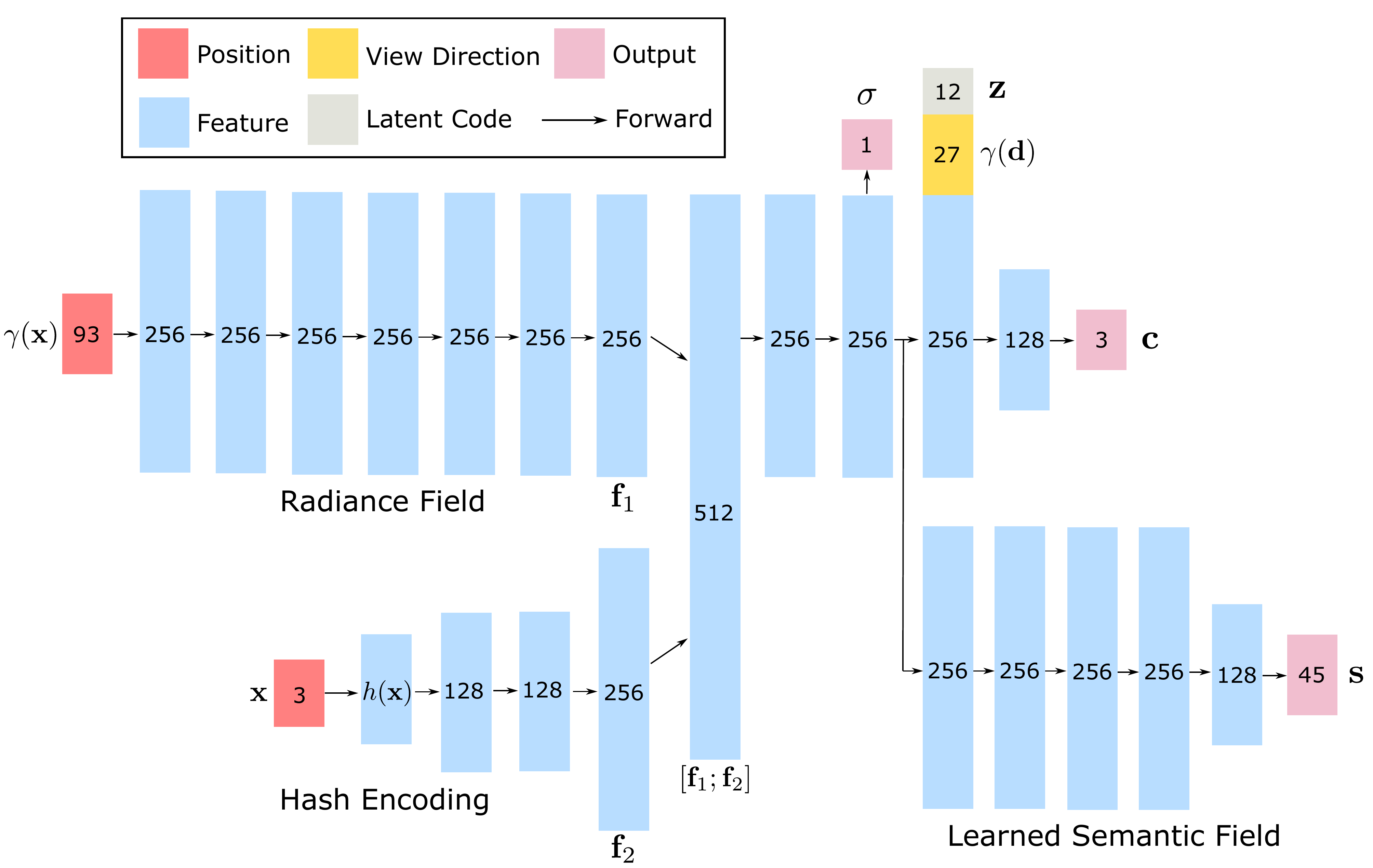}}
\caption{\textbf{Network Architecture.} PanopticNeRF-360 takes as input the 3D location $\bx$ (each element normalized to $[-1,1]$), the viewing direction $\bd$, and outputs radiance $\bc$ and semantic logits $\bs$. Scene feature $\textbf{f}$ is the concatenation of MLP features $\textbf{f}_{1}$ and hash encoding features $\textbf{f}_{2}$. $\bz$ is the per-frame latent embeddings.
}
\label{fig:network_supp}
\end{figure}

\figref{fig:network_supp} shows the trainable part of our PanopticNeRF-360 model. We adopt the same network architecture in all experiments. 

For radiance field, we map $\bx$ to a higher dimensional space using positional encoding (PE): $\gamma(p)=(\sin (2^{0} \pi p), \cos (2^{0} \pi p), \cdots, \sin (2^{L-1} \pi p), \cos (2^{L-1} \pi p))$. Note that we also apply PE in $\bd$ to produce a view-dependent effect. To learn high-frequency components in unbounded outdoor environments, we set $L=15$ for $\gamma(\bx)$ and $L=4$ for $\gamma(\bd)$. 

For hash encoding, we use multi-resolution grids~\cite{muller2022instant}, with each grid cell vertice mapped to a hash entry: $h(\mathbf{x})=\left(\bigoplus_{i=1}^3 x_i \pi_i\right) \quad \bmod T$. Each hash entry stores a trainable feature. We set the number of levels $16$ and grid resolutions  $N_{min}=16 \sim N_{max}=524,288$ with hash table size $T=2^{19}$. We query features via trilinear interpolation and concatenate the features at all spatial resolutions as input to a shallow MLP.

Our learned semantic field  is conditioned only on the 3D location $\bx$ rather than the viewing direction $\bd$ in order to predict view-independent semantic logits. The logits are then transformed into categorical distributions through a softmax layer.

\boldparagraph{Biased Density Initialization} To accelerate convergence when sampling points from coarse annotated bounding primitives, we set the bias of the density layer to 0.2, irrespective of whether the primitives belong to the ``thing" or ``stuff" categories.

\subsection{Sampling Strategy}
We sample points within the bounding primitives to skip empty space. As our bounding primitives are convex\footnote{The cuboids and ellipsoid are both convex. The extruded 3D plane is convex in a local region.}, each ray intersects a bounding primitive exactly twice which determines the sampling interval.
For each camera ray, we sort all bounding primitives that the ray hits from near to far and save the intersections offline. To save storage and to speed up training, we keep the first 10 sorted bounding primitives as the rest are highly likely to be occluded.  If a camera ray intersects less than 10 bounding primitives, we additionally sample a set of points to model the sky in $[t_{max}, t_{max} + t_{int}]$, where $t_{max}$ denotes the distance from the origin to the furthest bounding primitive in the 360$^{\circ}$ scene and $t_{int}$ is a constant distance interval. 

\subsection{Fusing Far-region Class} \label{fuse_farregion_class}
Urban scenes often contain objects that are located in far regions, resulting in missing bounding boxes in the corresponding images and misclassification as ``sky'' class. For pixels that have been identified as belonging to foreground classes within the 2D pseudo semantic label map, but which do not intersect bounding primitives along the corresponding ray, we replace these with available pseudo labels.

\subsection{Training and Inference}
As mentioned in Section 4.4 of the main paper, our total loss function comprises six terms, including three semantic losses $\cL^{2D}_{\hat{\bS}}$, $\cL^{2D}_{\bS}$, $\cL^{3D}_{\bs}$, the instance loss $\cL^{\text{2D}}_{\hat{\bT}}$, the photometric loss $\cL_{\bc}$ and the weak depth loss $\cL_{d}$. 
During per-scene optimization, we resize perspective images to 704$\times$188 pixels and fisheye images to 350$\times$350 pixels. We further mask out invalid regions in the fisheye, including the periphery region and the ego-vehicle. The photometric loss $\cL_{\bc}$ is defined on the stereo images and two-side fisheye images.  
We apply the 2D semantic losses $\cL^{2D}_{\hat{\bS}}$, $\cL^{2D}_{\bS}$ to the left perspective images and fisheye images and the 3D semantic loss $\cL^{3D}_{\bs}$ directly on 3D points sampled along the camera rays of these images. The instance loss $\cL^{\text{2D}}_{\hat{\bT}}$ is applied to buildings on stereo perspective views as line regression is not applicable on distorted fisheye label maps. The weak depth loss $\cL_d$ is defined only on the left images as the information gain is marginal on the right views and monocular depth estimation on fisheye views is struggling.

For inference, we compare our method to the baselines on the left perspective views and two-side fisheye views of which the manually labeled 2D Ground Truth is defined. Note that our method is not constrained to the fixed viewpoints during inference. We show label transfer results on the 360$^{\circ}$ outward rotated views in \secref{sec:360_outward} and panoramic views in \secref{sec:panoramic}.

\subsection{Evaluation Metric}
We evaluate mIoU and pixel accuracy following standard practice~\cite{cordts2016cityscapes,liao2021kitti}. Here, we provide more details of the multi-view consistency and panoptic quality metrics.

\boldparagraph{Multi-view Consistency} To evaluate multi-view consistency, we use depth maps obtained from LiDAR points to retrieve matching pixels across two consecutive frames. A similar multi-view consistency metric is considered in~\cite{Tong2021IJCAI} where optical flow is used to find  corresponding pixel pairs. We instead use LiDAR depth maps as they are more accurate compared to optical flow estimations. The details of generating the LiDAR depth maps will be introduced in \secref{sec:lidar_depth}. Given LiDAR depth maps at two consecutive test frames, we first unproject them into 3D space and find matching points. Two LiDAR points are considered matched if their distance in 3D is smaller than 0.1 meters. For each pair of matched points, we retrieve the corresponding 2D semantic labels and evaluate their consistency. The MC metric is evaluated as the number of consistent pairs over all matched pairs. Despite being not 100$\%$ accurate as the 3D points may not match exactly in 3D space, we find this metric meaningful in reflecting multi-view consistency. 

\boldparagraph{Panoptic Quality}
Following~\cite{kirillov2019panoptic}, we use the PQ metric to evaluate the performance of panoptic segmentation. PQ can be seen as the multiplication of a segmentation quality (SQ) term and a recognition quality (RQ) term. To mitigate the over-penalization errors related to stuff classes in PQ, we further involves PQ†~\cite{porzi2019seamless} for comprehensive evaluation. As the ground truth panoptic labels are not precise in distant areas and have a lot of small noises of things, we set ground truth labels of areas less than 100 pixels to ``void''. Correspondingly, segment matching will not be performed in void regions. In addition, Panoptic maps of the 3D-2D CRF and our method contain very small-region objects that are usually less than 100 pixels. To avoid being biased by those extremely small-region objects in the segment matching, we omit them by setting the predicted labels of the areas less than 100 pixels to the sky class.  To ensure a fair comparison across all methods, we adopt the same evaluation protocol for all baselines and our method.

\section{Data Preparation} \label{datapre}
\subsection{Stereo Depth for Weak Depth Supervision}
\begin{figure*}[tb]
 \centering
 \newcommand{\mywidth}{0.33 \textwidth}
 \setlength\tabcolsep{0.2em}
 \begin{tabular}{ccc}
      \includegraphics[width=\mywidth]{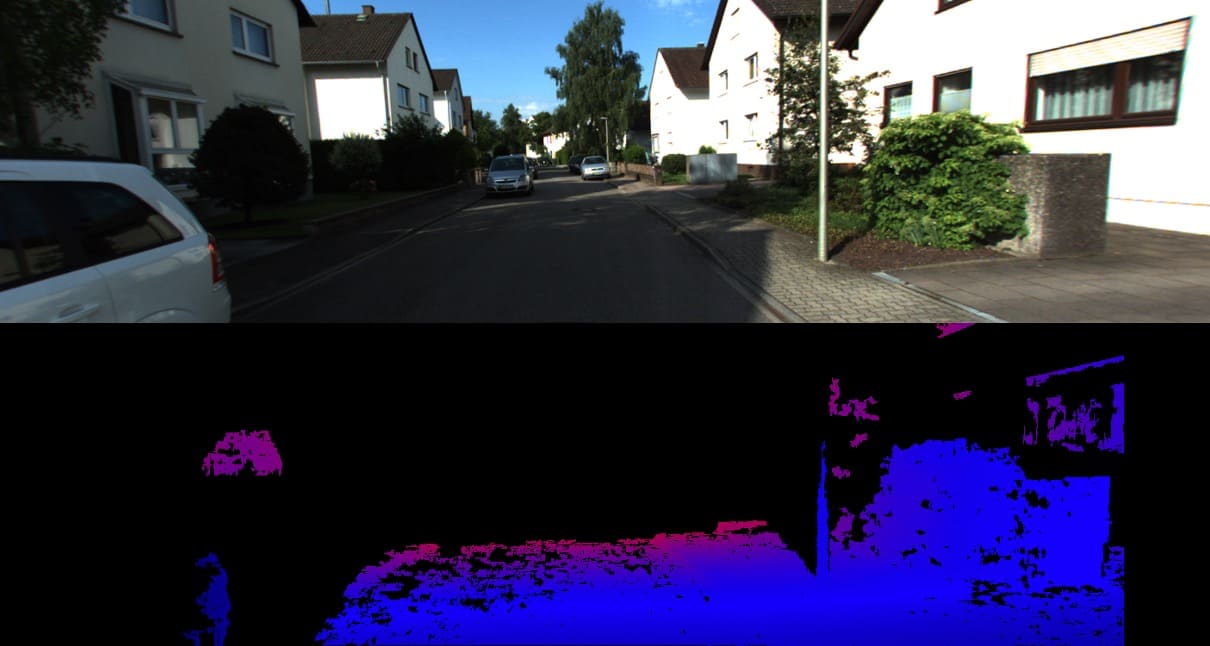}&
  \includegraphics[width=\mywidth]{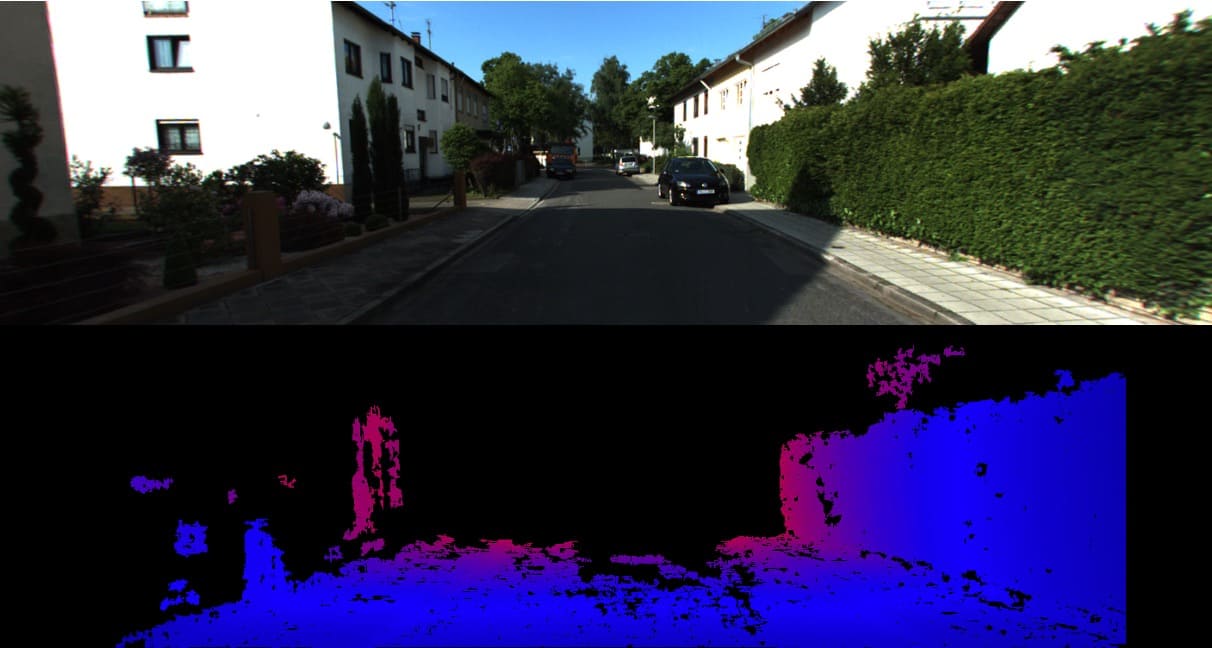}&       \includegraphics[width=\mywidth]{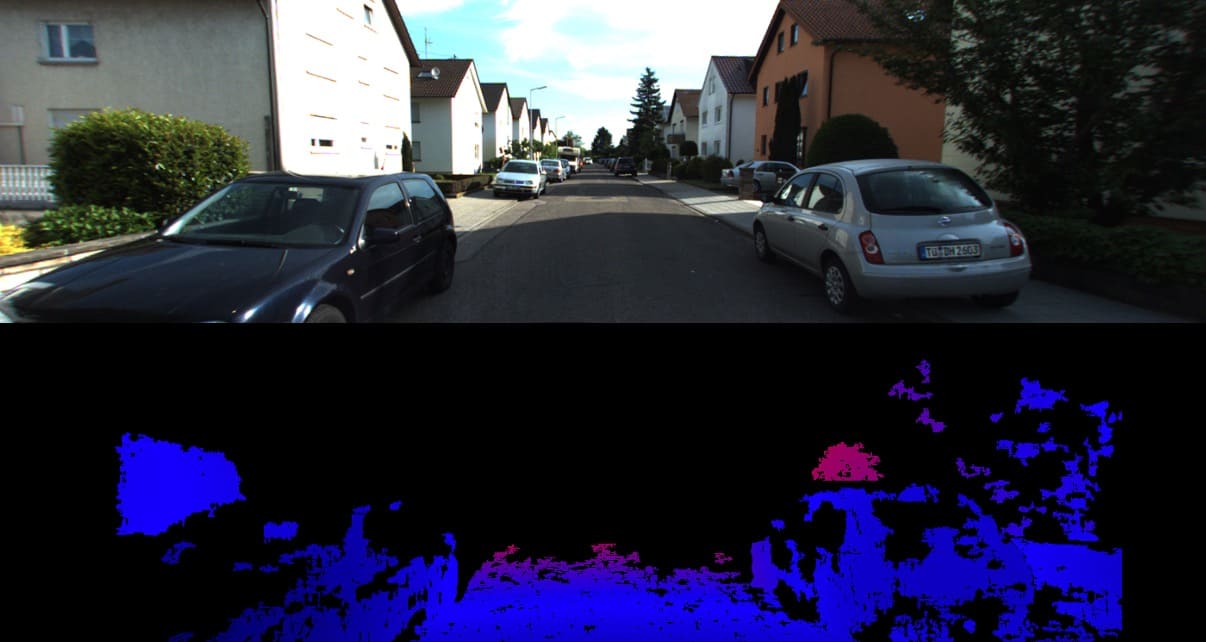}\\
   \includegraphics[width=\mywidth]{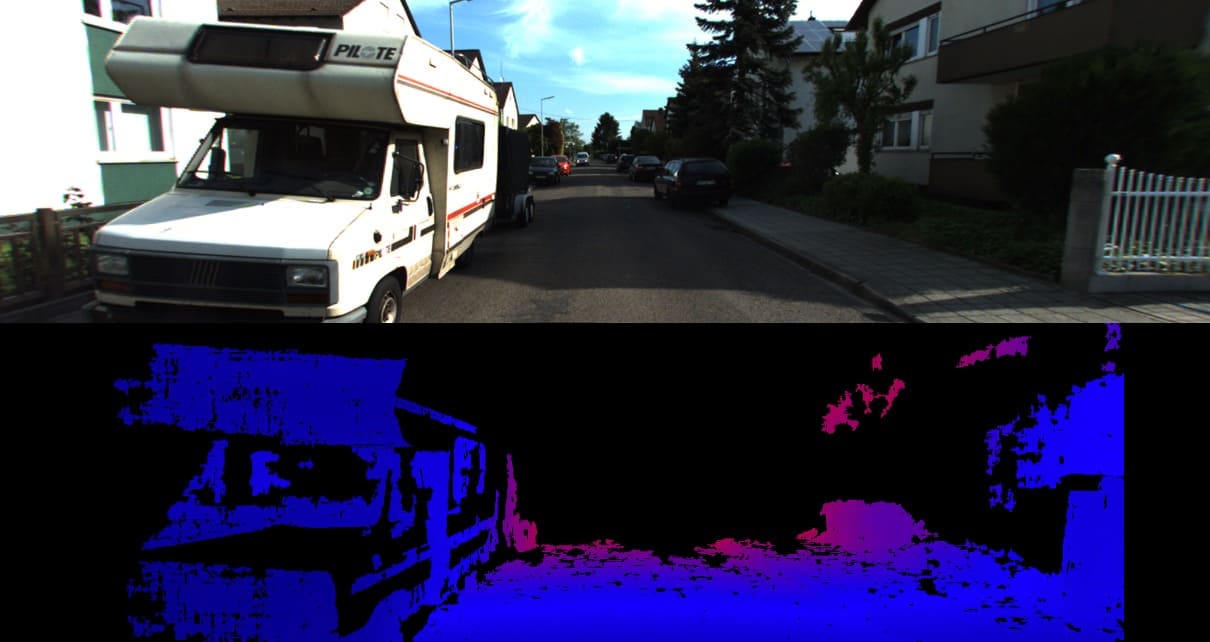}&
  \includegraphics[width=\mywidth]{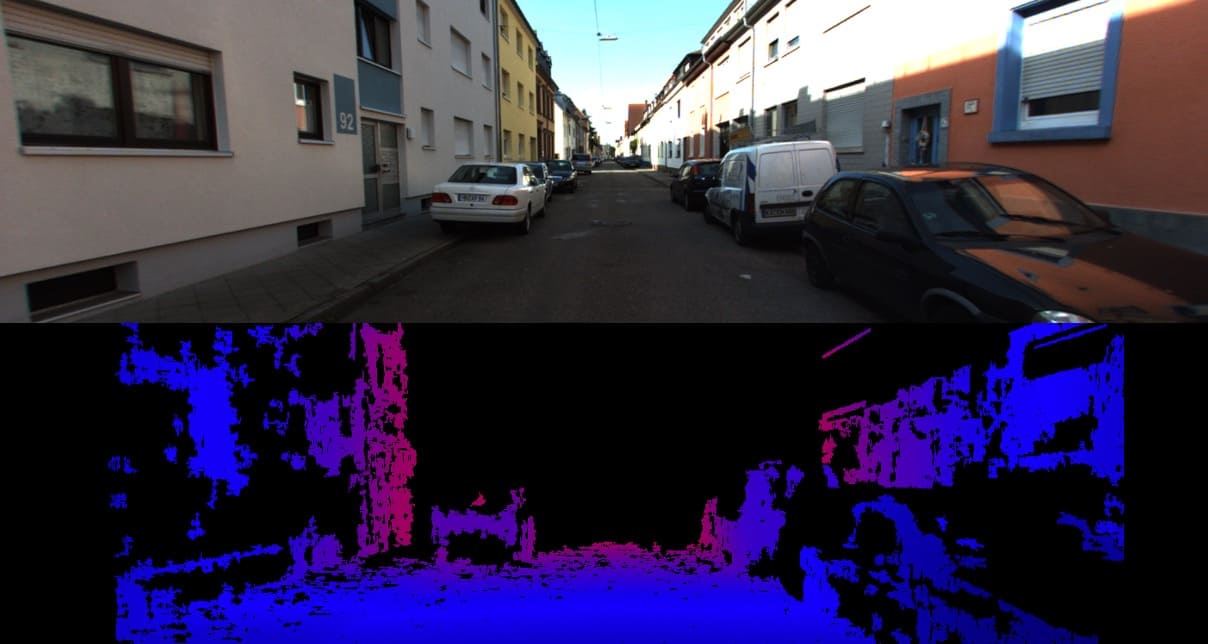}&       \includegraphics[width=\mywidth]{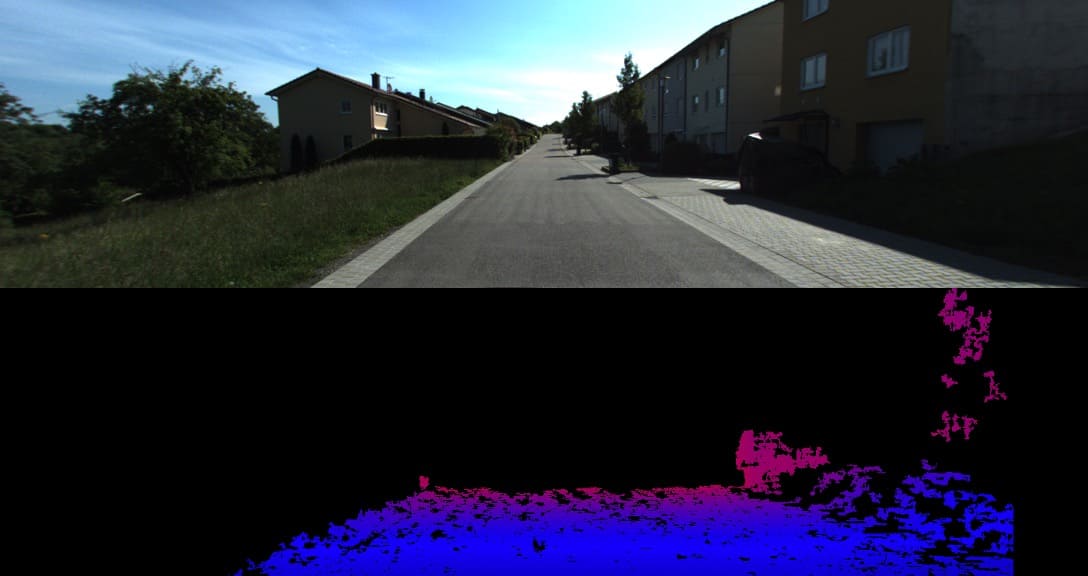}\\

 \end{tabular}
 \caption{
  \textbf{Depth Maps for Weak Depth Supervision.} Each group shows the RGB image (top) and the corresponding depth maps (bottom) used for supervision.}
 \label{fig:sgmdepth}

\end{figure*}

To provide weak depth supervision to PanopticNeRF-360, we use Semi-Global Matching (SGM)~\cite{hirschmuller2007stereo} to estimate depth given a stereo image pair on perspective views. We perform a left-right consistency check and a multi-frame consistency check in a window of $5$ consecutive frames to filter inconsistent predictions. We further omit depth predictions further than 15 meters in each frame as disparity is better estimated in nearby regions, see \figref{fig:sgmdepth}.

\subsection{LiDAR Depth for Evaluation} \label{sec:lidar_depth}
We evaluate the rendered depth maps against the LiDAR measurements. We refrain from using LiDAR as input as 1) this allows us to evaluate our depth prediction against LiDAR and 2) it makes our method more flexible to work with settings without any LiDAR observations. As LiDAR observations at each frame are sparse, we accumulate multiple frames of LiDAR observations and project the visible points to each frame similar to~\cite{uhrig2017sparsity}. 

\subsection{Manually Annotated 2D GT}
\begin{figure*}[tb]
 \centering
 \newcommand{\mywidth}{0.48\textwidth}
 \setlength\tabcolsep{0.8em}
 \begin{tabular}{cc}
  \includegraphics[width=\mywidth]{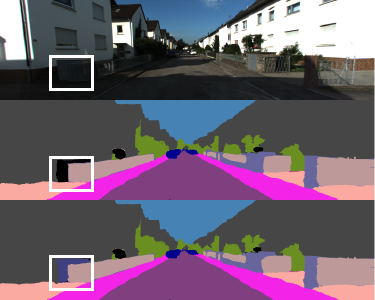}&
  \includegraphics[width=\mywidth]{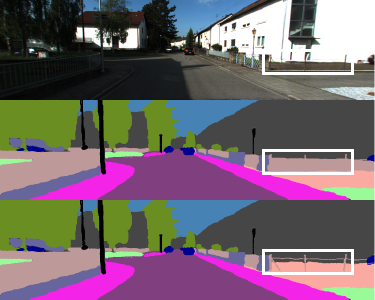}    

 \end{tabular}
 \caption{
  \textbf{Examples of Modified Ground Truth.} We correct some GT pixels that were incorrectly labeled in the KITTI-360 dataset. Top: Input RGB images. Middle: Original ground truth. Bottom: Modified ground truth. In the first column, we add the ``box" class. In the second column, we correct the ``parking" area.
 }
 
 \label{fig:modifiedgt}
\end{figure*}

The manually annotated 2D ground truth of KITTI-360~\cite{liao2021kitti} is inferior at some regions. For a fair comparison, we improve the label quality by manually relabeling ambiguous classes, see \figref{fig:modifiedgt} for illustrations.

\md{\subsection{Acquirement of Weak 3D\&2D Labels}}
\md{
We use the bounding primitives provided by the KITTI-360. As for the acquirement of these 3D labels, 
Liao et al.~\cite{liao2021kitti} first capture 3D point clouds utilizing LiDAR and stereo sensors. Next, using the KITTI-360 annotation toolkit\footnote{\href{https://github.com/autonomousvision/kitti360labeltool}{https://github.com/autonomousvision/kitti360labeltool}}, the 3D point clouds are annotated in the form of bounding primitives, \textit{i.e.}, by placing cuboids and ellipsoids to enclose objects in 3D and assigning a semantic label to each of them. 
The 3D scene is annotated with 37 label classes, including 24 ``thing'' classes and 13 ``stuff'' classes. Labels are defined in accordance with the Cityscapes. To obtain more 3D bounding primitives in other scenarios, we may also utilize tools like the KITTI-360 annotation toolkit, as demonstrated in the experiments on Waymo. Besides, it is promising to leverage off-the-shelf 3D understanding algorithms~\cite{qi2021offboard,he2023msf,shi2019pointrcnn,zhang2023hyperspherical} to reduce the cost of labeling.  We believe improving the labeling efficiency augmented by these 3D perception methods is an interesting yet orthogonal direction for future work. As for coarse 2D semantic segmentation, state-of-the-art models have intensively investigated semantic segmentation of self-driving scenarios~\cite{kirillov2023segment,cheng2022masked,chen2017deeplab,zhao2017pyramid,tao2020hierarchical}. Applying these models to obtain coarse 2D semantic segmentation masks is cost-effective.}

\section{Additional Experimental Results} \label{additionalexper}

\subsection{Quantitative Evaluation of Bbox Intersection}
\begin{table}[!tb]
\centering
\resizebox{.48\textwidth}{!} {

\begin{tabular}{c||cc}
\toprule
                & Number            & Volume ($\text{m}^{3}$)       \\
\midrule 
 \multicolumn{3}{c}{Cross-semantics Statistic}\\ 
\midrule 
sem.-sem.   & 140,229 (82.4\%) & 2,557,030.10 (94.2\%) \\
sem.-inst.  & 27,345 (16.1\%)  & 112,805.18 (4.2\%)    \\
inst.-inst. & 2,561 (1.5\%)    & 43,313.79 (1.6\%)   \\
\midrule 
 \multicolumn{3}{c}{Distribution of Volume ($\text{m}^{3}$)}\\ 
\midrule
0-1               & 100,864 (59.3\%) & 23,028.01 (0.8\%)\\
1-5               & 35,498 (20.9\%)  & 85,393.67 (3.1\%)\\
5-10              & 11,604 (6.8\%)   & 82,593.65 (3.0\%)\\
10-100            & 19,341 (11.4\%)  & 557,977.52 (20.6\%)\\
\textgreater{}100 & 2,828 (1.7\%)   & 1,964,156.23 (72.4\%)\\

\bottomrule
\end{tabular}
}
\caption{\textbf{Evaluation of Bbox Intersection.} `sem.-sem.' for semantic-semantic intersection, `sem.-inst.' for semantic-instance intersection, and `inst.-inst.' for instance-instance intersection, respectively.}

\label{tab:bbox_sect}
\end{table}

We provide quantitative bbox intersection evaluation in correspondence to Fig. 5 in the main paper. We evaluate on two sequences (``$2013\_05\_28\_drive\_0000\_sync$'' and ``$2013\_05\_28\_drive\_0004\_sync$'') that contain the 10 test scenes in KITTI-360. As shown in Table~\ref{tab:bbox_sect}, the intersection between different semantic class (``sem.-sem.'' and ``sem.-inst.'') accounts to 98.5$\%$ in number and 98.4$\%$ in volume. Thus the learned semantic field is crucial to resolve label ambiguity using pre-trained semantic prior transferred from other datasets. As the ``inst-inst'' intersection is small (1.5$\%$ in number and 1.6$\%$ in volume), we ignore resolving instance intersection in our experiments. From the distribution of volume, we find that the intersected volume size and quantity of intersection numbers are inversely proportional.

\subsection{More Panoptic Label Transfer Results} 
\begin{figure*}[!h]
 \centering
 \newcommand{\mywidth}{0.49 \textwidth}
 \setlength\tabcolsep{0.2em}
 \begin{tabular}{cc}
\includegraphics[width=\mywidth]{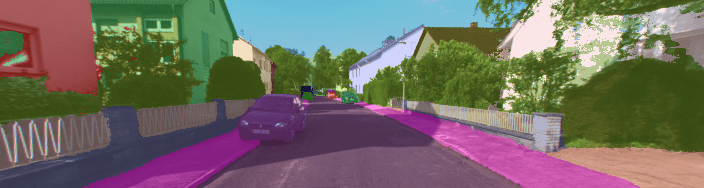}& 
\includegraphics[width=\mywidth]{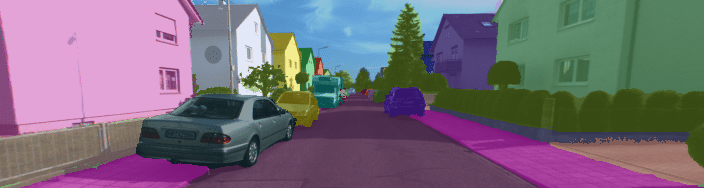} \\
\includegraphics[width=\mywidth]{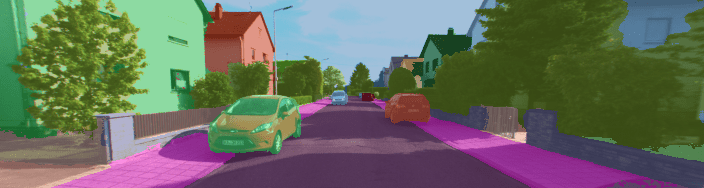}& 
\includegraphics[width=\mywidth]{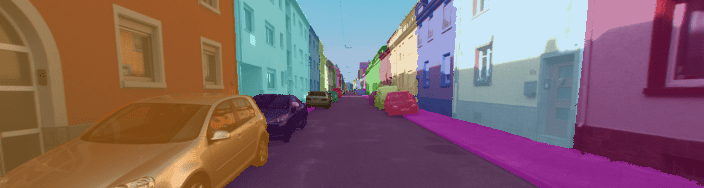} \\
\includegraphics[width=\mywidth]{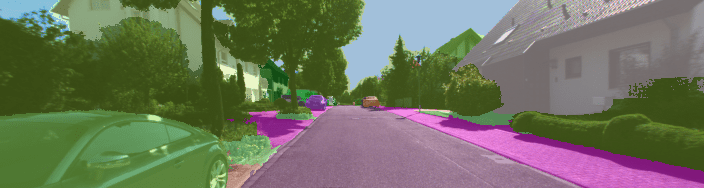}& 
\includegraphics[width=\mywidth]{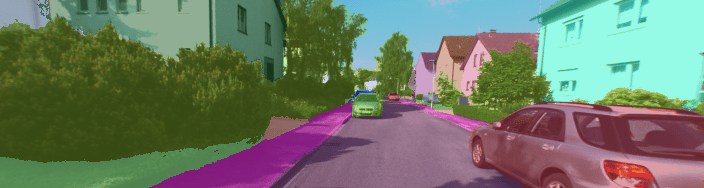} \\
\includegraphics[width=\mywidth]{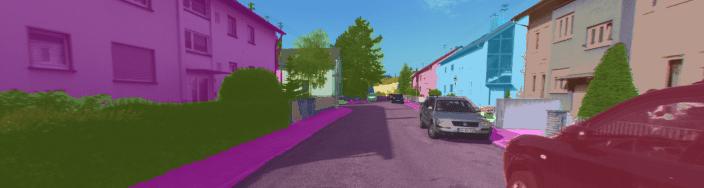}& 
\includegraphics[width=\mywidth]{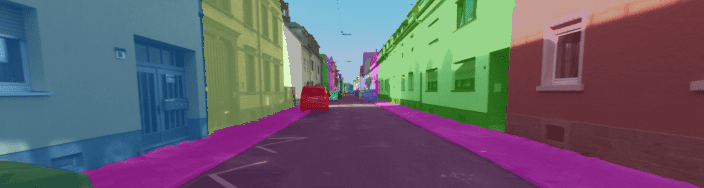} \\
\includegraphics[width=\mywidth]{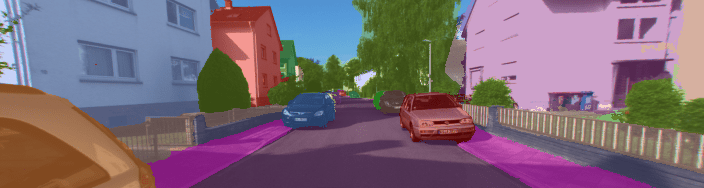}& 
\includegraphics[width=\mywidth]{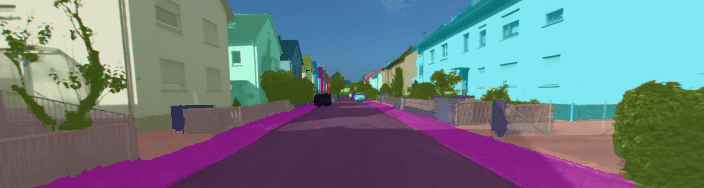} \\
\includegraphics[width=\mywidth]{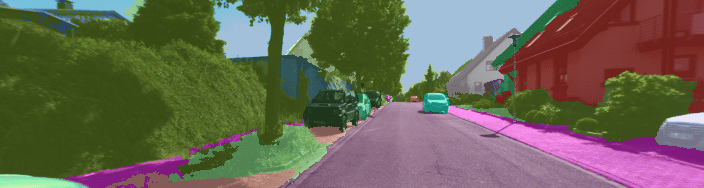}& 
\includegraphics[width=\mywidth]{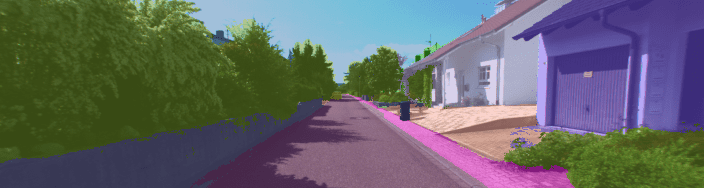} \\
 \end{tabular}
 \caption{\textbf{More Perspective Panoptic Label Transfer Results} overlaid with GT RGBs and predicted panoptic labels.}
 \label{fig:more_panoptic_perspective}

\end{figure*}

\begin{figure*}[!h]
 \centering
 \newcommand{\mywidth}{0.19 \textwidth}
 \setlength\tabcolsep{0.2em}
 \begin{tabular}{ccccc}
  \includegraphics[width=\mywidth]{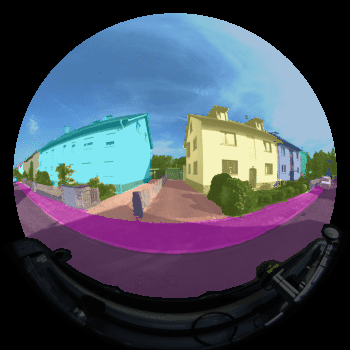}& 
  \includegraphics[width=\mywidth]{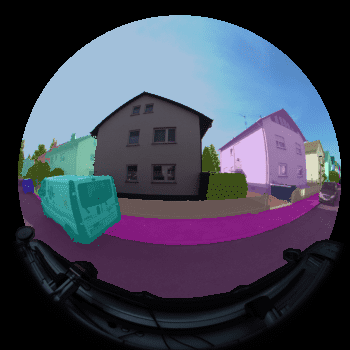}& 
  \includegraphics[width=\mywidth]{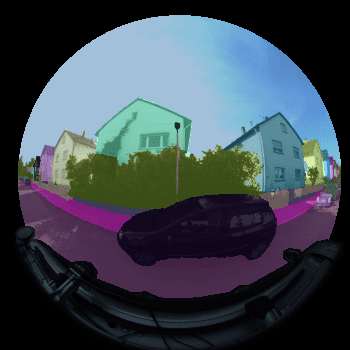}& 
  \includegraphics[width=\mywidth]{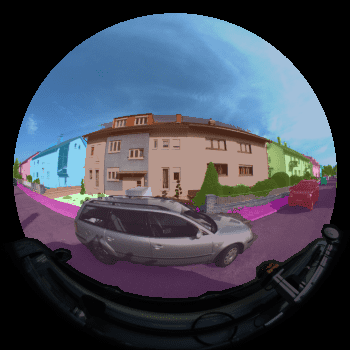}& 
  \includegraphics[width=\mywidth]{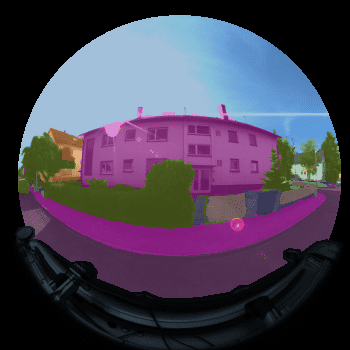}\\
  \includegraphics[width=\mywidth]{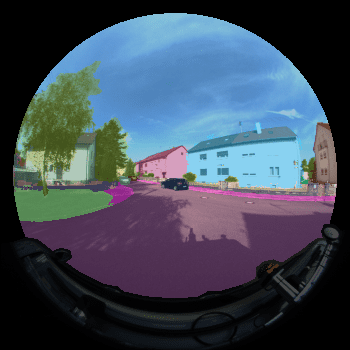}& 
  \includegraphics[width=\mywidth]{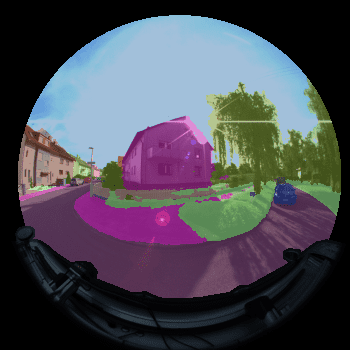}& 
  \includegraphics[width=\mywidth]{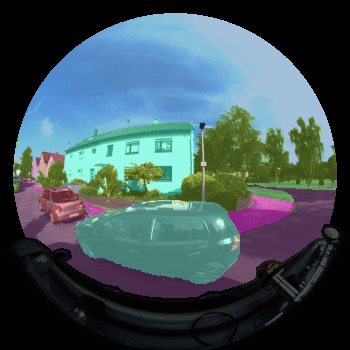}& 
  \includegraphics[width=\mywidth]{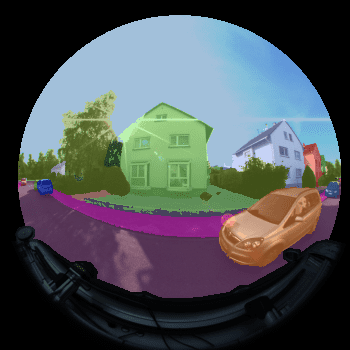}& 
  \includegraphics[width=\mywidth]{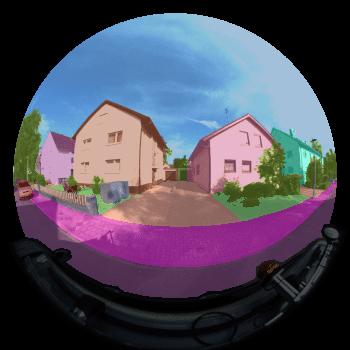}\\
  \includegraphics[width=\mywidth]{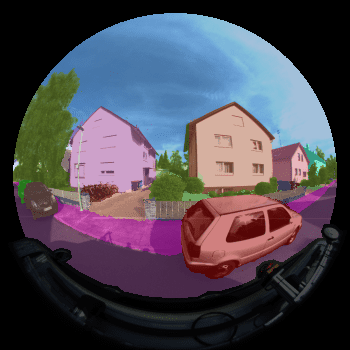}& 
  \includegraphics[width=\mywidth]{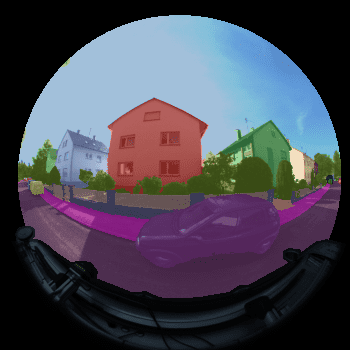}& 
  \includegraphics[width=\mywidth]{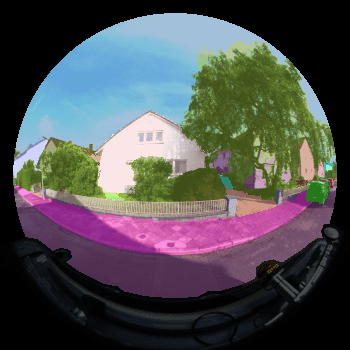}& 
  \includegraphics[width=\mywidth]{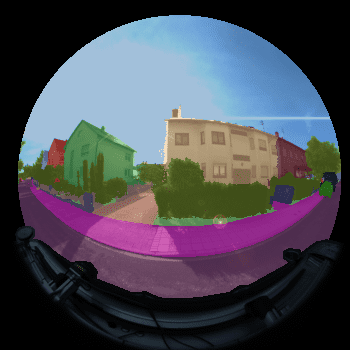}& 
  \includegraphics[width=\mywidth]{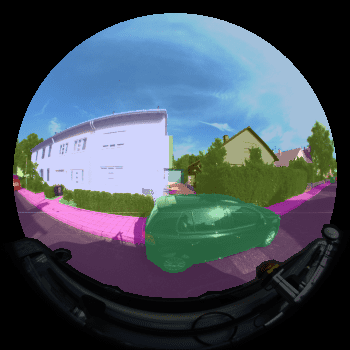}\\
  \includegraphics[width=\mywidth]{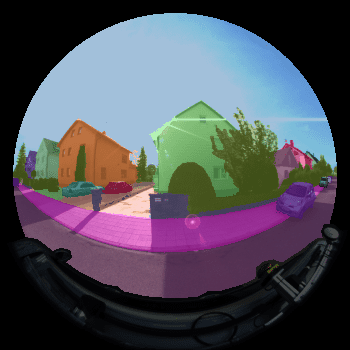}& 
  \includegraphics[width=\mywidth]{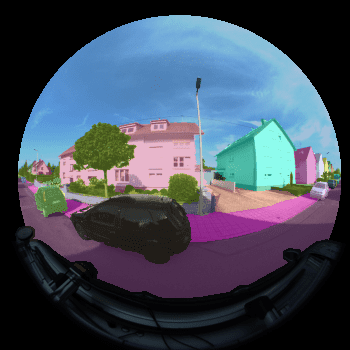}& 
  \includegraphics[width=\mywidth]{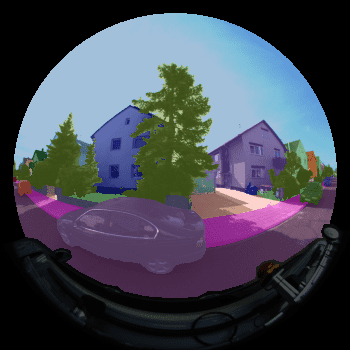}& 
  \includegraphics[width=\mywidth]{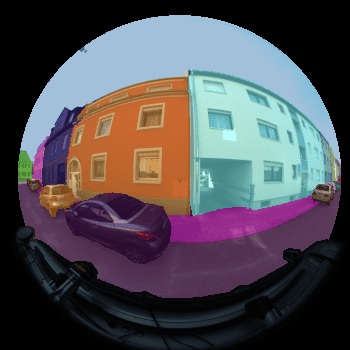}& 
  \includegraphics[width=\mywidth]{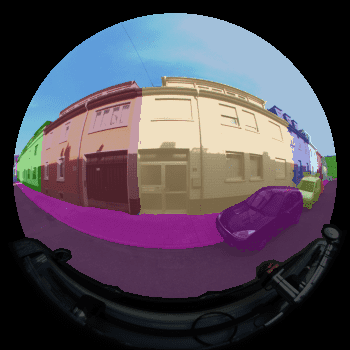}\\
  \includegraphics[width=\mywidth]{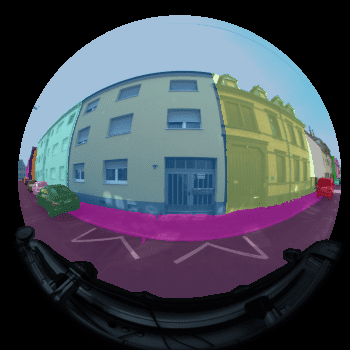}& 
  \includegraphics[width=\mywidth]{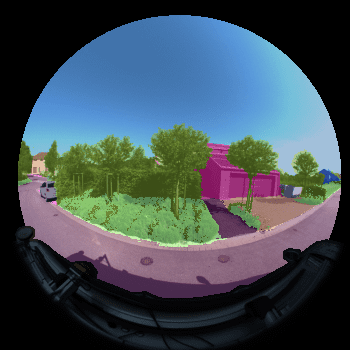}& 
  \includegraphics[width=\mywidth]{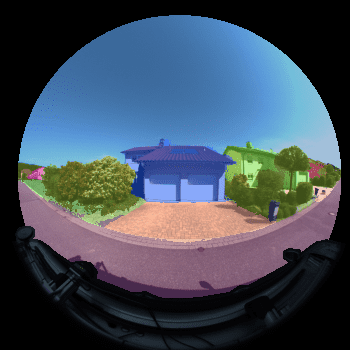}& 
  \includegraphics[width=\mywidth]{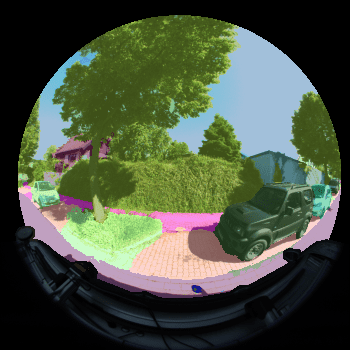}& 
  \includegraphics[width=\mywidth]{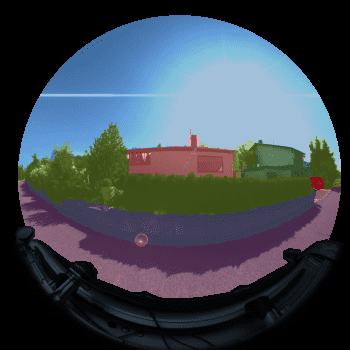}\\
 \end{tabular}
 \caption{\textbf{More Fisheye Panoptic Label Transfer Results} overlaid with GT RGBs and predicted panoptic labels.}
 \label{fig:more_panoptic_fisheye}

\end{figure*}

We provide visualization of more panoptic label transfer results both on perspective views~(see \figref{fig:more_panoptic_perspective}) and fisheye views~(see \figref{fig:more_panoptic_fisheye}).

\subsection{360$^{\circ}$ Outward Rotated Label Synthesis} \label{sec:360_outward}
\begin{figure*}[!h]
 \centering
 \newcommand{\mywidth}{0.33 \textwidth}
 \setlength\tabcolsep{0.2em}
 \begin{tabular}{ccc}
 \includegraphics[width=\mywidth]{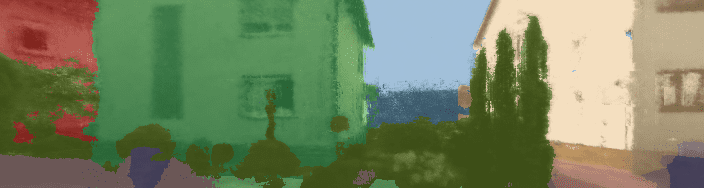}& 
 \includegraphics[width=\mywidth]{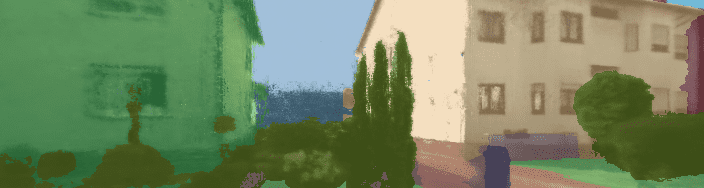}& 
 \includegraphics[width=\mywidth]{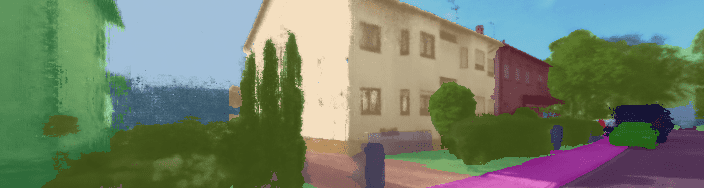} \\
 \includegraphics[width=\mywidth]{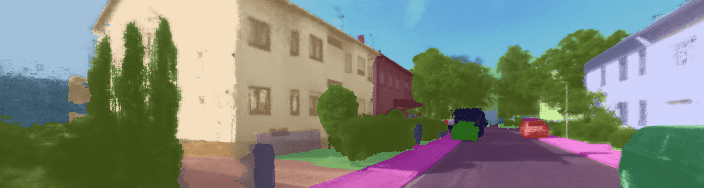}& 
 \includegraphics[width=\mywidth]{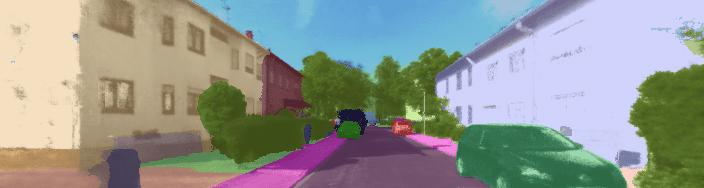}& 
 \includegraphics[width=\mywidth]{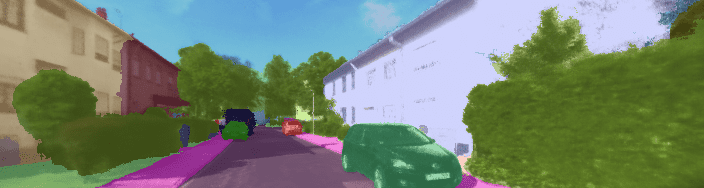} \\
 \includegraphics[width=\mywidth]{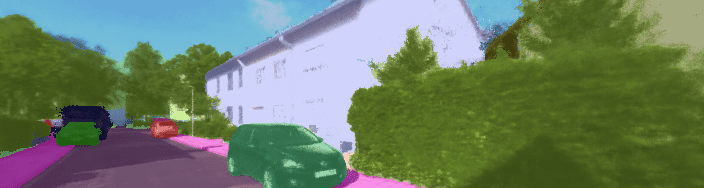}& 
 \includegraphics[width=\mywidth]{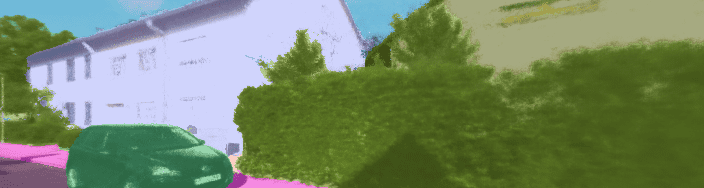}& 
 \includegraphics[width=\mywidth]{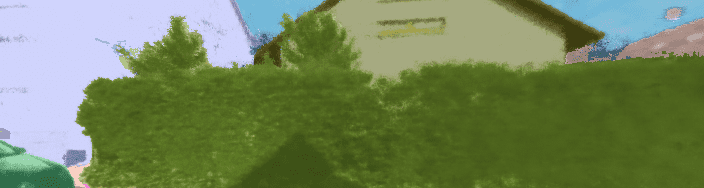} \\
 \includegraphics[width=\mywidth]{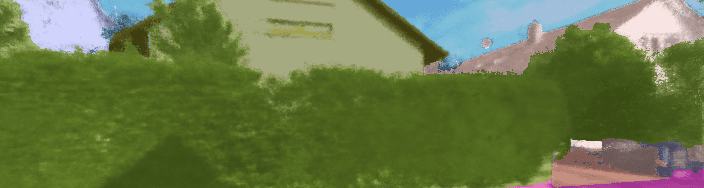}& 
 \includegraphics[width=\mywidth]{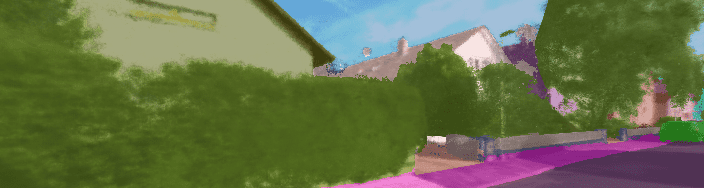}& 
 \includegraphics[width=\mywidth]{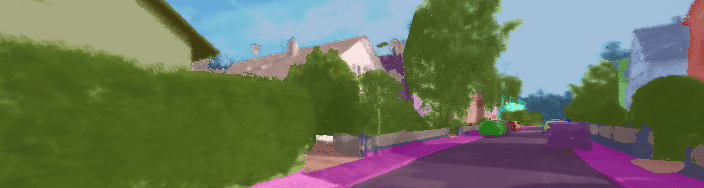} \\
 \includegraphics[width=\mywidth]{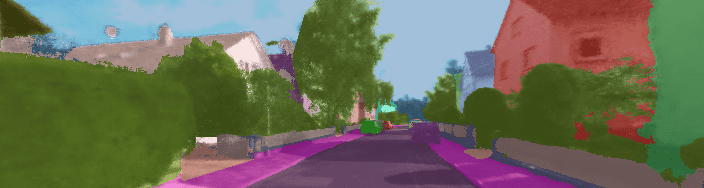}& 
 \includegraphics[width=\mywidth]{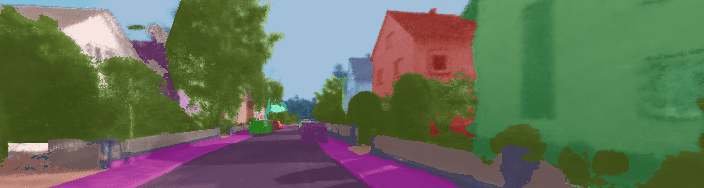}& 
 \includegraphics[width=\mywidth]{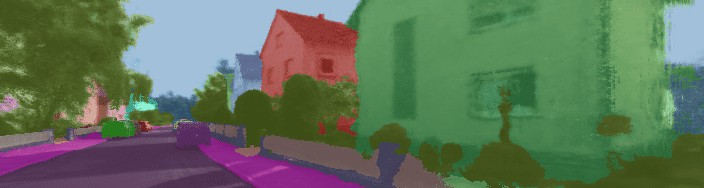} \\
 \\ 
 \\
 \\
 \includegraphics[width=\mywidth]{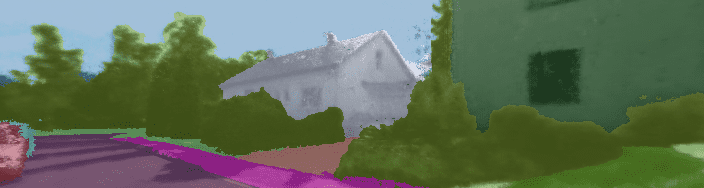}& 
 \includegraphics[width=\mywidth]{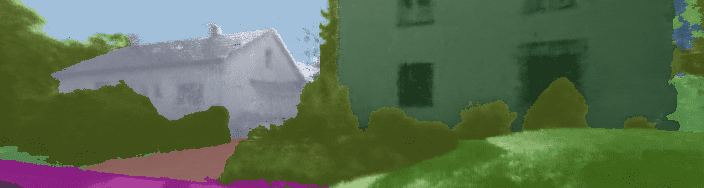}& 
 \includegraphics[width=\mywidth]{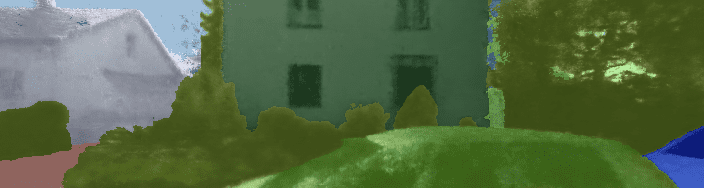} \\
 \includegraphics[width=\mywidth]{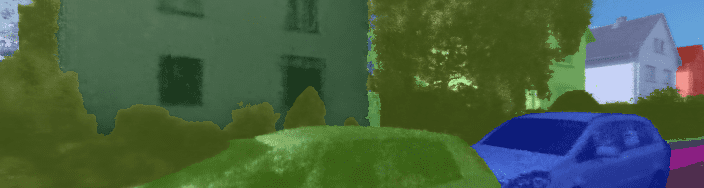}& 
 \includegraphics[width=\mywidth]{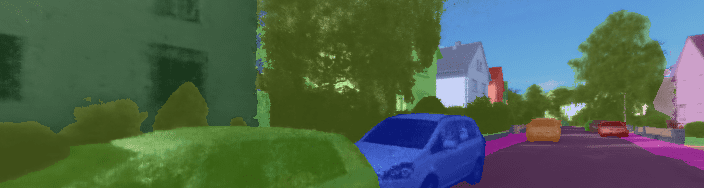}& 
 \includegraphics[width=\mywidth]{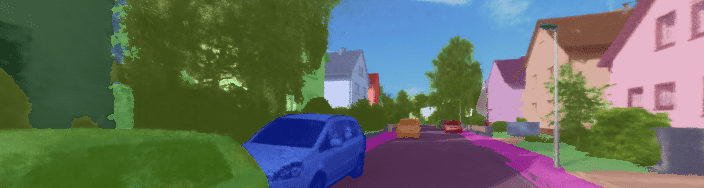} \\
 \includegraphics[width=\mywidth]{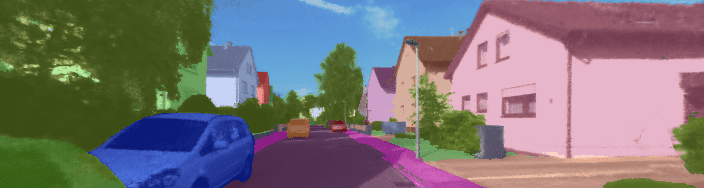}& 
 \includegraphics[width=\mywidth]{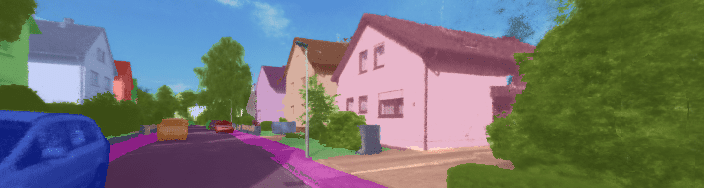}& 
 \includegraphics[width=\mywidth]{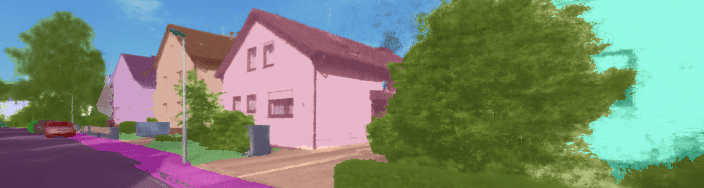} \\
 \includegraphics[width=\mywidth]{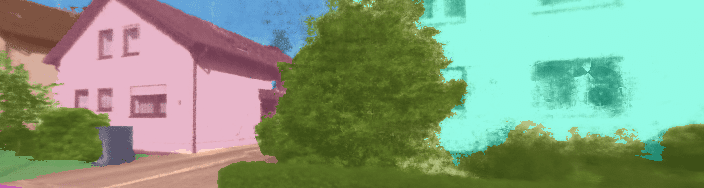}& 
 \includegraphics[width=\mywidth]{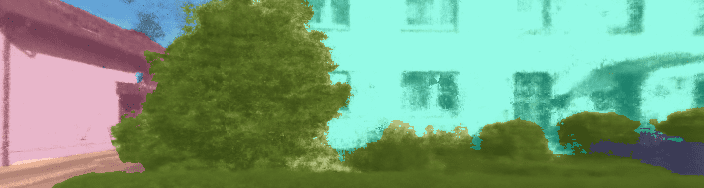}& 
 \includegraphics[width=\mywidth]{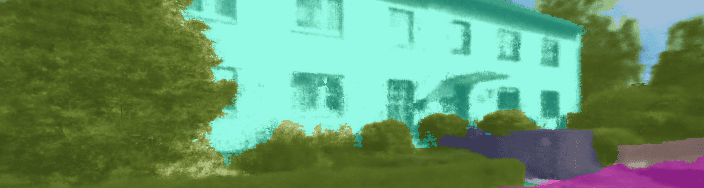} \\
 \includegraphics[width=\mywidth]{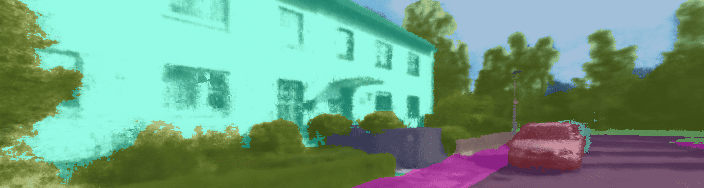}& 
 \includegraphics[width=\mywidth]{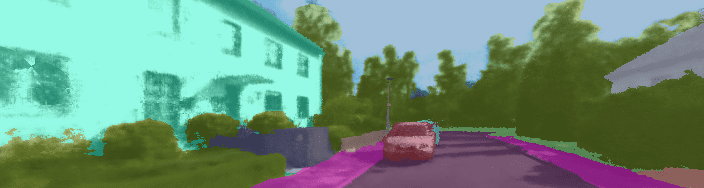}& 
 \includegraphics[width=\mywidth]{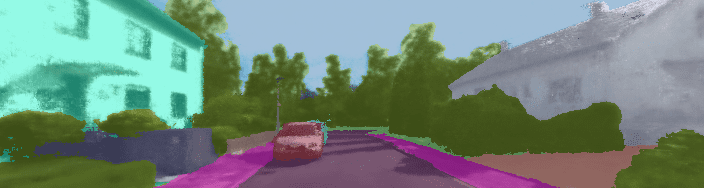} \\
 \end{tabular}
 \caption{\textbf{Panoptic Label Transfer Results on 360 Outward Rotated Viewpoints.} The frames are overlaid with predicted panoptic labels and rendered appearance. The rotation can be recognized in the image groups from left to right, from top to down.}
 \label{fig:360_outward_rotated}

\end{figure*}

PanopticNeRF-360 enables omnidirectional rendering of label and appearance. In \figref{fig:360_outward_rotated}, we showcase sampling at 360$^\circ$ rotated viewpoints around the z-axis in a scene. The angle between adjacent images is 24$^\circ$. Please refer to te website for videos with 64 frames.

\subsection{Panoramic Synthesis} \label{sec:panoramic}
\begin{figure*}[!h]
 \centering
 \newcommand{\mywidth}{0.47 \textwidth}
 \setlength\tabcolsep{0.2em}
 \begin{tabular}{cc}
\includegraphics[width=\mywidth]{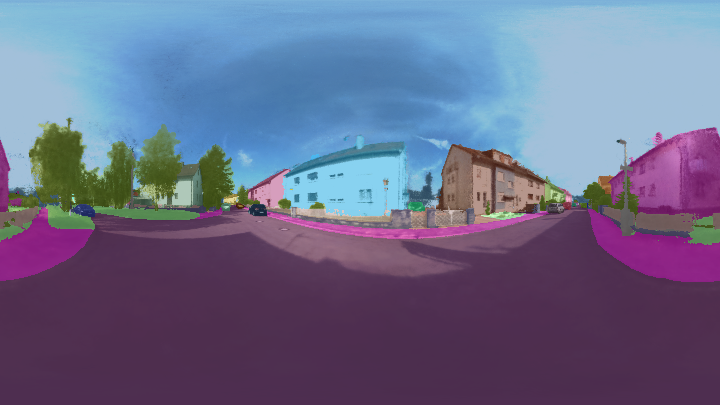}& 
\includegraphics[width=\mywidth]{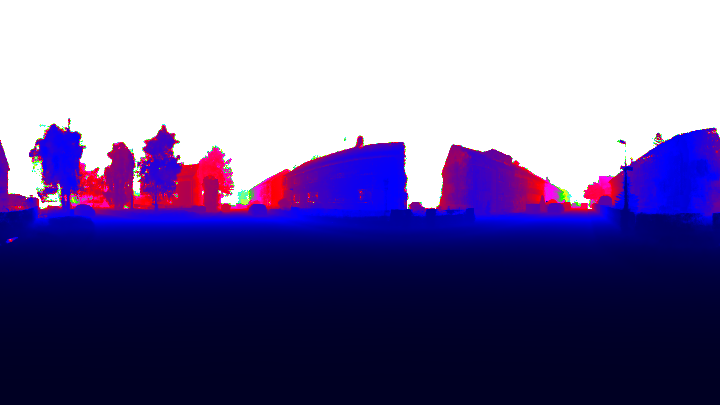}\\ 
\includegraphics[width=\mywidth]{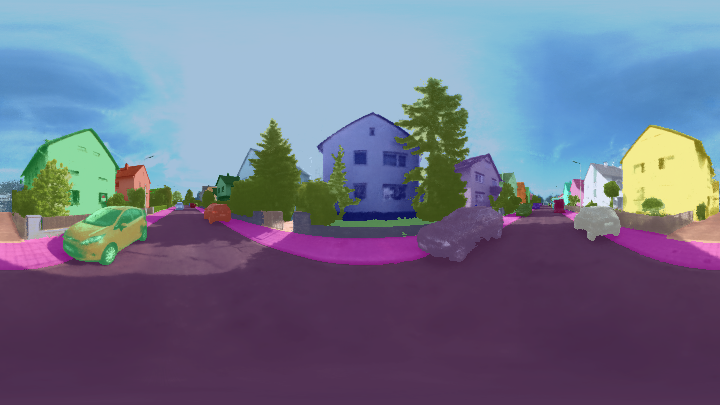}& 
\includegraphics[width=\mywidth]{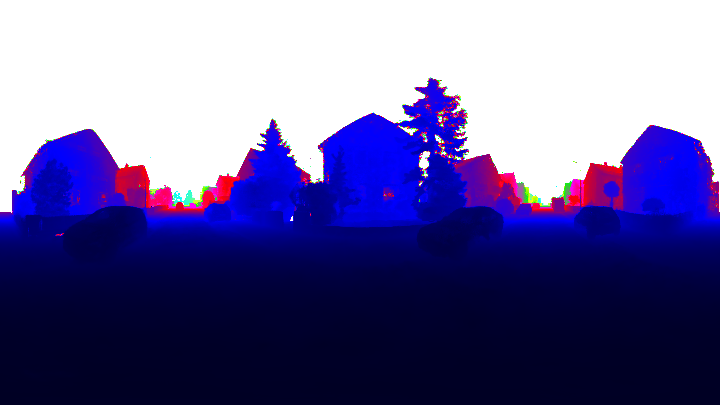}\\ 
\includegraphics[width=\mywidth]{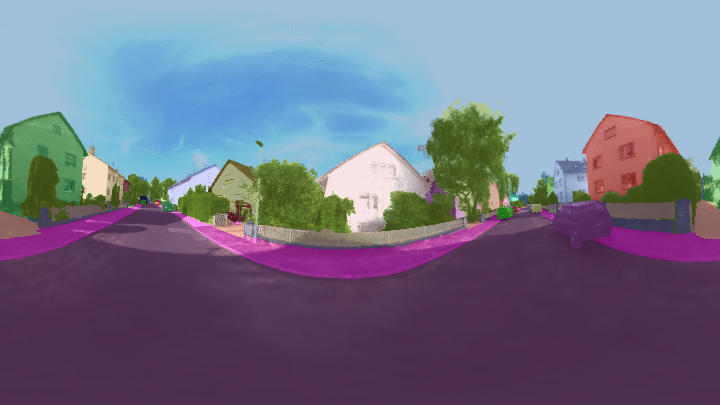}& 
\includegraphics[width=\mywidth]{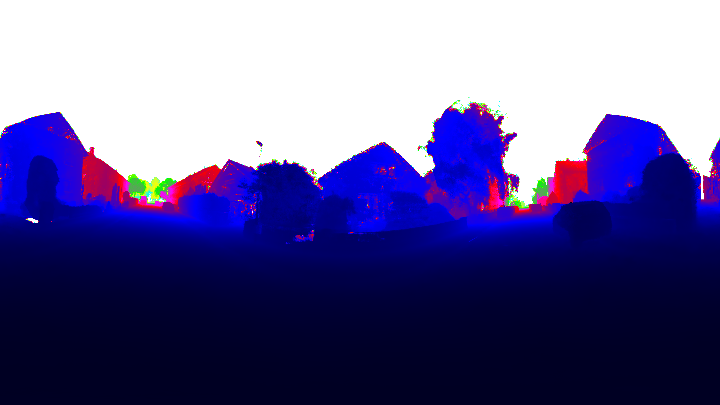}\\ 
\includegraphics[width=\mywidth]{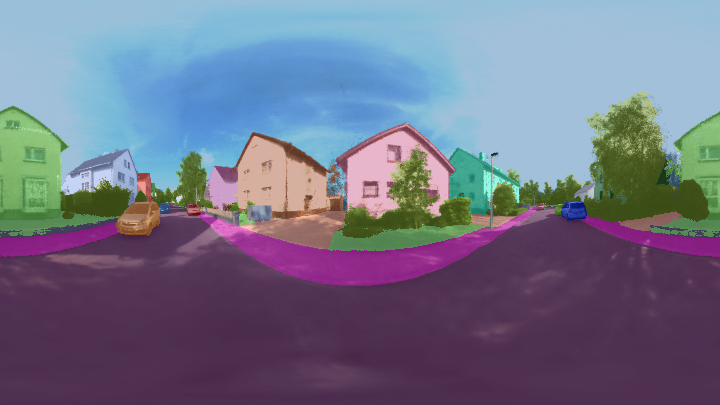}& 
\includegraphics[width=\mywidth]{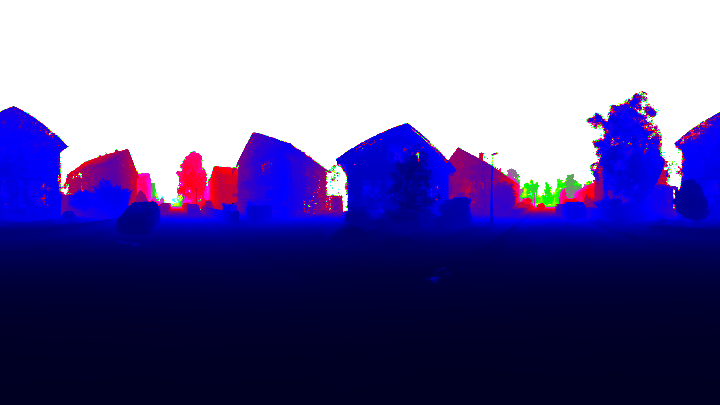}\\ 
\includegraphics[width=\mywidth]{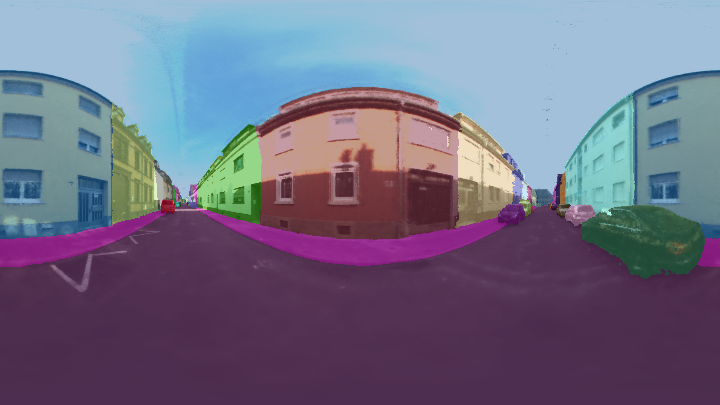}& 
\includegraphics[width=\mywidth]{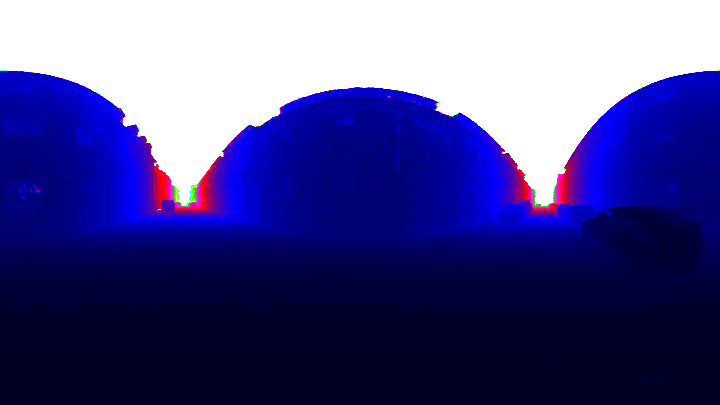}\\ 

 \end{tabular}
 \caption{\textbf{Panoramic Label $\&$ Depth Synthesis at 960$\times$540 pixels}.}
 \label{fig:more_panorama_label_depth}

\end{figure*}

In Fig. 11 of main paper, we show examples of panoramic semantic/instance labels at the resolution of 960$\times$540 pixels but crop the regions that are too high and too low (50 marginal pixels). In \figref{fig:more_panorama_label_depth}, we visualize panoramic panoptic label and depth map synthesis at full-resolution.

\subsection{Qualitative Comparison of Label Transfer}
\begin{figure*}[!t]
 \centering
 \newcommand{\mywidth}{0.87\textwidth}
 \setlength\tabcolsep{0.05em}
 \newcolumntype{P}[1]{>{\centering\arraybackslash}m{#1}}
 \def\arraystretch{0.50}

  \begin{tabular}{P{0.5em}P{0.5em}P{\mywidth}}
    \rot{\tiny{RGB}}&& \includegraphics[width=\mywidth]{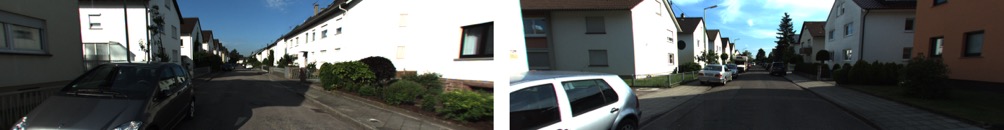}\\
   \rot{\tiny{FC CRF + Manual GT}}&& \includegraphics[width=\mywidth]{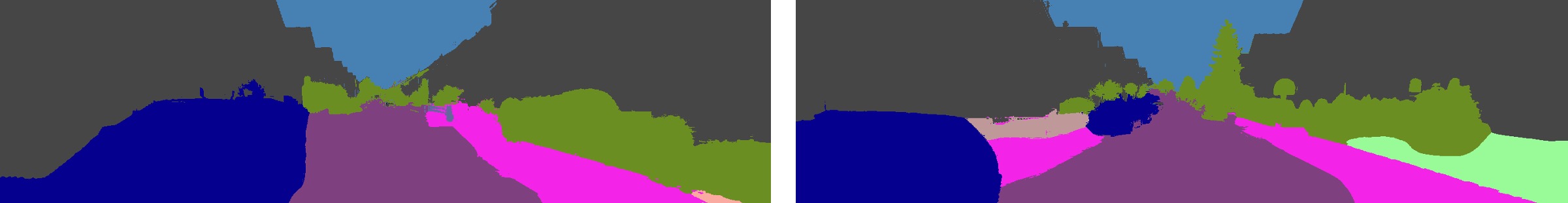}\\
   \rot{\tiny{S-NeRF + Manual GT}}&& \includegraphics[width=\mywidth]{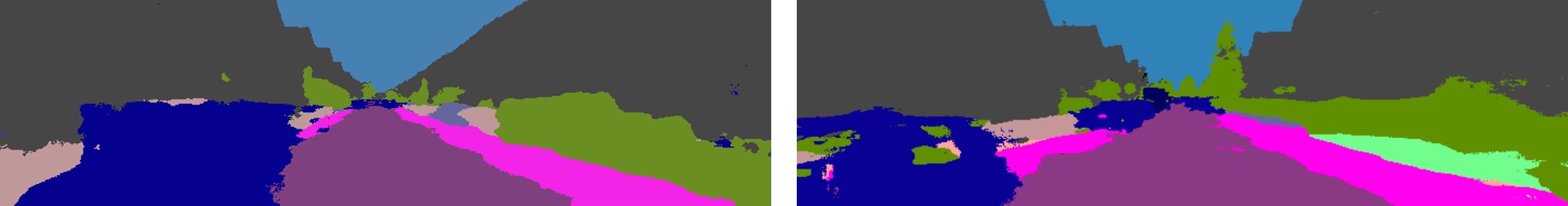}\\
   \rot{\tiny{S-NeRF + Pseudo GT}}&& \includegraphics[width=\mywidth]{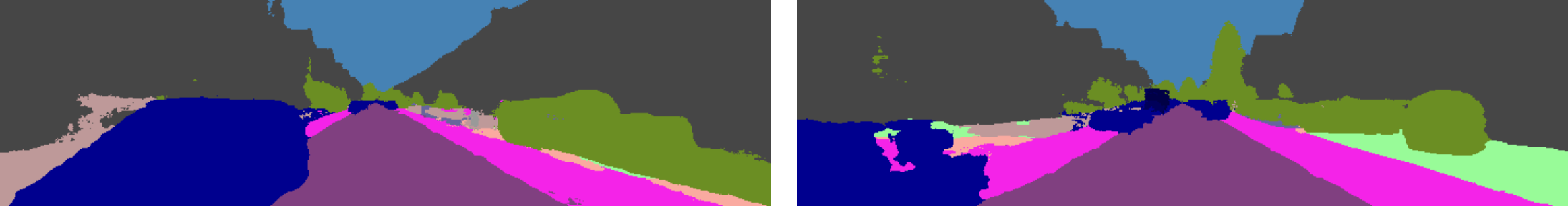}\\
   \rot{\tiny{J-NeRF + Pseudo GT}}&& \includegraphics[width=\mywidth]{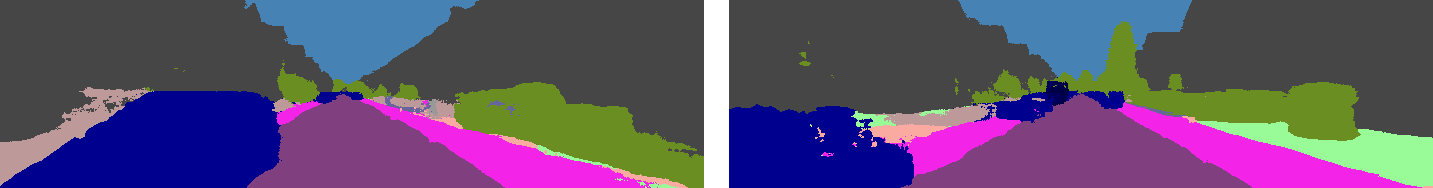}\\
   \rot{\tiny{PSPNet*}}&& \includegraphics[width=\mywidth]{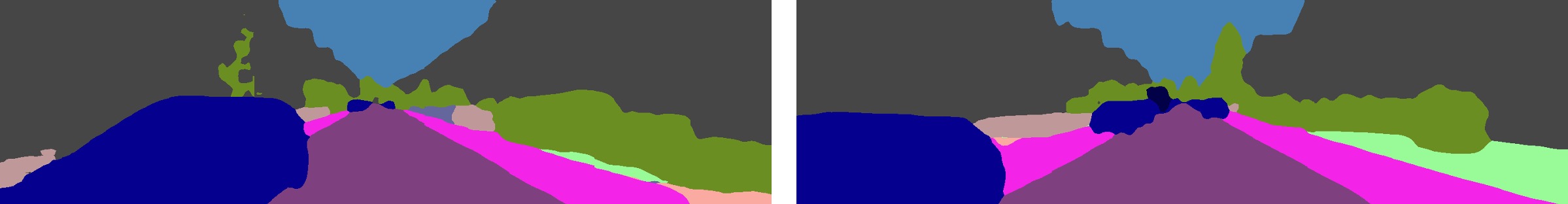}\\
   \rot{\tiny{3D Primitives + GC}}&& \includegraphics[width=\mywidth]{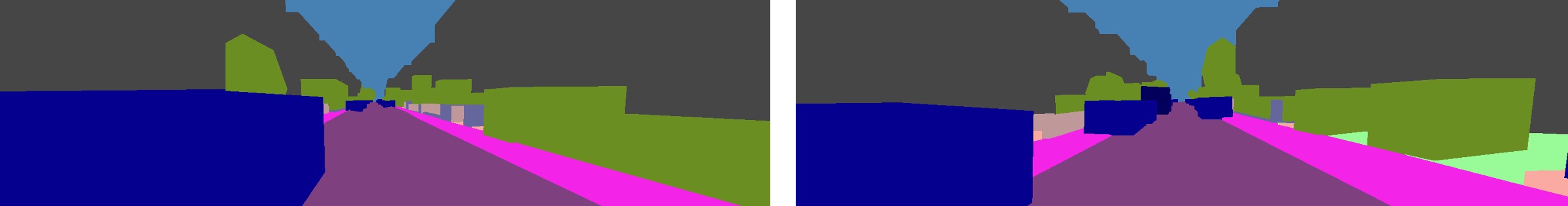}\\
   \rot{\tiny{3D Mesh + GC}}&& \includegraphics[width=\mywidth]{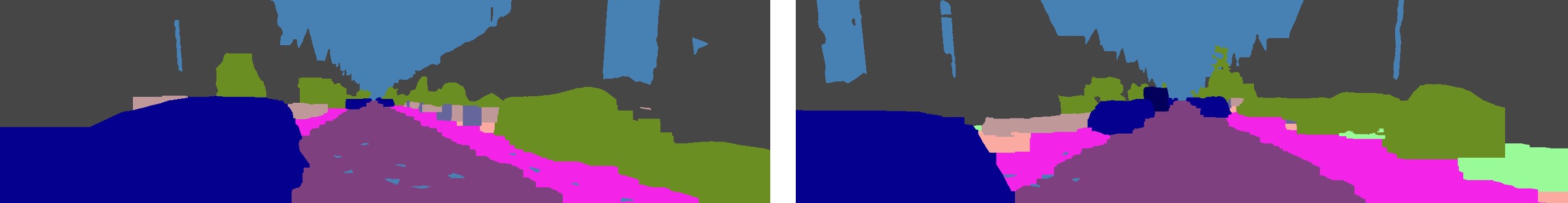}\\
   \rot{\tiny{3D Point + GC}}&& \includegraphics[width=\mywidth]{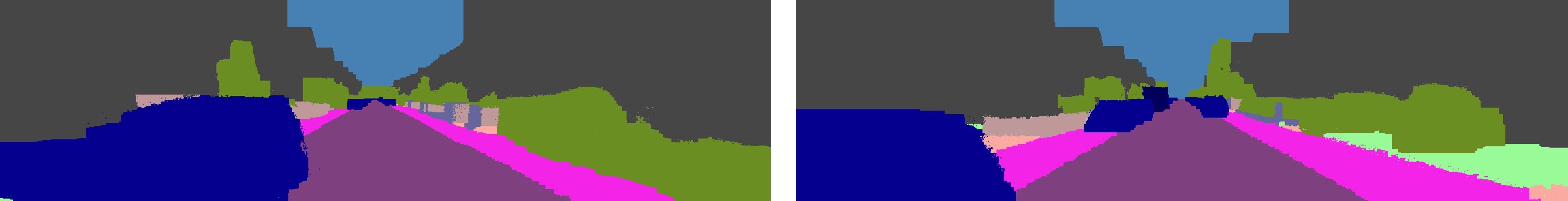}\\
   \rot{\tiny{3D-2D CRF}}&& \includegraphics[width=\mywidth]{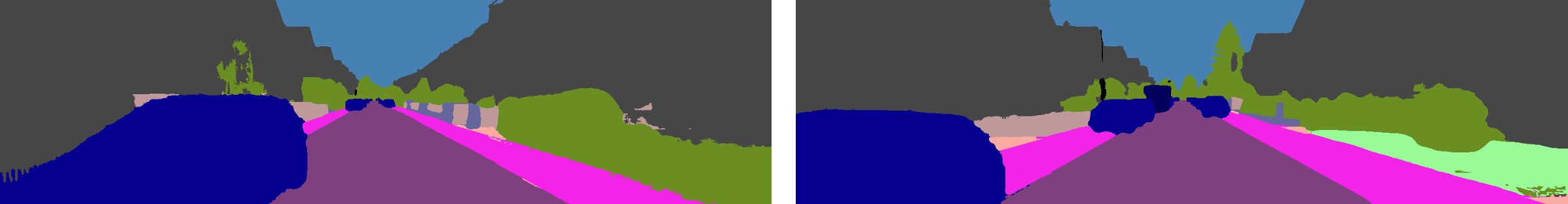}\\
   \rot{\tiny{Ours}}&& \includegraphics[width=\mywidth]{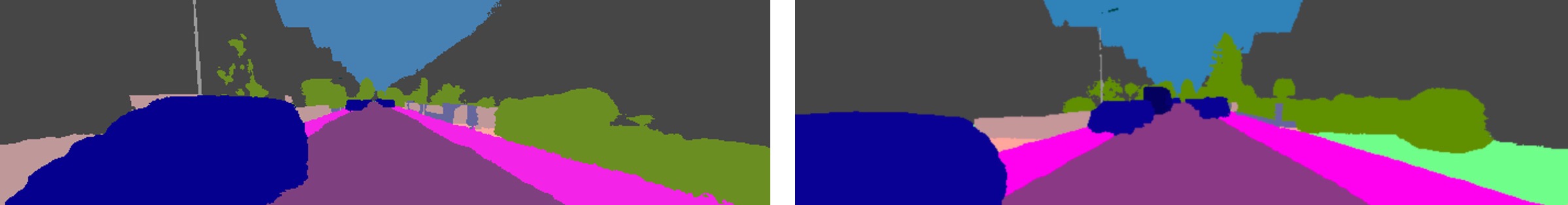}\\
   \rot{\tiny{GT}}&& \includegraphics[width=\mywidth]{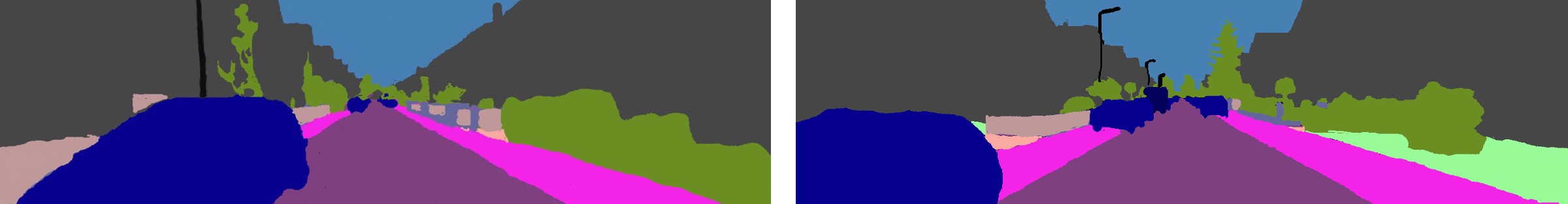}\\
 \end{tabular}
 \caption{
  \textbf{Qualitative Comparison of Perspective Semantic Label Transfer} on frames with manually labeled ground truth.
 }
 \label{fig:semantic_perspective_comparison}
\end{figure*}

\begin{figure*}[tb]
 \centering
 \newcommand{\mywidth}{0.49 \textwidth}
 \setlength\tabcolsep{0.2em}
 \begin{tabular}{cc}
  \includegraphics[width=\mywidth]{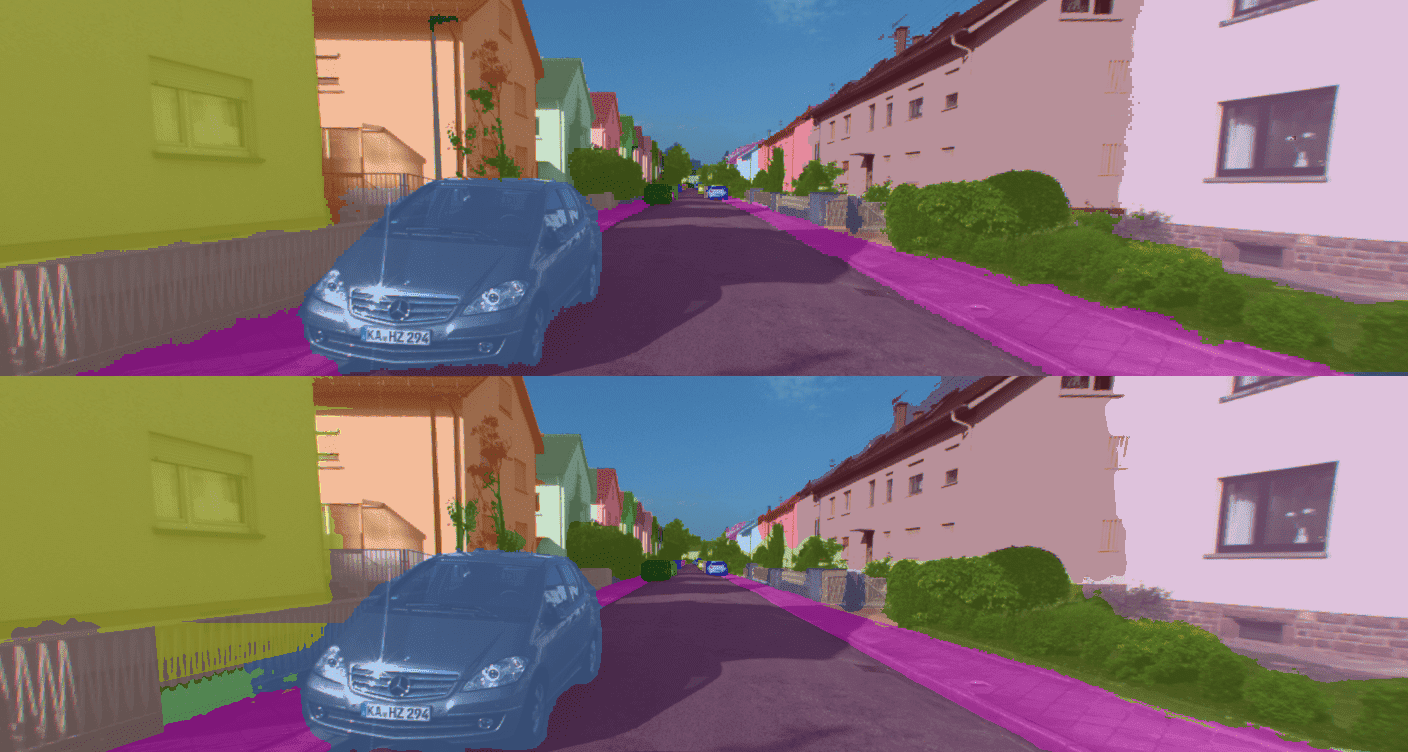}&
  \includegraphics[width=\mywidth]{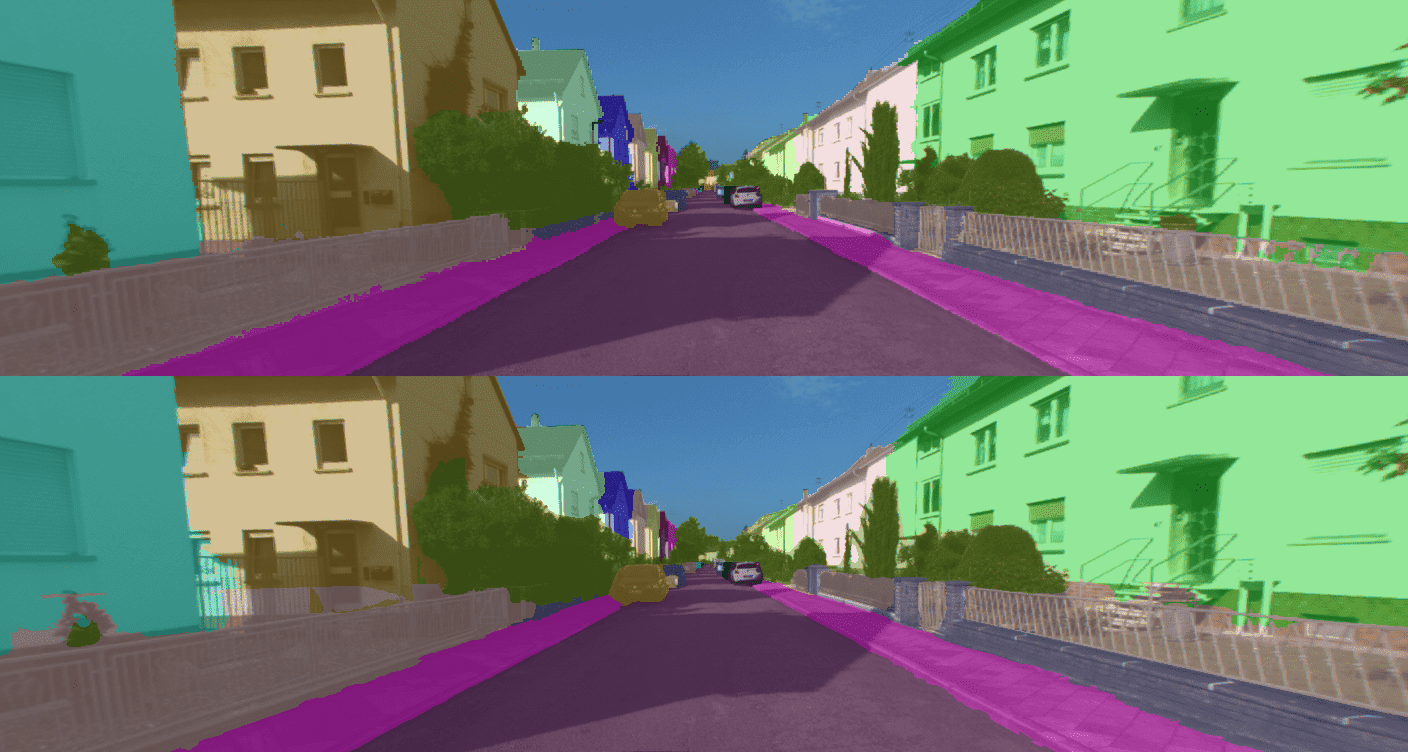}\\
  \includegraphics[width=\mywidth]{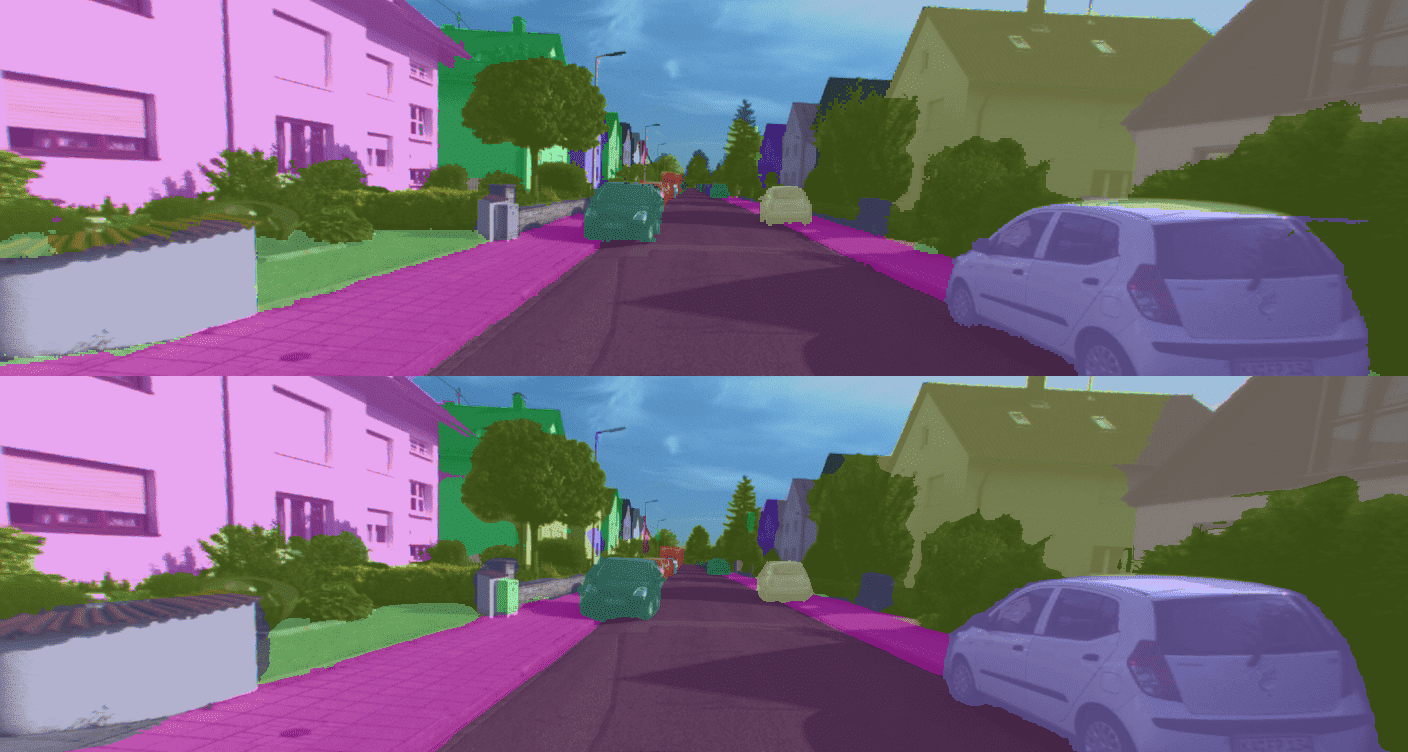}&
  \includegraphics[width=\mywidth]{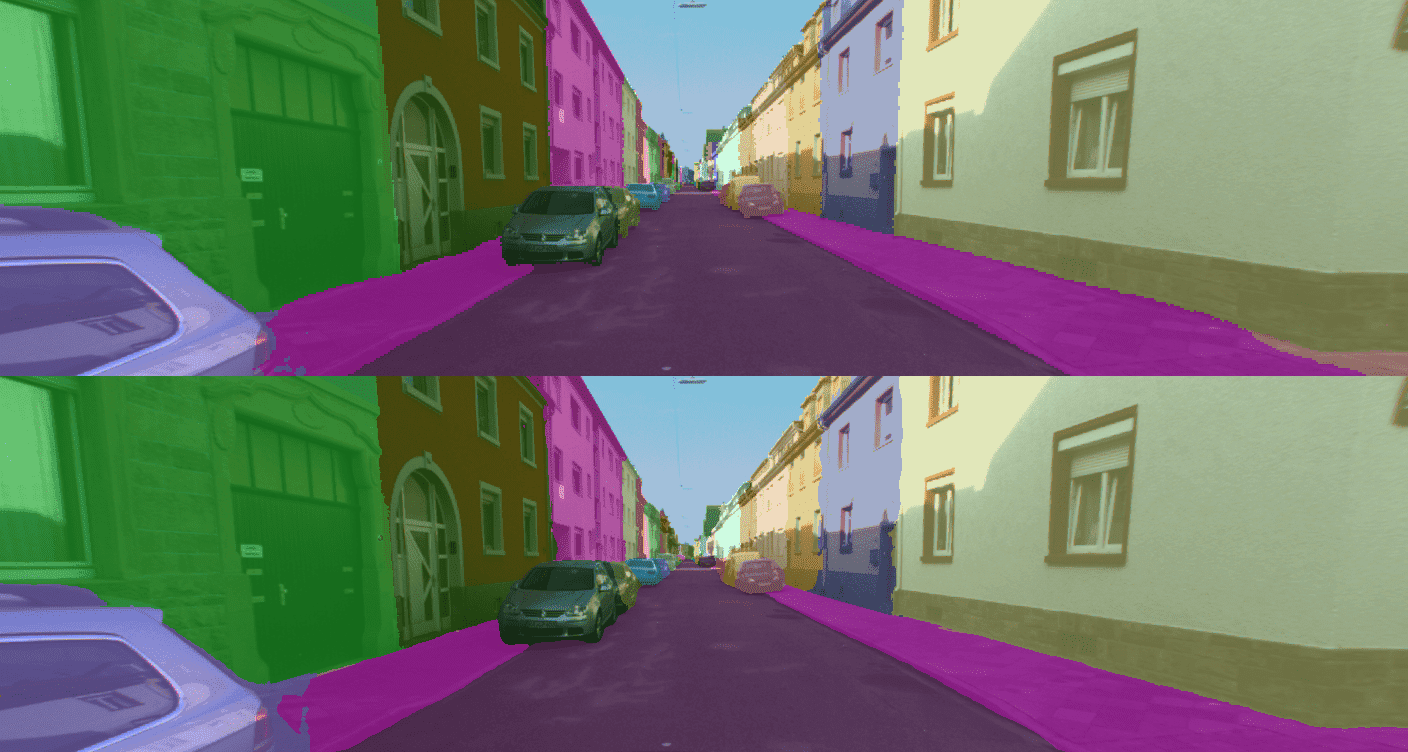}\\
 \end{tabular}
 \caption{
  \textbf{Qualitative Comparison of Perspective Panoptic Label Transfer on frames without manually labeled ground truth}. Each group shows the prediction of ours (top) and 3D-2D CRF~\cite{liao2021kitti} (bottom). 
  }
 \label{fig:panoptic_perspective_comparison}
\end{figure*}

\begin{figure*}[tb]
 \centering
 \newcommand{\mywidth}{0.195 \textwidth}
 \setlength\tabcolsep{0.2em}
 \begin{tabular}{ccccc}
  \includegraphics[width=\mywidth]{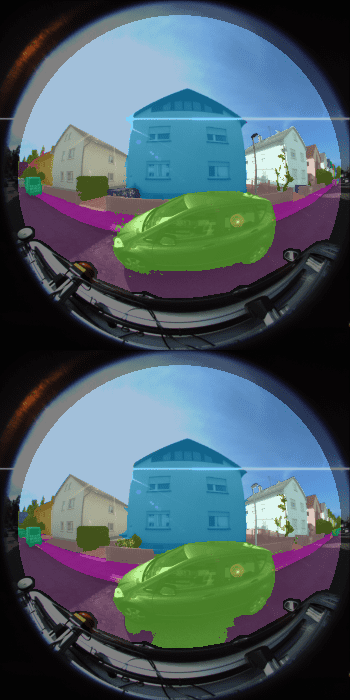}&
  \includegraphics[width=\mywidth]{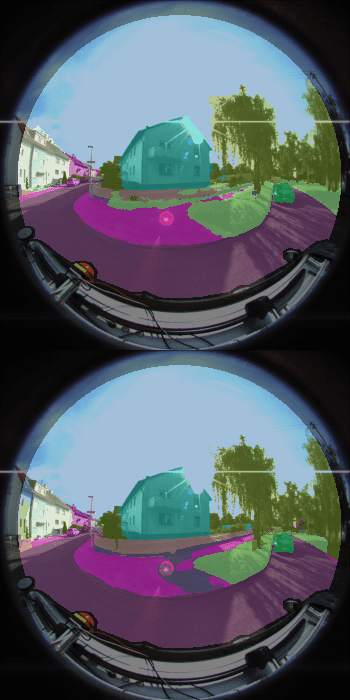}&
  \includegraphics[width=\mywidth]{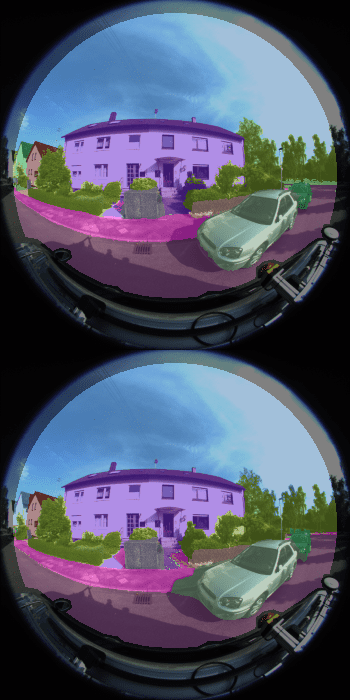}&
  \includegraphics[width=\mywidth]{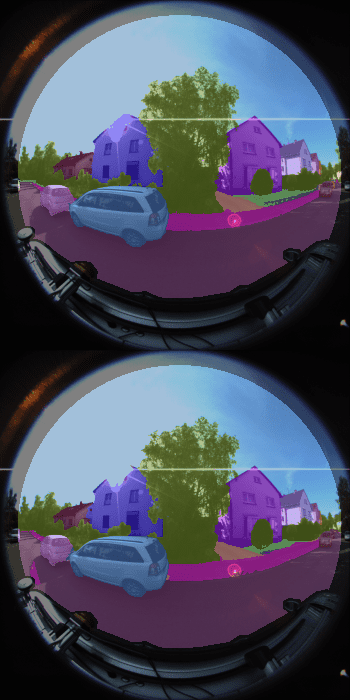}&
  \includegraphics[width=\mywidth]{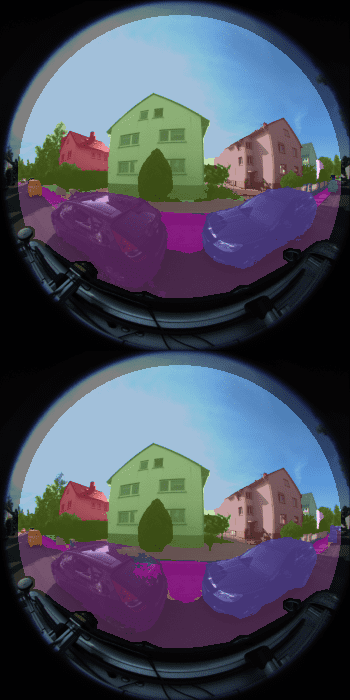}

 \end{tabular}
 \caption{
  \textbf{Qualitative Comparison of Fisheye Panoptic Label Transfer on frames without manually labeled ground truth}. Each group shows the prediction of ours (top) and 3D-2D CRF~\cite{liao2021kitti} (bottom). We can infer from groups 2 and 4 that Ours is superior to 3D-2D CRF on over-exposed areas on buildings.
  }
 \label{fig:panoptic_fisheye_comparison}
\end{figure*}

\figref{fig:semantic_perspective_comparison} shows additional qualitative perspective comparisons corresponding to the Table 1 of the main paper. Consistent with the quantitative results, our method outperforms all baselines qualitatively. We further show qualitative comparisons to 3D-2D CRF in terms of panoptic label transfer on a set of unlabeled 2D perspective frames (see \figref{fig:panoptic_perspective_comparison}) and fisheye frames (see \figref{fig:panoptic_fisheye_comparison}).

\subsection{Qualitative Comparison of Scene Representations}
\begin{figure*}[!t]
 \centering
 \newcommand{\mywidth}{0.97\textwidth}
 \setlength\tabcolsep{0.05em}
 \newcolumntype{P}[1]{>{\centering\arraybackslash}m{#1}}
 \def\arraystretch{0.50}
  \begin{tabular}{P{0.5em}P{0.5em}P{\mywidth}}
   \rot{\scriptsize{Ours}}&& \includegraphics[width=\mywidth]{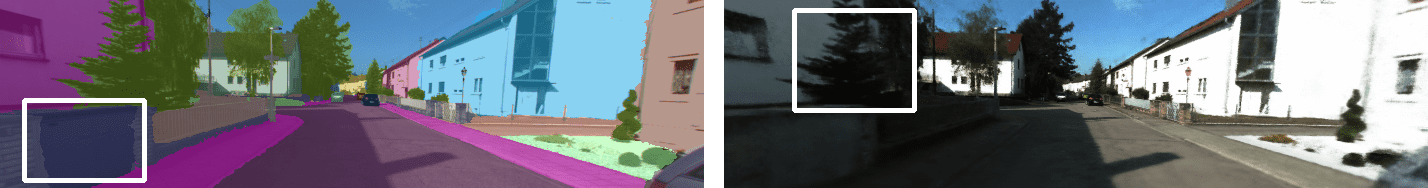}\\
   \rot{\scriptsize{iNGP}}&& \includegraphics[width=\mywidth]{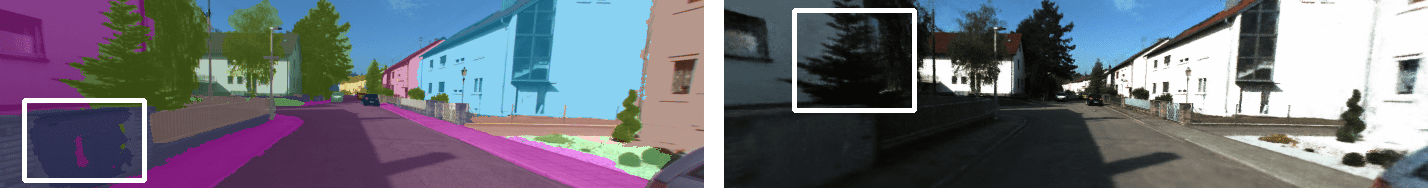}\\
   \rot{\scriptsize{MLP}}&& \includegraphics[width=\mywidth]{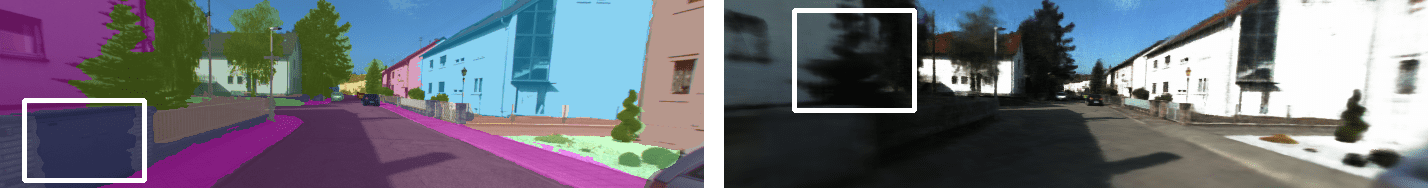}\\
   \rot{\scriptsize{Tri-planes}}&& \includegraphics[width=\mywidth]{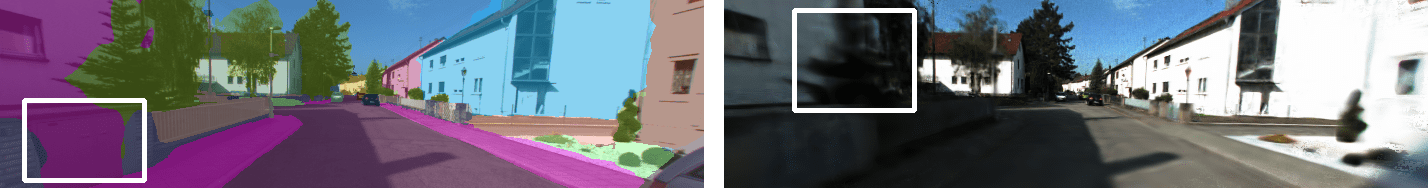}\\
   \\
   \\
   \\
   \rot{\scriptsize{Ours}}&& \includegraphics[width=\mywidth]{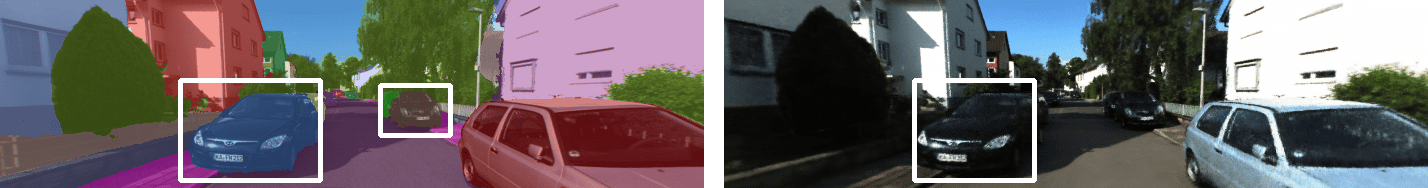}\\
   \rot{\scriptsize{iNGP}}&& \includegraphics[width=\mywidth]{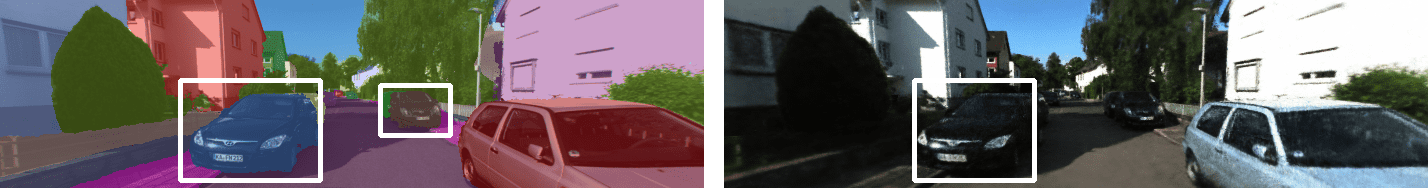}\\
   \rot{\scriptsize{MLP}}&& \includegraphics[width=\mywidth]{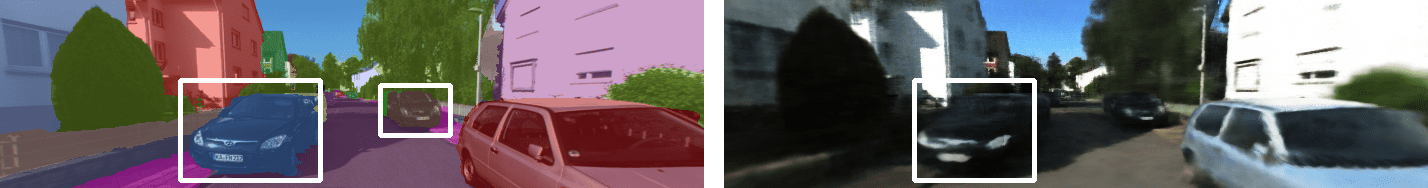}\\
   \rot{\scriptsize{Tri-planes}}&& \includegraphics[width=\mywidth]{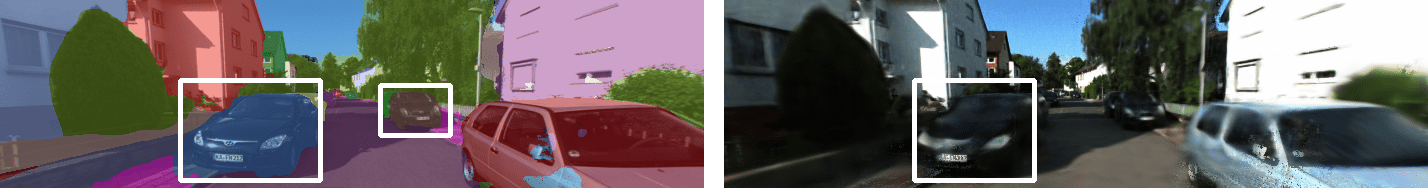}\\
 \end{tabular}
 \caption{
  \textbf{Qualitative Comparison of Neural Scene Representations}. We visualize the predicted panoptic labels overlaid with GT images (left) and rendered appearance (right).
 }
 \label{fig:scene_representation}
\end{figure*}

We study the influence of different scene representations based on pure MLP~\cite{mildenhall2020nerf}, iNGP~\cite{muller2022instant}, and Tri-planes~\cite{chen2022tensorf}. MLP conditions on official 8 fully-connected ReLU layers as the scene feature. iNGP adopts the lower branch of Our's network architecture in \figref{fig:network_supp}, and the network parameters also follows the implementation in \ref{sec:network_architecture}. Tri-planes takes TensoRF-VM-192 ($R_{\sigma}=16, R_{c}=48$) architecture with $512^{3}$ grid voxels. We also involve the total variation (TV) loss to avoid high-frequency noise. The comparison is shown in \figref{fig:scene_representation}. In label synthesis, iNGP generates more undesired noise and ours is comparable to MLP. The inductive smoothness bias benefits MLP to produce smoother boundaries across frequency-varying regions. For rendering appearance, Ours and iNGP are able to reconstruct high-frequency imagery, while MLP is inferior to them. Tri-planes struggles when scaled to larger scenes, while it is excellent at small-scale objects.

\subsection{Weak Depth Supervision}
\begin{figure*}[!t]
 \centering
 \newcommand{\mywidth}{0.95\textwidth}
 \setlength\tabcolsep{0.05em}
 \newcolumntype{P}[1]{>{\centering\arraybackslash}m{#1}}
 \def\arraystretch{0.50}
  \begin{tabular}{P{0.5em}P{0.5em}P{\mywidth}}
    \rot{\scriptsize{Complete Model}}&& \includegraphics[width=\mywidth]{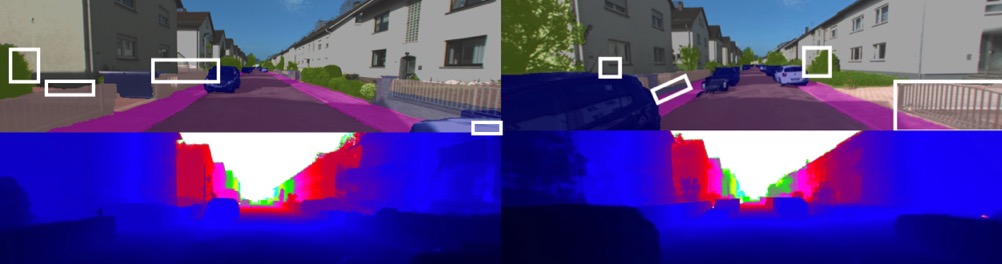}\\
   \rot{\scriptsize{w/ $\cL_d$, w/o $\cL^{2D}_{\hat{\bS}}$}}&& \includegraphics[width=\mywidth]{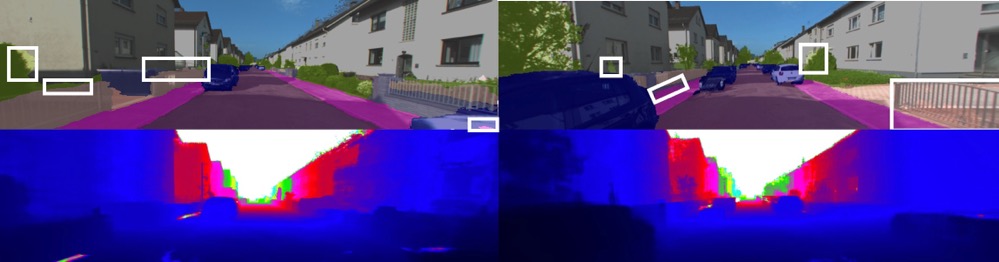}\\
 \end{tabular}
 \vspace{-0.2cm}
 \caption{
  \textbf{Qualitative Comparison of Ablation Study.} We visualize the semantic map and depth map of the complete model (top) and the model without fixed semantic field (bottom). 
 }
 \label{fig:fixcomparison}

\end{figure*}

We show that using the depth loss $\cL_{d}$ alone is not able to recover accurate object boundaries in \figref{fig:fixcomparison}. In contrast, adding the semantic loss $\cL^{2D}_{\hat{\bS}}$ to the fixed semantic field further improves the object boundary. These improvements can be explained as follows: Firstly, the weak stereo depth supervision is not fully accurate, especially at far regions. Furthermore, even with perfect depth supervision, the model receives very small penalty if the predicted depth is close to the GT depth.  In contrast, the cross entropy loss $\cL^{2D}_{\hat{\bS}}$ defined on the fixed semantic field provides a strong penalty as small errors in depth lead to wrong semantics.

\subsection{Analysis of 3D-2D CRF}
\begin{figure*}[tb]
 \centering
 \newcommand{\mywidth}{0.33 \textwidth}
 \setlength\tabcolsep{0.2em}
 \begin{tabular}{ccc}
 
    \includegraphics[width=\mywidth]{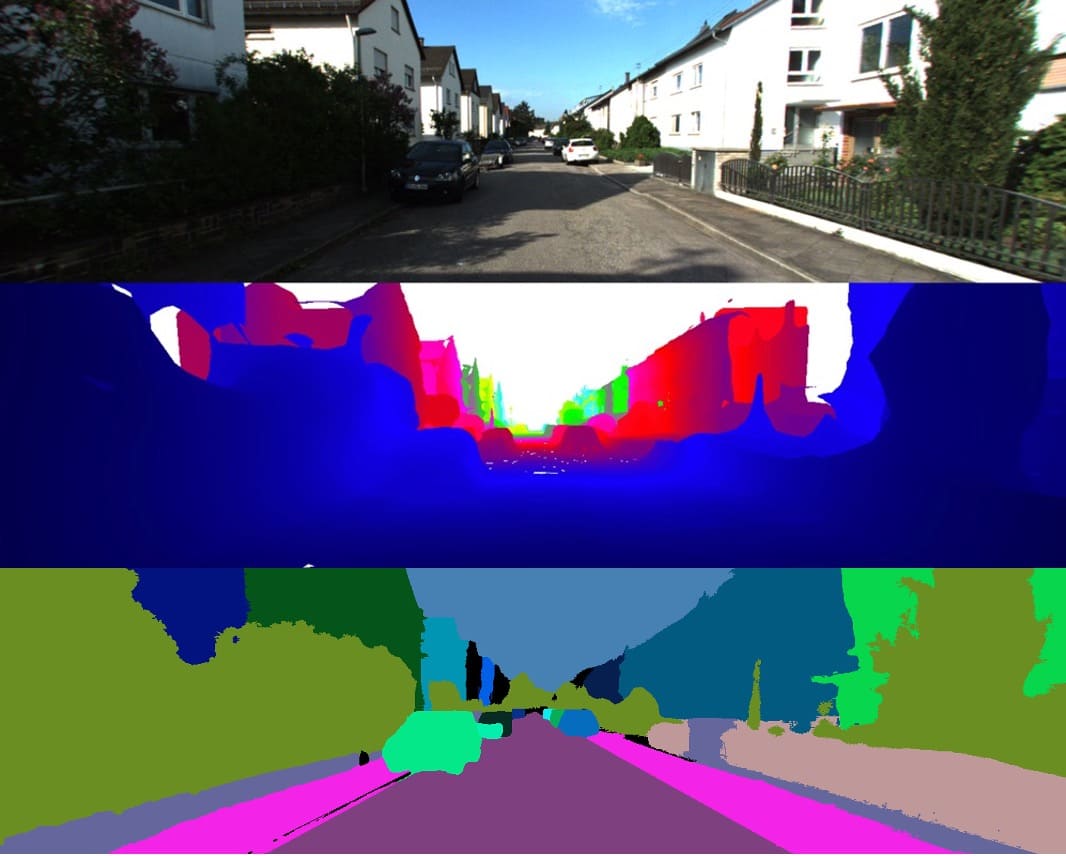}&
    \includegraphics[width=\mywidth]{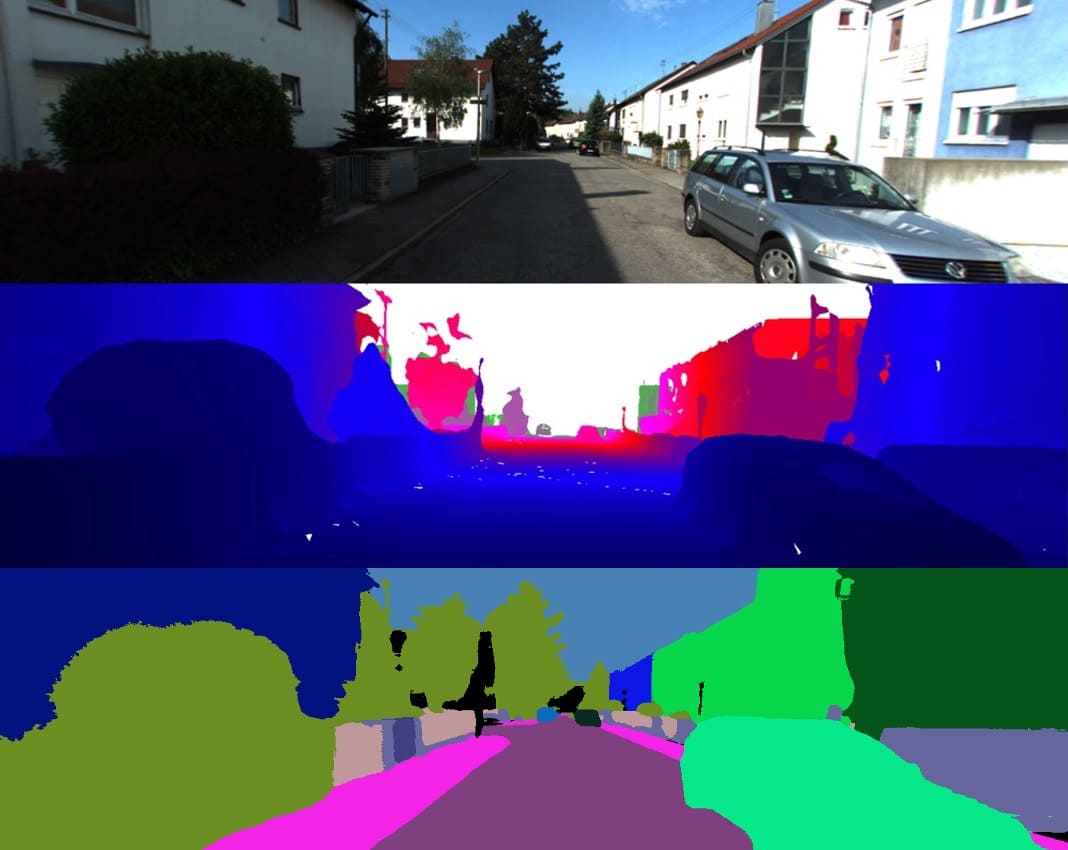}&      
    \includegraphics[width=\mywidth]{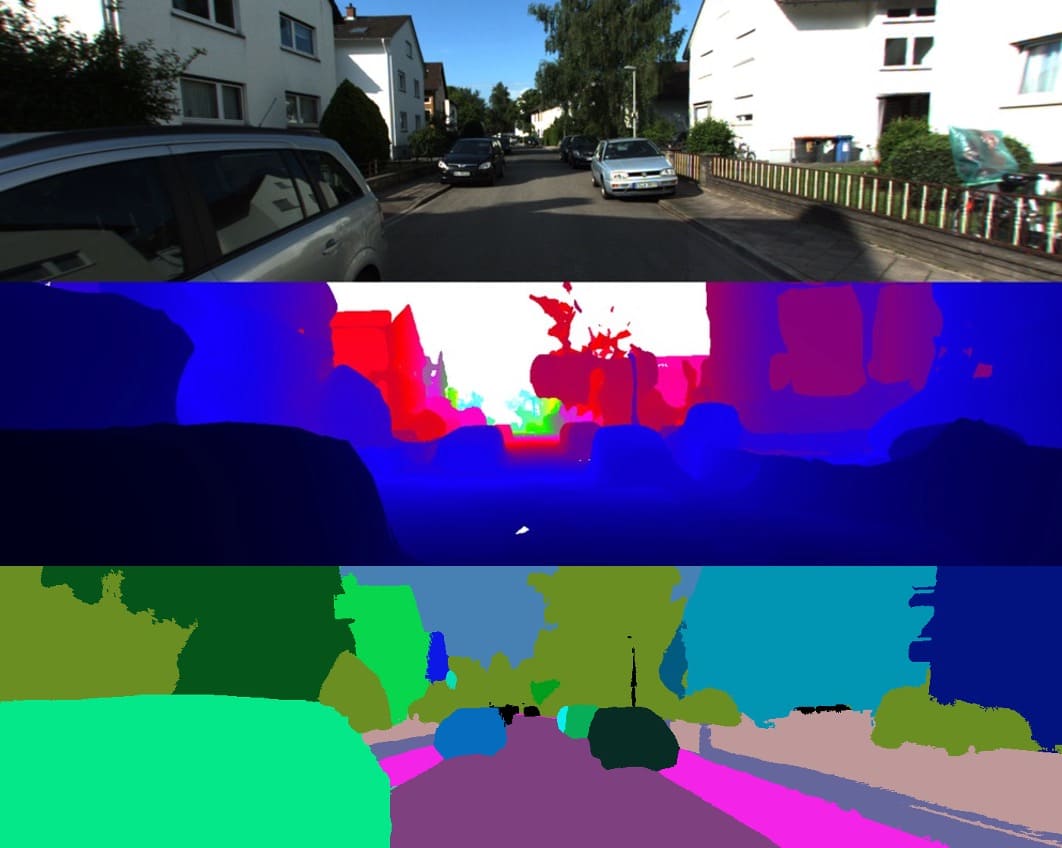}

 \end{tabular}
 \caption{
  \textbf{Qualitative Results of 3D-2D CRF.} Top: Input RGB images. Middle: 3D-2D CRF mesh depth. Bottom: Panoptic label transfer results of the 3D-2D CRF method.}
 \label{fig:crf_mesh}
\end{figure*}

The 3D-2D CRF performs inference based on a multi-field CRF which reasons jointly about the labels of the 3D points and all pixels in the image. To obtain dense 3D points, it accumulates LiDAR observations over multiple frames and project visible 3D points to the image based on a reconstructed mesh. \figref{fig:crf_mesh} shows depth maps of the reconstructed mesh corresponding to Fig. 5 of the main paper. As can be seen, the side of the building can hardly be scanned by the LiDAR, leading to incomplete mesh reconstruction. Consequently, 3D-2D CRF lacks 3D information in these regions and needs to distinguish building instances mainly based on 2D image cues. It is not surprising that the 3D-2D CRF fails at overexposed image regions in this case.

\subsection{Generating Self-distilled Pseudo GT Using SAM}
\begin{figure*}[tb]
 \centering
 \newcommand{\mywidth}{0.33 \textwidth}
 \setlength\tabcolsep{0.2em}
 \begin{tabular}{ccc}
  \includegraphics[width=\mywidth]{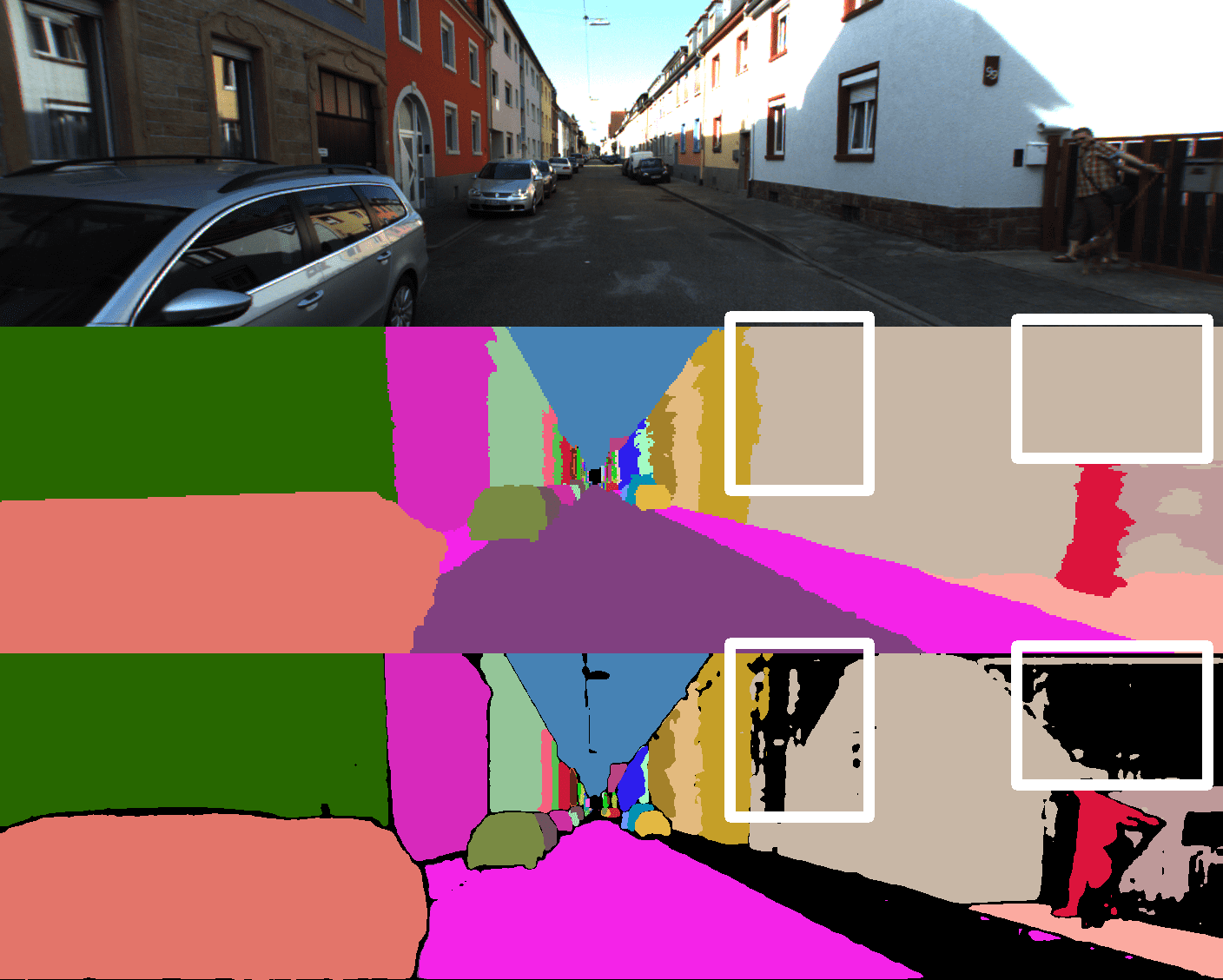}&
  \includegraphics[width=\mywidth]{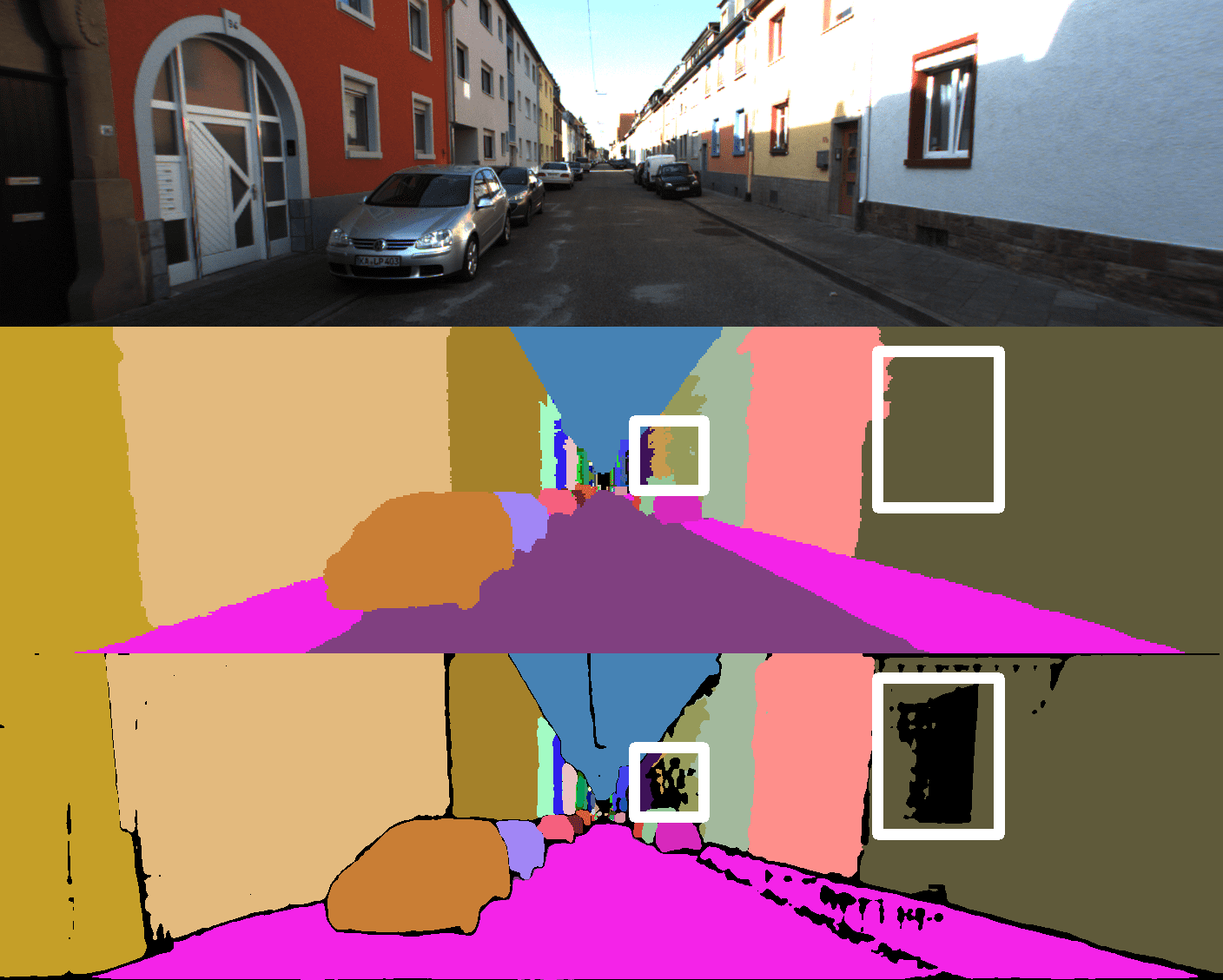}&
  \includegraphics[width=\mywidth]{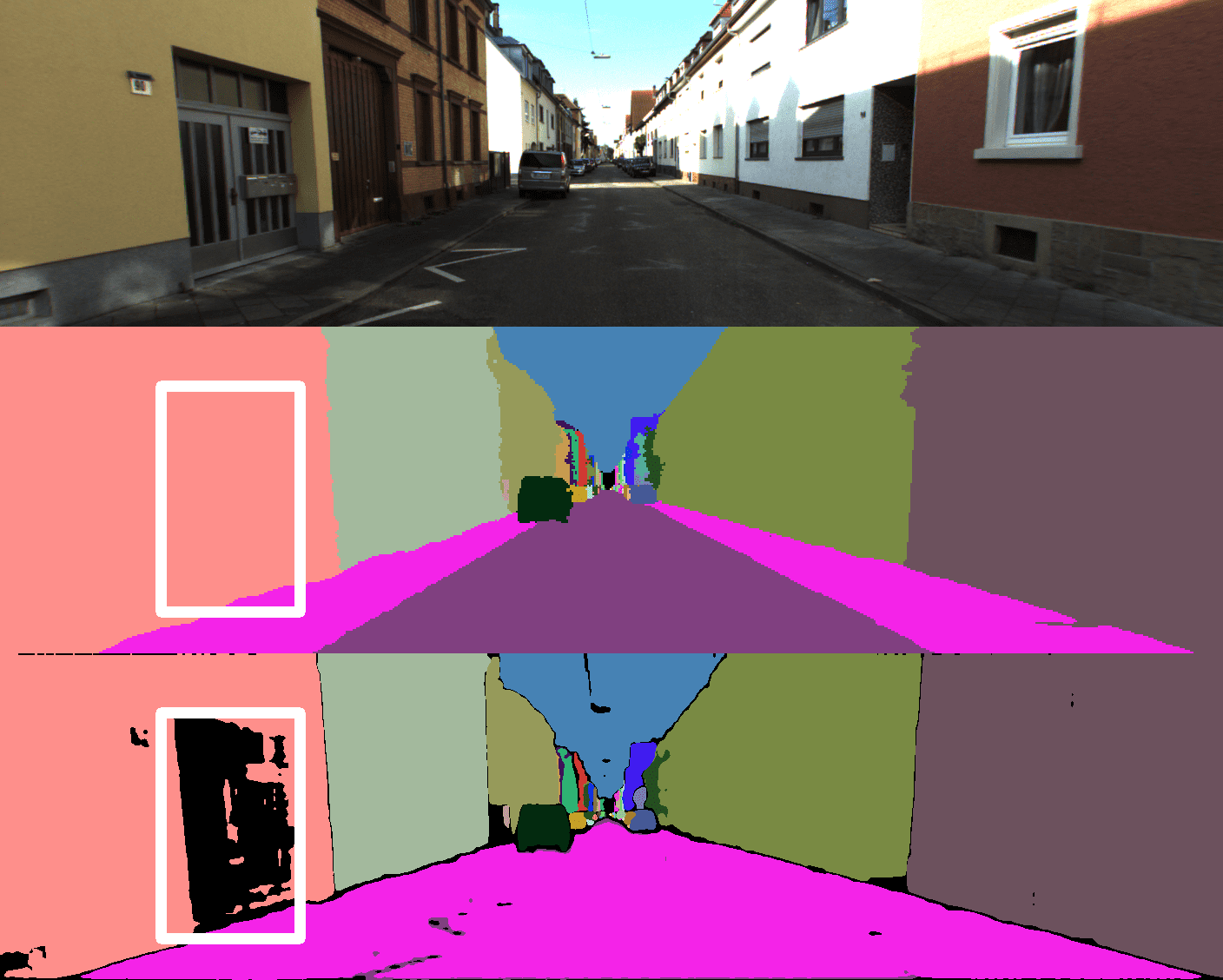}

 \end{tabular}
 \caption{
  \textbf{Self-distilled Pseudo GT Using SAM.} Top: Input RGB images. Middle: Middle results generated by our method. Bottom: Pseudo GT generated by SAM.}
 \label{fig:SAM_pseudo_GT}

\end{figure*}

SAM~\cite{kirillov2023segment} has extraordinary performance on object segmentation and supports multi-forms of input. We try to utilize it to generate general self-distilled pseudo panoptic GT. However, we find its open source code can not support panoptic masks as input and we have to transform them into panoptic 2D bounding boxes before sending them to the network. As shown in \figref{fig:SAM_pseudo_GT}, the pseudo GTs generated by SAM have some flaws in buildings. We suppose that SAM tends to segment some components of the buildings (like doors and windows) separately instead of regarding them as part of the buildings and the performance of SAM can be deteriorated by over-exposure.

\section{Failures} \label{failure}

\subsection{Far-Region Label Synthesis}

\begin{figure*}[tb]
 \centering
 \newcommand{\mywidth}{0.33 \textwidth}
 \setlength\tabcolsep{0.2em}

 \begin{tabular}{ccc}

  \includegraphics[width=\mywidth]{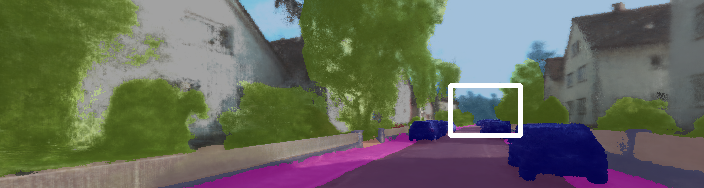}&
  \includegraphics[width=\mywidth]{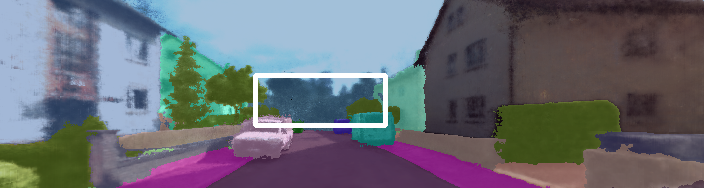}&
  \includegraphics[width=\mywidth]{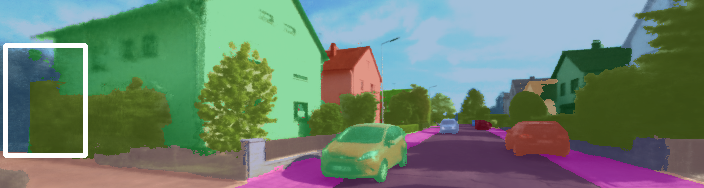}

 \end{tabular}
 \caption{
  \textbf{Failure in Label Synthesis in Far-Region at Novel Viewpoints}. }
 \label{fig:failure_far}

\end{figure*}

As there are missing bounding boxes in far-regions, labels rendered at novel viewpoints at these areas will be classified to ``sky'' as shown in \figref{fig:failure_far}. Although we can improve the label quality via pseudo label fusion in overfitted views in \ref{fuse_farregion_class}, the ability to render precise labels in regions with arbitrary distances at omnidirectional viewpoints remains a problem.

\subsection{Fisheye Geometric Reconstruction}

\begin{figure*}[tb]
 \centering
 \newcommand{\mywidth}{0.195 \textwidth}
 \setlength\tabcolsep{0.2em}
 \begin{tabular}{ccccc}
  \includegraphics[width=\mywidth]{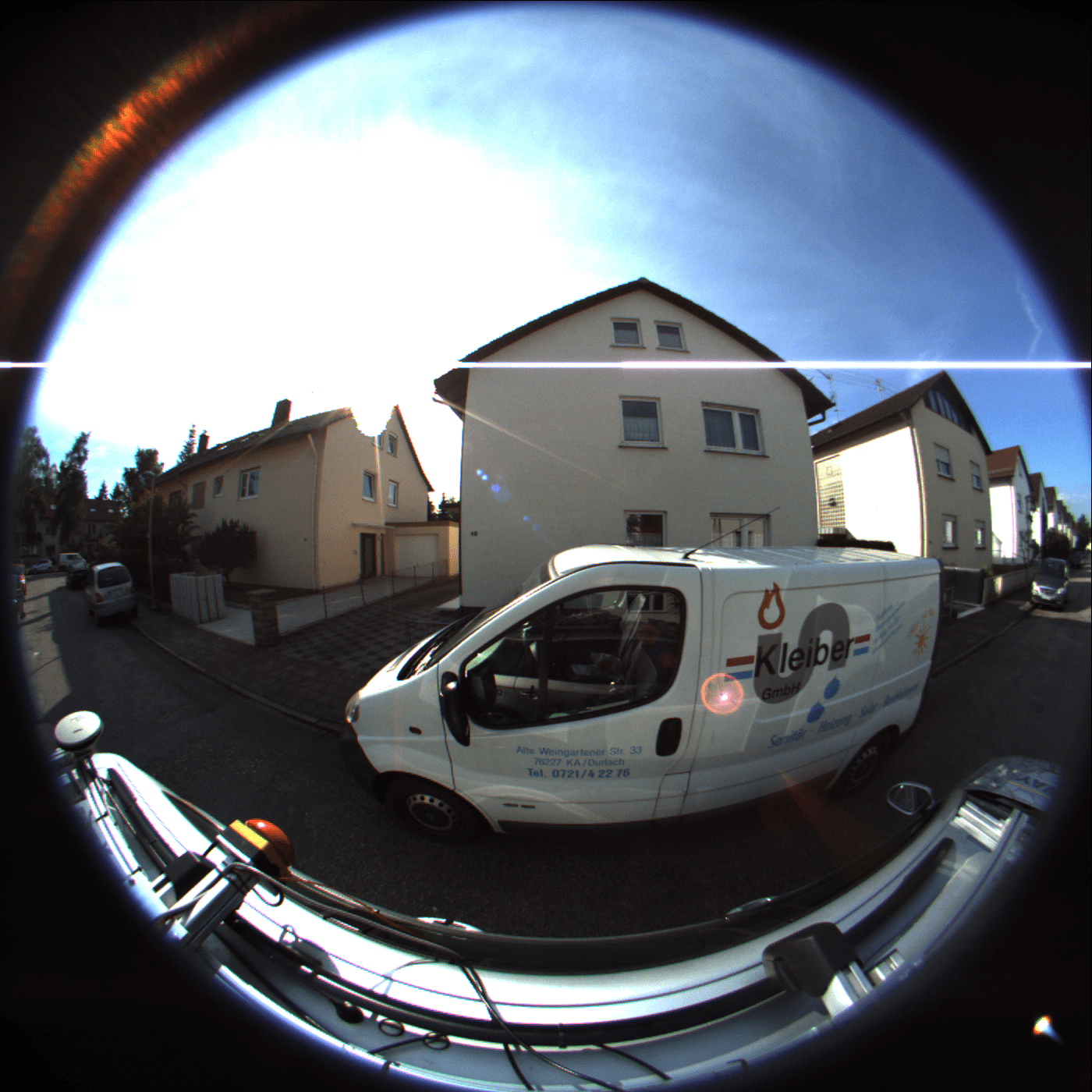}&
  \includegraphics[width=\mywidth]{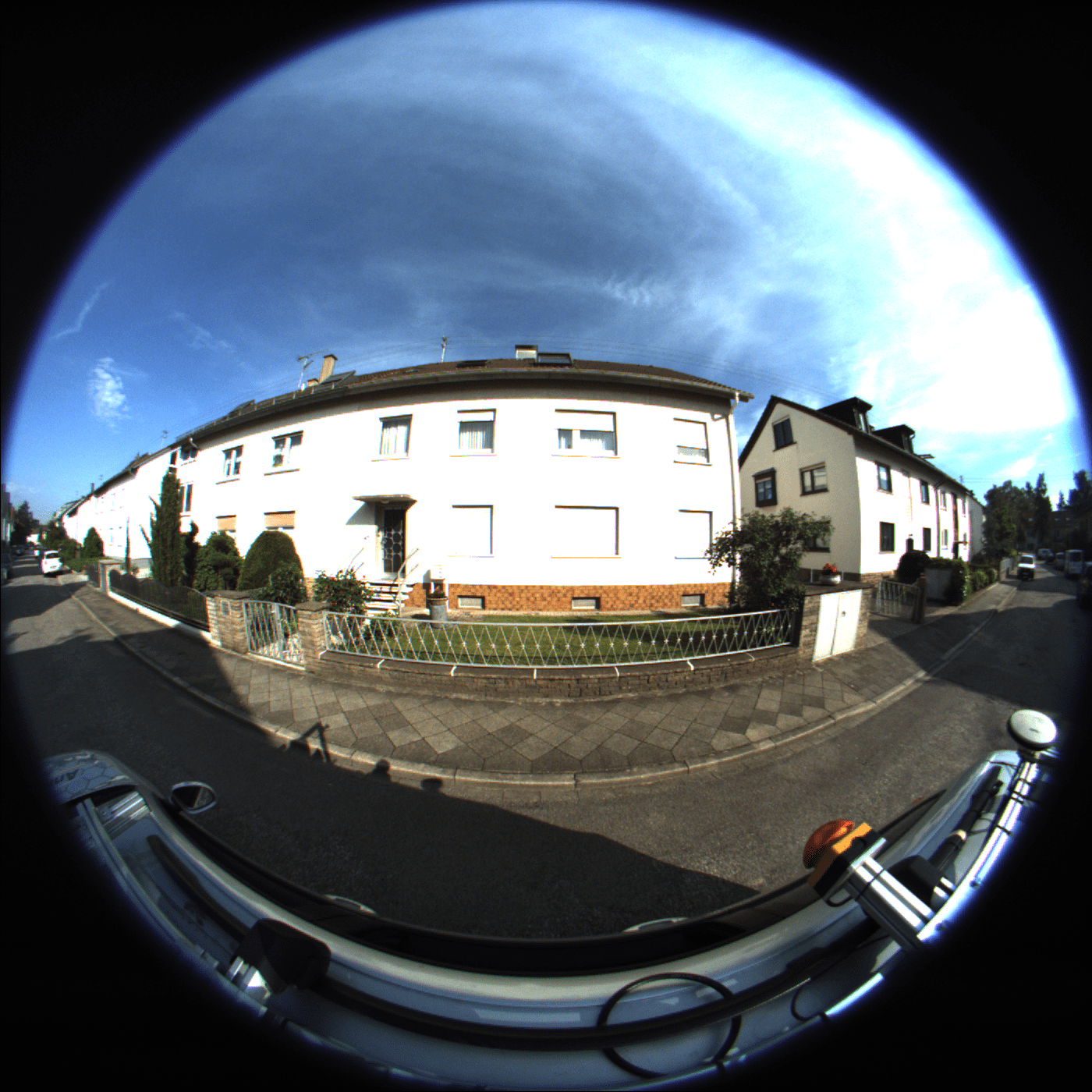}&
  \includegraphics[width=\mywidth]{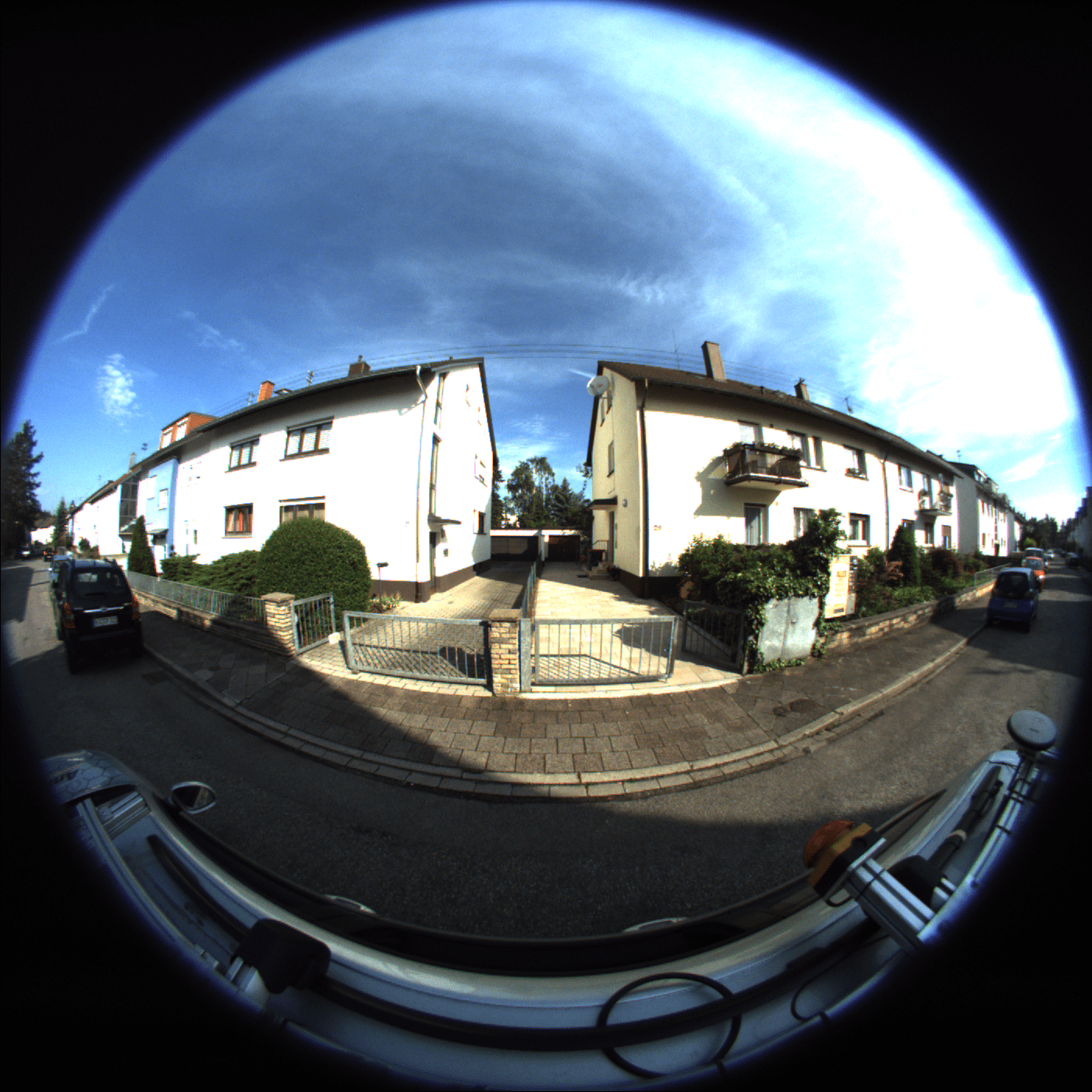}&
  \includegraphics[width=\mywidth]{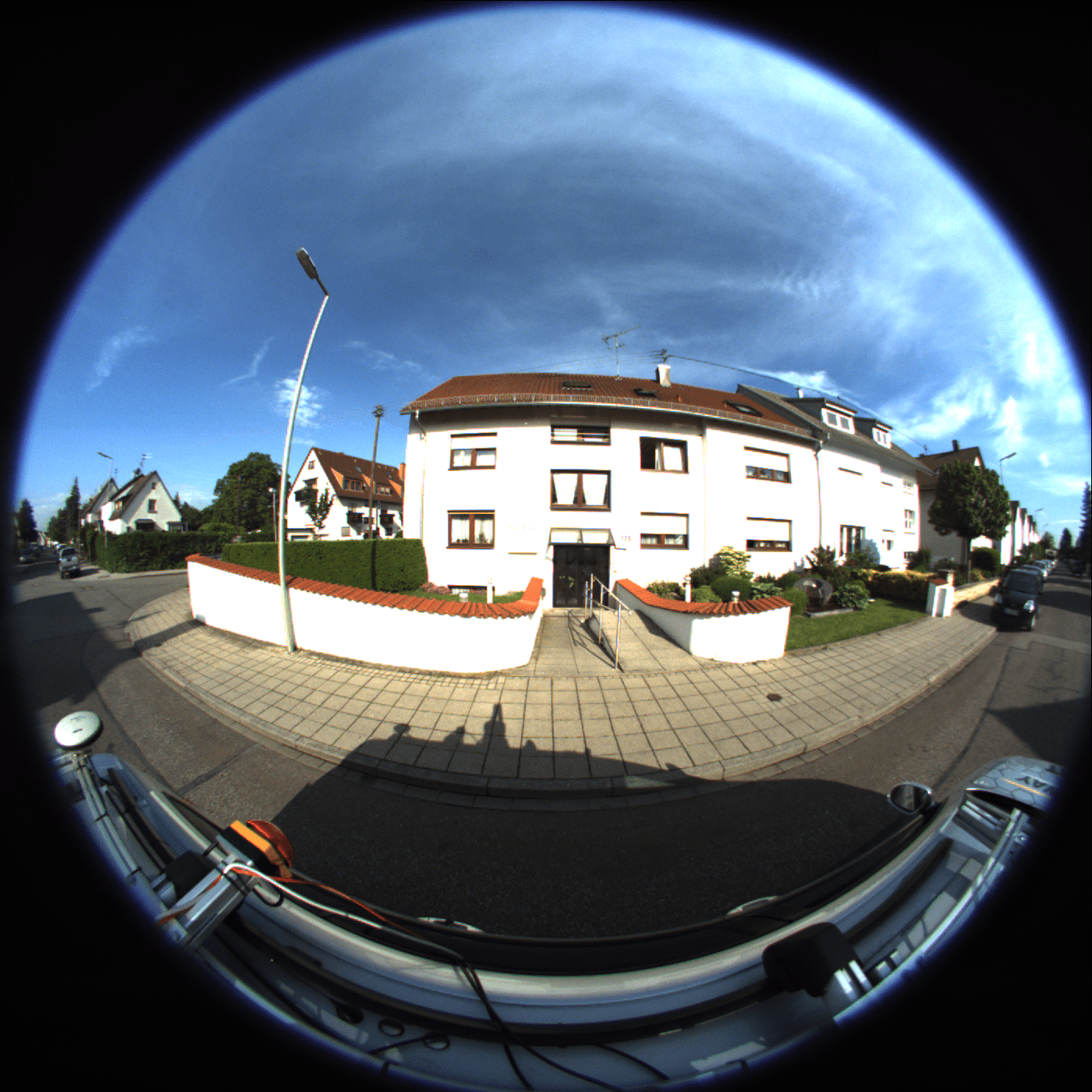}&
  \includegraphics[width=\mywidth]{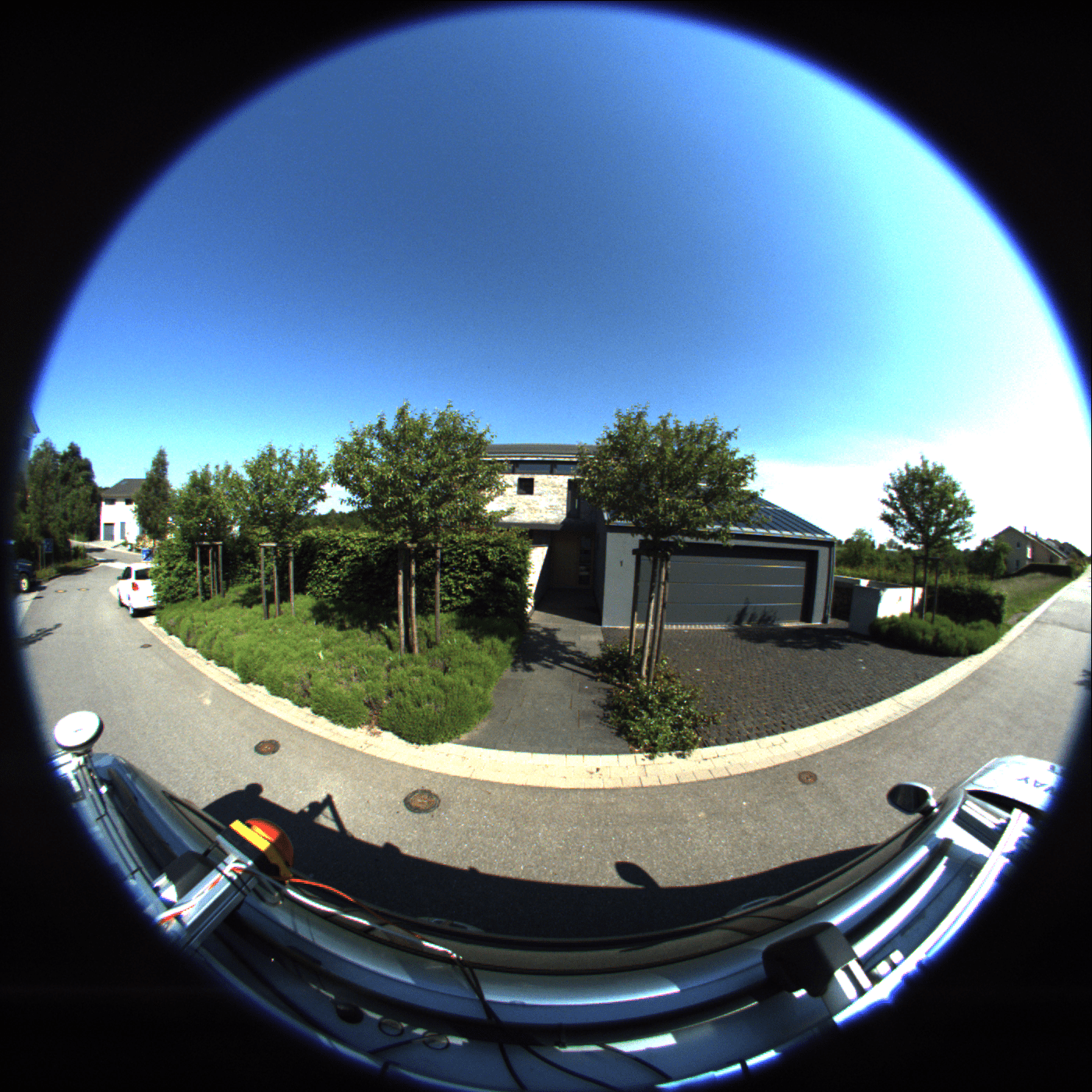} \\
  \includegraphics[width=\mywidth]{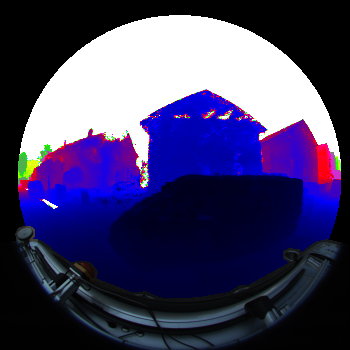}&
  \includegraphics[width=\mywidth]{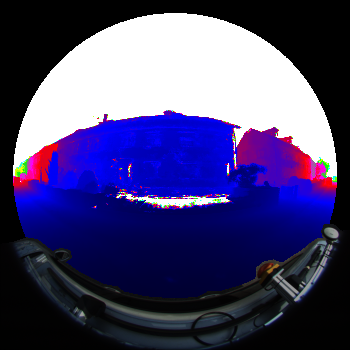}&
  \includegraphics[width=\mywidth]{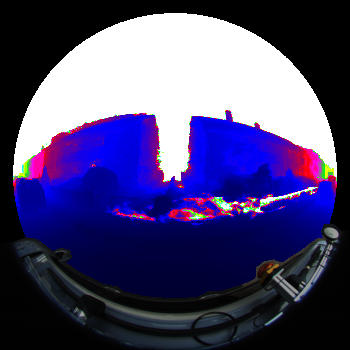}&
  \includegraphics[width=\mywidth]{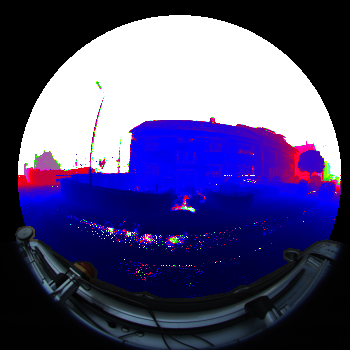}&
  \includegraphics[width=\mywidth]{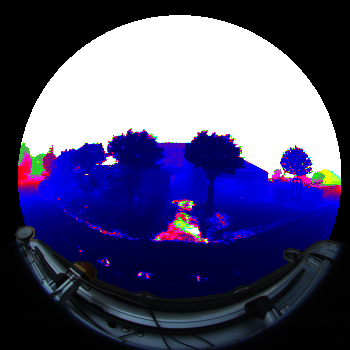}
 \end{tabular}
 \caption{
  \textbf{Failure in Geometric Reconstruction on Fisheye Views.} Each group contains a fisheye RGB input (upper) and reconstructed depth map (bottom).
  }
 \label{fig:failure_fe_depth}
\end{figure*}

In our experiments, we find that the geometric reconstruction in two-side fisheye is unstable though perspective information could serve as a complementation. As illustrated in \ref{fig:failure_fe_depth}, there are irregular holes in the reconstructed geometry especially in low-texture and over-exposed regions.

\end{appendices}

\end{document}